# Knowledge Reduction and Discovery Based On Demarcation Information

**Yuguo HE** heyuguo@163.net
*Department of Computer Science, Beijing Institute of Technology*
*Beijing, P.R. China, 100081*

**Abstract**

Knowledge reduction, includes attribute reduction and value reduction, is an important topic in rough set literature. It is also closely relevant to other fields, such as machine learning and data mining.

In this paper, an algorithm called TWI-SQUEEZE is proposed. It is so named because it can find a *reduct*, or an irreducible attribute subset that maintains certainty of classification, after two scans, which is a bit similar to the process of squeezing water from sponge by *pressing on both sides of the sponge. Its soundness and computational complexity are given, which show that it is the fastest algorithm at present*.

A measure of difference contained in a system, or variety, is brought forward. The quantity character of it is measured by demarcation information, of which algorithm TWI-SQUEEZE can be regarded as an application. Demarcation information measure can also be used as heuristic information to guide the algorithm to find a suboptimal *reduct*. In this paper, this measure will be compared with Shannon entropy in details. Some basic concepts, such as uncertainty, distinctiveness, variety, similarity, difference and their relationships will be studied. The author also argues the rightness of this measure as a measure of information, which can make it a unified measure for "*differentiation*", a concept appeared in cognitive psychology literature.

Value reduction is another important aspect of knowledge reduction. It is interesting that using the same algorithm we can execute a complete value reduction efficiently. A complete knowledge reduction, which results in an irreducible table, can therefore be accomplished after four scans of table. The byproducts of reduction are two classifiers of different styles.

Traditionally, attribute reduction is regarded as data preprocessing phase in knowledge discovery. The author will show in this paper that, knowledge reduction, especially approximate knowledge reduction which is based on attribute reduction, is a process of data mining.

In this paper, various cases and models will be discussed to prove the efficiency and effectiveness of the algorithm. Some topics, such as how to integrate user preference to find a local optimal attribute subset will also be discussed.

## 1. Introduction

How to draw useful knowledge from a table, especially a decision table, is an important topic in data mining and knowledge discovery. A decision table is a relational table with a special dimension[1] which indicates the class or category an object belongs to. A decision table can be regarded as a matrix. The problem of knowledge reduction can be addressed as how to remove elements of the matrix without violating its performance of classification, and in the same time ensure that further removal of it will definitely decrease the accuracy of classification. From this definition, knowledge reduction can be regarded as finding an irreducible description of the original table. However, the techniques invented are also useful for prediction. In this paper, I'll show why knowledge reduction is a kind of knowledge discovery, and the concepts and ideas that can not only used to tackle this problem, but also can benefit to other researches.

One of the main objectives of knowledge reduction is attribute reduction, or dimensionality reduction. However, in rough set theory, attribute reduction is a bit special from traditional dimensionality reduction in that, it aims to find *reduct*, a subset of the original attribute set, which satisfies two conditions: 1. it can replace the original attribute set *without* loss of accuracy of classification. 2. if remove any attribute in the reduct, it will definitely cause new inconsistency, and as a consequence violate the accuracy of classification. The concept "reduct" is a contribution of rough set theory, and therefore it is discussed mainly in rough set literature. However, it can also benefit to other relevant areas. For example, it reveals how "complicated" a function is, according to definition 4 of relevance (Blum&Langley, 1997); the process of finding reducts can be used in feature selection, and so on. This paper is just such an illustrative example.

In a decision table, all the attributes are divided into two classes: condition attributes and decision attributes. Reduct in decision table is also called as *relative reduct*. For two rules, if they satisfy the same condition but have different decisions, we say they cause inconsistency in decision table. A relative reduct won't cause new inconsistency to original table, which means it maintains classification accuracy of the original table. In this paper, I will still call "relative reduct" as "reduct" for simple.

Clearly, finding a reduct, especially an approximate one, has something to do with feature selection[2]. Their relationship will be discussed in section 7.8.1. Indeed, all the heuristic algorithms mentioned in section 5 can be regarded as counterparts of sequential suboptimal search methods in feature selection literature. And it is clear that achievements of either problem will benefit to the other.

Finding reducts in large relation database is a basic problem in the increasing applications of rough set theory (Pawlak, 1991; Shi, 2002), which is also one of the bottlenecks of the rough set methodology (Komorowski, Pawlak, Polkowski, &Skowron, 1998; Drewry, Gu, Hocking et al., 2002). Here I narrow my interest in decision table, because a non-decision table can be transferred to a decision table efficiently. There are many researches on this topic and two objectives lie in these works. One is to find a minimal reduct, which is proved to be NP-hard (Rauszer, 1991). All the known algorithms use some heuristic information and their strategies are greedy in nature. Therefore they are all incomplete for this problem (Wang&Miao, 1997). Another goal is to find one reduct. It especially makes sense in large database. Several recent applications, especially those in data mining, involve thousands of attributes. In such case, the computational requirement, such as time and space, is extremely important. Therefore a suboptimal algorithm is needed to trade off optimality for computational efficiency, since all the known optimal algorithms are computationally prohibitive for large database in practical applications, a similar case to feature selection (Jain, Duin, & Mao, 1999). In this paper the latter goal is in pursuit of.

---

[1] Without this dimension, the table can be called information system, or non-decision table here. It is proved that any non-decision table can be transformed to decision table efficiently.

[2] There are mainly two families of methods that are used in dimensionality reduction. One *selects* subset of input attributes. The other *creates* new features to replace the original attributes, which is often called *extraction*. Attribute reduction and feature selection both belong to the first "*selection*" family.

There were many algorithms aimed at finding reducts in a decision table previously. They can mainly be classified into two classes. One is based on observation of change of positive region. The other is based on Discernibility Matrix.

There are two subclasses of the first sort of method. One is the classical data analysis method (Pawlak, 1991; Liu, 2001), which can be called *backward elimination*[3] method. The thought is to remove attributes in the condition attribute set one by one and observe the change of positive region. Although classical methods based on comparing objects are very slow, Nguyen Sinh Hoa et al. (Hoa & Son, 1996) found an efficient way to find a reduct, by evaluate cardinality of positive region efficiently[4]. The essential point is how to compute positive region efficiently. This kind of method is similar to Sequential Backward Selection (SBS) method in feature selection literature.

On the contrary, the other methods, which can be called *forward selection* method, adds attributes to the core gradually and observes the change of positive region (e.g. Wang et al., 1998. And Jelonek et al., 1995, according to Shen et al., 2001, which is also an example). This kind of method is similar to Sequential Forward Selection (SFS) method in feature selection literature. But we must cautious that when we add one attribute, it is possible that we must remove more attributes, which was ignored by many previous works, such as the algorithms introduced by Hu et al. (1994), Miao and Wang (1997), Wang et al. (1998), etc.

(Conditional) Information entropy (Yu, Yang, Wu, &Li, 2001; Wang, Yu &Yang, 2002), mutual information (Miao & Wang, 1999; Miao& Hu, 1999) and significance of attribute defined by change of positive region (Pawlak, 1991; Hu, Pao, Yu, 2002) are the heuristic information adopted in this family of methods. They are used to find a suboptimal reduct.

The other popular method uses Discernibility Matrix as a tool and compute Discernibility Function to find reducts (Skowron & Rauszer, 1992). There are many varieties of this method (e.g. Liu, 2001a, 2001b; Wang & Wang, 2001). It is efficient when core is relative large in reduct and the table is small. And it is the best choice when effectiveness is stressed because it is complete for finding all reduct and optimal reduct.

Occurrence frequency of an attribute in discernibility matrix is used as heuristic information in this kind of methods (Hoa &Son, 1996; Wang et al., 1998).

After comparing all these methods, we can see that, as far as the second goal is concerned,

---

[3] This is a term borrowed from feature selection literature, which can also release the close relationship between attribution reduction and feature selection.

[4] However, in this paper *reduct* is defined as a minimal subset of attributes that maintains indiscernibility relation: “A minimal subset *B* of *A* such that *IND*(*A*)=*IND*(*B*) is called a reduct”. It is not true. What I will show in this paper is that a reduct is a minimal subset of attributes that maintains the demarcation relation.

Nguyen Sinh Hoa’s method will work better, if it remove repetitive objects first after sorting.

the work of Nguyen Sinh Hoa et al. is the fastest among these methods.

Some works use genetic algorithm to find reducts (Wróblewski, 1995, 1998). However, the computational complexity of genetic algorithms cannot be analyzed thoroughly and it is impossible to obtain their computational complexity except computing price of fitness function. Therefore, although they maybe efficient in some case, they will not be discussed in this paper.

I have also designed a series of efficient algorithms (He, 2003)**.**

In this paper I bring forward a new algorithm based on previous work, which can find a reduct more efficiently than any other works proposed previously. Moreover, it is suitable for parallel computing and has a good increment version.

Complete value reduction is a process to find a subset of values in table which satisfies that removing any value in this subset will definitely cause new inconsistency. That is, some classification ability of the original table is lost. It is a problem received less concern in rough set literature and there is no efficient enough method for long.

Fortunately, we will see that after running the algorithm TWI-SQUEEZE twice, both attribute reduction and the corresponding value reduction are finished, which make up a whole knowledge reduction efficiently. It will be discussed in section 6. Furthermore, knowledge reduction can be regarded as a process of selective inductive machine learning. The results of knowledge reduction by algorithm TWI-SQUEEZE are two different sorts of classifiers. Because of the efficiency and scalability of the algorithm, it can be used in data mining and knowledge discovery. There are some excellent books that introduce data mining and knowledge discovery from different aspects (Zhou, 2003), such as the books written by Han and Kamber (2000), Witten and Frank (2000), Hand and Mannila (2001). Some reviews and surveys are also available, such as the articles written by Chen, Han and Yu (1996), Mitchell (1999), Drewry, Gu and Hocking et al.(2002), Mitra, Pal and Mitra (2002), etc. Tutorials can be found on the web, such as http://www-2.cs.cmu.edu/~awm/tutorials.

The rest of this paper is organized as follows.

In section 2, algorithm TWI-SQUEEZE will be proposed and analyzed in details. Section 2.1 discusses the basic ideas and concepts in this algorithm. Section 2.2 gives outline of the algorithm and a running example to illustrate the process. In section 2.3 its computational complexity will be analyzed. Section 2.4 discusses how to find relative core and irrelevant attributes.

In section 3, parallel versions of algorithm TWI-SQUEEZE will be discussed shortly.

In section 4, scalability of the algorithm will be discussed. In section 4.1, its incremental version will be discussed in details. An example will illustrate the process. Its decrease version will be mentioned briefly in section 4.2. In section 4.3 we will talk about how to renew original reduct efficiently to adapt to new situations and requirement. All those discussions show that algorithm TWI-SQUEEZE is a scalable algorithm which can adapt to change of need efficiently.

Section 5 discusses how to find a short reduct. Section 5.1 is one of the *emphases* of this paper, in which I propose a new sort of information measure and prove some key properties of it. Algorithm TWI-SQUEEZE can be regarded as an application of it[5]. In this section I'll discuss measure of information, variety or difference contained in a system, and their relationship. Measure of similarity is the same as that of difference, since similarity and difference are two sides of one coin. This section can be read alone, without concerning too much of Appendix D. Section 5.2 discusses how to utilize this kind of information to guide algorithm TWI-SQUEEZE to find a short reduct. Section 5.3 discusses how to renew a reduct to a shorter one.

Section 6 discusses value reduction, which combines with attribute reduction to make up a complete knowledge reduction. Section 6.1 gives description of an algorithm that can perform a complete value reduction. Section 6.2 talk about another algorithm that handle value reduction more efficiently, regardless of the fact that incomplete table is created in the process. Section 6.3 analyzes the computational complexity of the whole knowledge reduction, including attribute reduction and value reduction. Section 6.4 discusses incomplete and approximate value reduction. Section 6.5 discusses how to find core values that cannot be omitted in table.

In section 7, we will discuss some relevant and important topics thoroughly. Section 7.1 discusses how to perform knowledge reduction on non-decision table, using the same algorithm TWI-SQUEEZE. Section 7.2 is an interesting result, which discusses how to integrate user preference to find a local optimal attribute subset.

In section 7.3, we'll discuss models of approximate reduction. Section 7.3.1 gives an advice to handle non-monotonicity in reduction based on variable precision model. Section 7.3.2 gives a solution to reduction based on graded rough set model briefly. The functions of approximation will be mentioned shortly in Section 7.3.3.

In section 7.4, we talk about incomplete cases in data mining. In fact, I've proposed methods to handle incomplete data efficiently in section 6.1, 6.2. In this section, we mainly discuss incomplete phenomenon itself rather than algorithms to solve it.

In section 7.5, the importance of data preprocessing techniques, especially discretization and concept hierarchy generation, are stressed.

In section 7.6, dynamic redcut will be introduced. The possibility of sampling on attributes is also discussed.

In section 7.7, how to find more reducts will be discussed briefly.

In section 7.8 we will discuss some closely relevant topics.

Feature selection will be compared with attribute reduction in section 7.8.1. How to conduct feature selection using algorithms proposed in this paper is discussed.

It is interesting that, after knowledge reduction, we have two identical classifiers: one is a rule system and the other is a tree that is different from classical decision tree. It will be discussed in section 7.8.2.1. A novel classifier called order-deterministic classifier tree is proposed in section 7.8.2.2.

In section 7.8.3 we will discuss two topics in a casual way. One topic is from where comes generalization in knowledge reduction. The other topic is the inductive bias of algorithm TWI-SQUEEZE, and how can we amend the algorithm a bit in value reduction, assimilating

[5] However, I designed the algorithm independently before I discovered the measures.

the advantage of AQ11.

In section 7.9, the soundness of algorithm TWI-SQUEEZE in tolerance approximation space model, which has been proved in Appendix A, will be discussed in details by examples. Some method is proposed to improve its efficiency in this case.

Computational complexity of various attribute reduction algorithms will be compared in section 8.

At last, some conclusions will be made in section 9.

In Appendix A, I'll prove the soundness of algorithm TWI-SQUEEZE, based on tolerance approximation space model, of which Pawlak's rough set model is a special case.

Some special technique details of algorithm will be mentioned in Appendix B.

In Appendix C, a small example will be given, which illustrates the process of computing amount of demarcation information.

Appendix D is an indispensable complement for section 5.1. However, it can be read alone. In this sense, it can be treated as a separate part from section 5.1.

## 2. Squeeze a Reduct from Large Decision Table

A decision table or knowledge system can be represented as $KS=(U, C \bigcup D)$. Here $U$ is a non-empty, finite set, which is called universe; an element in the universe is called an object (or a record, a rule); $C$ and $D$ are all non-empty, finite set of attributes. $C \bigcap D=\emptyset$. For any $a \in C \bigcup D$, $a$: $U \rightarrow V_a$, where $V_a$ is called the value set or domain of $a$. $C$ is condition attributes set and $C_i\,(1 \le i \le t_C)$ denotes the $i$ th condition attribute. $D$ is decision attributes set. $t_C$ is the number of condition attributes and $t_D$ is the number of decision attributes[6]. After sorting under $D$, decision table is divided into $r_D$ regions. I call these regions as "*D-Region*". $r_D$ is the number of equivalence classes sorted by decision attributes.

### 2. 1 The Basic Elements of Thought

In a table, an attribute can generally induce an equivalence relation or indiscernibility relation. In other words, it can induce a partition of the universe. The complement of equivalence relation can be called as discernibility (Skowron & Stepaniuk, 1996) or demarcation relation. For example, the complement of equivalence relation $R=\{(a, a), (a, b), (b, a), (b, b), (c, c)\}$ is $R'=\{(a, c), (b, c), (c, a), (c, b)\}$. These two sets are complementary with respect to $\{a, b, c\} \times \{a, b, c\}$. In this case, the values under the attribute is $\{a, b, c\}$, and the demarcation relation induced by the attribute is $\{(a, c), (b, c), (c, a), (c, b)\}$.

$(x, y)$ is called a *demarcation* (or $x$ and $y$ are demarcated), if it is an element in a demarcation relation.

In a decision table, all the objects are grouped into some classes by decision attribute value. These classes can be called as "D-Regions"[7]. Two objects are demarcated by decision attributes if they belong to different D-Regions. The nature of attribute reduction is that any condition attribute can be removed safely only if any two objects in different D-Regions that can be demarcated by it can also be demarcated by other condition attributes.

---

[6] A decision table can also be denoted as $KS=(U, C \bigcup D, V, f)$, where $V$ is the united domain of attributes, and $f: U \times (C \bigcup D) \rightarrow V$.

[7] Formally, any set belongs to $U/D$ is called a D-Region. $U/D$ is a set composing of all the equivalent classes induced by attribute D.

Here is a small example. In table 1, object 1, 2 belong to one class, which is labeled by "0", and object 3, 4, 5 belong to the other class, which is labeled by "1", according to decision attribute *D*. Therefore, we know at once that there are two D-Regions (D-Region "0" and D-Region "1") and object 1, 2 are demarcated from objects 3, 4, 5. All the demarcations, e.g.

| | $C_1$ | $C_2$ | $C_3$ | D |
|---|---|---|---|---|
| O.1 | 1 | 0 | 1 | 0 |
| O.2 | 0 | 0 | 1 | 0 |
| O.3 | 1 | 0 | 0 | 1 |
| O.4 | 1 | 1 | 0 | 1 |
| O.5 | 0 | 1 | 1 | 1 |

Table 1: a small example

(1, 3), form the demarcation relation of attribute *D*. Furthermore, if the table is consistent[8], then all these demarcations induced by attribute *D* must also be induced by condition attributes for any attribute reduction algorithm that aims to find reducts. The attribute reduction problem can be regarded as: can we find a subset of condition attributes which can offer (induce) these demarcations and in the same time any further reduction will surely fail?

In the above case, we find that $C_3$ can offer all the demarcations except the demarcations between objects 1, 2 and object 5: {(1, 5), (2, 5)}. Afterwards, we find that these two demarcations can be offered by $C_2$, which means $C_2$ and $C_3$ can offer all the necessary demarcations. Therefore, $C_1$ is a redundant attribute relative to $\{C_2, C_3\}$, and $\{C_2, C_3\}$ contains at least one reduct. Moreover, either $C_2$ or $C_3$ is indispensable to $\{C_2, C_3\}$ because each of them offers some demarcations that cannot offer by the other. For example, if we retain only $C_3$ in the table, objects 1, 2 and object 5 will be inconsistent because they satisfy the same condition but have different decisions. Hence, $\{C_2, C_3\}$ is a reduct of this decision table.

If an element in a demarcation relation induced by some condition attribute is also the element in the demarcation relation induced by decision attributes *D*, the element is called one ***relative demarcation***. For example, (1, 3) is a relative demarcation because it not only can be induced (offered) by attribute $C_3$, But also can be induced by decision attribute *D*.

The algorithm is based on a simple thought. That is, for the attribute set $\{C_1, C_2, \cdots, C_n\}$, $C_i$ is dispensable to the other attributes and can be removed without loss of accuracy of classification, if and only if the relative demarcations that can be offered by attribute $C_i$ with respect to decision attributes *D* is the subset of that can be offered by the other attributes. The proof can be found in Appendix A, which is based on a more general model.

Before introducing my algorithm, I'll first bring forward some important concepts. One basic concept is "demarcation task". A demarcation task is a subset of demarcation relation induced by decision attributes. Another basic concept is "absolute cascade task at attribute $C_i$" or "absolute task at layer (stage) *i* " or "layered task *i* ", which is the intersection of equivalence relation induced by $C_i$ and demarcation relation induced by decision attributes. It can be denoted by $alT_i$. For instance, the absolute cascade task at attribute $C_3$ is: {(1, 1), (2, 2), (3, 3), (4, 4), (5, 5), (1, 2), (2, 1), (1, 5), (5, 1), (2, 5), (5, 2), (3, 4), (4, 3)} $\bigcap$ {(1, 3), (1, 4), (1, 5), (2, 3), (2, 4), (2, 5)}={(1, 5), (2, 5)}. We can use *“**segment**”* in representation of task. For example, we can use {[1, 2], [5, 5]} to denote {(1, 5), (2, 5)}. Here, [1, 2] and [5, 5] are two segments. For segment [1, 2], “1” is the first position of this segment, and “2” is the last position of this segment. The objects belonging to this segment have a value “1” under attribute $C_3$. And this segment falls into D-Region “0”, according to the labels of objects in this segment. In this paper, segments will be frequently used, not only for concision and intuition, but for efficiency of computation that will be explained in Appendix B.

[8] A decision table is consistent if for any two objects, they belong to the same decision class (or D-Region) when they take an identical value, if the value is known in both objects, in every conditional attribute.

Now I will introduce the most important concept “relative cascade task at attribute $C_i$ (stage)” (or “relative task at layer $i$ ”, or “layer $i$ task”) or just “cascade task at attribute $C_i$ ” for short.

**Definition 1**: $lT_i$ is a “(relative) cascade task at attribute $C_i$ ” to the attributes set $\left\{C_{j1}, C_{j2}, \cdots C_{jl}\right\}$, if $lT_i = \bigcap_{k\in[1,l]} \left\{alT_{jk}\right\}$.

If we give the attributes an order, then we obtain a recursive definition:

$lT_i = alT_i \bigcap lT_{i+1}$, or $lT_i = alT_i \bigcap lT_{i-1}$.

Which formula should be taken is decided by which order we take. In this paper, $lT_i=alT_i \bigcap lT_{i+1}$ is adopted because we first check if the attributes on right side can offer all the necessary demarcations. This definition is especially useful in incremental algorithm which will be introduced in section 4.1.

For example, in the above small example, we have:

$lT_3 = alT_3 = \left\{[1,2],[5,5]\right\}$; $lT_2 = alT_2 \bigcap lT_3 = \phi$.

Although it is “clear” that $C_2$ and $C_3$ are indispensable for {$C_2$, $C_3$}, when there are many attributes, it is hard to “see” which one is dispensable and which one is indispensable. How can we find a reduct in the reduced attribute set?

Fortunately, we can see an interesting thing in the reduced attribute set. That is any attribute in it is indispensable to attributes in its right side. For example, in the reduced set {$C_2$, $C_3$}, $C_2$ is indispensable to $C_3$ because it is introduced to offer some demarcations that cannot be offered by $C_3$.

If we reverse the order of attributes in {$C_2$, $C_3$} and compute cascade tasks again, the results are: $lT_2^{'} = alT_2^{'} = \left\{[1,2],[3,3]\right\}$; $lT_3^{'} = alT^{'}_3 \bigcap lT_2^{'} = \phi$.

As have claimed before, if an attribute cannot offer any demarcation that cannot be offered by other attributes (that is the attributes in its right side and left side in the reduced attribute set), the attribute can be removed, or else it is indispensable and should be retained. We can know it by compare old cascade task and new cascade task to check if there exists some demarcation task that appears in both cascade tasks.

## 2. 2 Algorithm TWI-SQUEEZE

As have introduced in the beginning of this section, we are performing reduction on a decision table $KS=(U, C\bigcup D)$. There are $t_C$ attributes in condition attribute set $C$. In the algorithm to be proposed, (relative) cascade task $lT_i$ stores the demarcation tasks needed to be done by $\{C_1, C_2,\cdots, C_{i-1}\}$, since these tasks cannot be done by $\{C_i, C_{i+1},\cdots, C_p\}$ ( $p= t_C$). If it won't cause misunderstanding, use $lT$ instead of $lT_i$. Here $lT$ means the cascade task needed to be done at present. As have shown in Figure 1, $lT$ is composed of some queues. The elements of queues are segments. Each segment is composed of first position of segment, last position of segment, and a label which denotes the D-Region it belongs to. Each queue has a sole value, which denotes attribute value of all the objects belonging to this queue (and the segments in it). And each queue can be divided according to D-Regions. All the segments in D-Region "*i*" can be called D-R-*i*.

We need a data structure which can not only describe $lT_i$, but also the difference between two successive cascade tasks. The margin stores the demarcation tasks done by attribute $C_i$ and can be called "semi-tasks". We will know the reason it is so named in section 3 and 4. The data structure is called "cascade successive task" $lsT_i$, as shown in Figure 1. All the queues are indexed and linked[9]. And each one has a unit which records how many D-Rs the queue has. For example, the first queue in Figure 1 has 5 D-Rs correspond to value "*a*" under attribute $C_i$. It means objects that fall into different D-Rs cannot be discerned by $C_i$, therefore forming the demarcation tasks need to be done further.

In Figure 1, all the queues in $lsT_i$ are grouped into two classes. Queue of one class has more than one D-Regions. This class is the "cascade task" class $lT_i$; Queue of the other class has only one D-Regions. This class is the "semi-tasks" or "potential tasks" $sT_i$, which will be used in incremental algorithm and even parallel computing later. $sT_i$ has no use if we do not concern incremental problem. We can also do not distinguish these queues, which is a good choice when incremental algorithm is concerned. However, the current scheme is more suitable to illuminate my thoughts. Therefore, I'll use this scheme to describe the data structure.

An array called "Ad" is also needed in this algorithm. A number indicates the new relative address of an object in table after sorting.

In this paper, an algorithm called TWI-SQUEEZE is brought forward. *The algorithm needs only two "squeeze" processes*. Each squeeze removes "water", or redundant attributes, from the original system and both of these two squeezes gains their ends when $lT=\varnothing$. It is known that if we want to squeeze water from sponge, we should exert pressure on both sides of the sponge. The thought of algorithm TWI-SQUEEZE is just like that. And it is interesting that after two squeezes, no "water" is retained in the "sponge".

In the beginning of the algorithm, objects will be sorted by decision attribute set $D$. Assume the result is: $SEG_D=\{Seg_1, Seg_2, \cdots, Seg_p\}$ ($p=r_D$). These $r_D$ segments correspond to $r_D$ D-Regions. These segments will be sorted according certain condition attribute, forming some subsegments. The subsegments that have a same value indicate the objects that cannot be discerned by this condition attribute.

[9] If all values of attributes are known, we can also use other more efficient schemes.

Below is the outline of the algorithm:

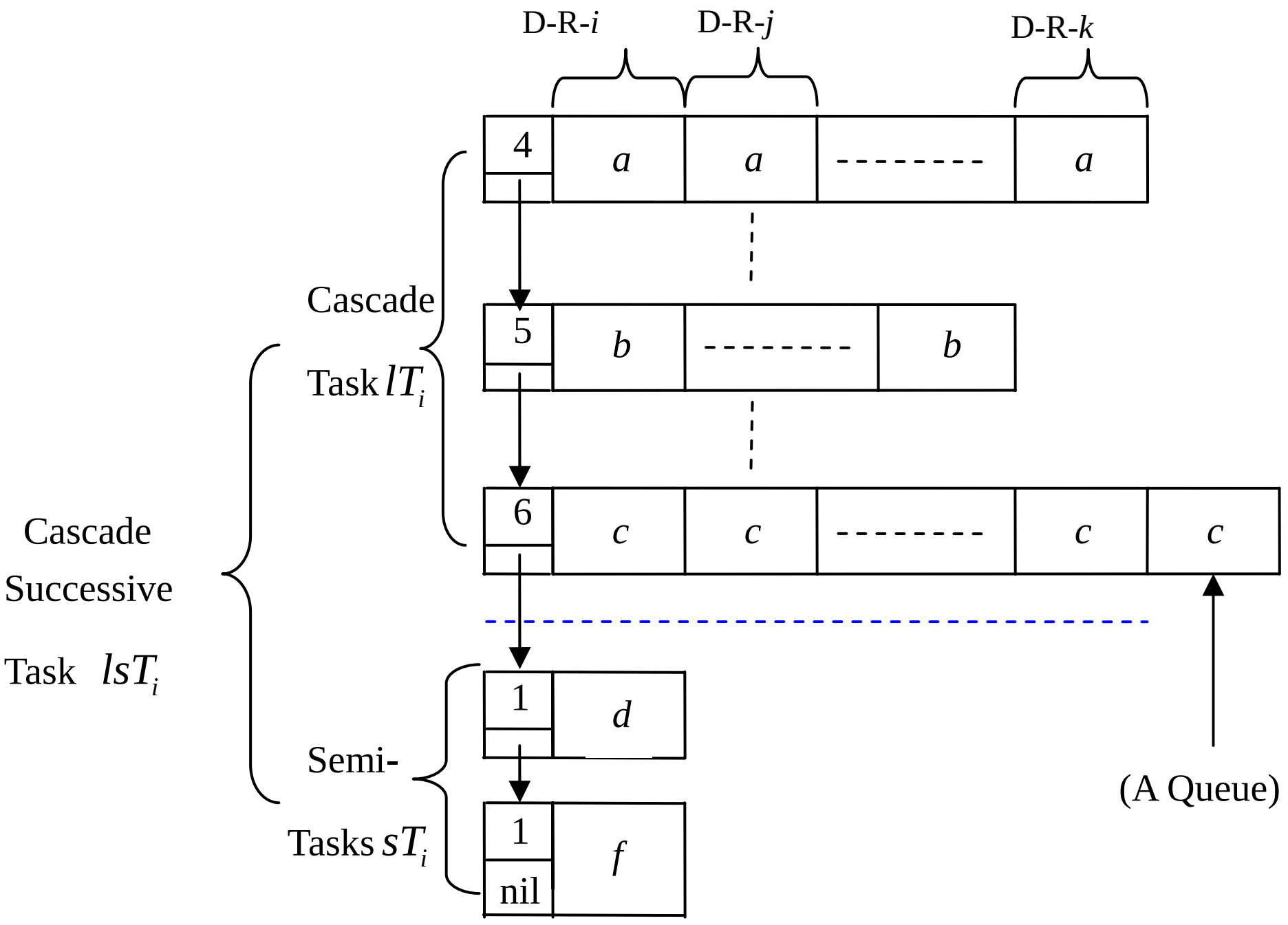

Figure 1: Data structure of a Cascade Successive Task

Algorithm R2L-SQUEEZE ($KS, C'.s_reduct$)

/* $KS'=(U, C'\cup D)$ *is a projection of* $KS$ *by* $C'$. *And sorting is done on* $KS'$. *This subroutine finds a s_reduct.*[10] *Assume* $Card(C')=r$. */

(1) */* Forming the original cascade task* $lT_r$ */

FOR $i=1$ TO $r$ DO

Sort $Seg_i$ under $C_r'$. Insert the subsegments of $Seg_i$ into $lT_r$.

(2) $k \leftarrow r$.

WHILE $lT_k \neq \phi$ DO

( IF $k=0$ THEN separate inconsistent elements from original table and renew cascade tasks, and RETURN. */*There still exists some queue of* $lT_0$, *in which there are more than one D-Rs. Elements in these D-Rs are inconsistent.*/

Sort the elements of $lT_k$ under $C_k'$; check&remove[11]; and form $lsT_{k-1}$ in the same time.

---

[10] A super reduct, or $s_reduct$, is a set of attributes which contain at least one reduct.

[11] In the process, we need check whether some queues should be put into $sT_{i-1}$ or not. If there exist such queues, move these queues from $lT_i$ to $sT_{i-1}$, (and in the same time form $lT_{i-1}$. Let $e(lT_i)$ be the set of objects felled into

IF $lT_{k-1} = lT_k$ THEN skip *\/*It means no D-R of $lT_k$ can be further demarcated under attribute $C'_{k-1}$ */*

ELSE $s_reduct \leftarrow s_reduct + \{C_k'\}$ *\/*It means that there exists some demarcation task which can be done by $C'_{k-1}$. */*

$k \leftarrow k-1$.

) ■ */*All tasks produced in the subroutine are kept.*/*

Algorithm L2R-SQUEEZE ( $KS$ , $C'$ . $reduct$ )

/* $KS' = (U, C' \cup D)$ *is a projection of* $KS$ *by* $C'$. *This subroutine finds a reduct.* */

This subroutine is similar to R2L-SQUEEZE except that at the same time it examines whether $lT_i'$ has some parts overlap with its counterpart. If there are some objects that cannot be discerned by all the attributes in $s_reduct$ except $C_k$, $C_k$ should be included in *reduct*, else it should not. Appendix B gives the necessary details that explain the work. ■

Algorithm TWI-SQUEEZE ( $KS$ . $reduct$ ) */* The MAIN PART*/*

/* *find a reduct in a decision table described in the paper.**/

(1) Give the conditional attributes an order.

*/*From the left end to the right end:* $\{C_1, C_2, \cdots C_{t_C} \mid D\}$*/

(2) Sort the objects by decision attributes $D$

*/*Assume the result is:* $SEG_D = \{Seg_1, Seg_2, \cdots Seg_{r_D}\}$*/

(3) $s_reduct \leftarrow$ R2L-SQUEEZE ( $KS, C$ ). */* Find an attributes subset s_reduct, which contains a reducet. s_reduct is a queue.*/*

(4) $reduct \leftarrow$ L2R-SQUEEZE ( $KS, rs_reduct$ ) */** $rs_reduct$ *is the reverse of* $s_reduct$**/*

■

Squeezing a $s_reduct$ is crucial in the algorithm. A small example can illustrate it as shown in table 2: condition attributes $C_1,\ldots,C_6$ are arranged from the left to right. *D* is the decision attribute. $t_C = 6$, $t_D = 1$. First, the decision attribute is sorted, forming three segments labeled with 0, 1 and 2. Therefore, the root demarcation task is formed: $T_D = \{[1,6]_0, [7,14]_1, [15,18]_2\}$. $r_D = 3$, which means there are three *D-Regions* in this table: $[1,6]_0, [7,14]_1$, and $[15,18]_2$. And the subscript denotes which D-Region a segment is belonged to. Then we sort these three segments under $C_6$ respectively. It is clear that objects which have a same value but belong to different D-Regions cannot be discerned, an $lsT_6$ is formed: $lsT_6 = \{lT_6 : \{T_{60} : [1,4]_0, [7,8]_1, [15,18]_2; T_{61} : [5,6]_0, [9,12]_1\}; sT_6 : \{[13,14]_1\}\}$. [12] There are two

---

$lT_i$, then $sT_{i-1} = e(lT_i) - e(lT_{i-1})$. $sT_{i-1}$ will be used in increment algorithm). This process is called as “check& remove”.

[12] It is a simplified representation of $lsT_6 = \{lT_6, sT_6\}; lT_6 = \{T_{60}, T_{61}\}; T_{60} = \{[1,4]_0, [7,8]_1, [15,18]_2\}; T_{61} = \{[5,6]_0, [9,12]_1\}$;

subtasks in $lT_6$. That is $T_{60}$ and $T_{61}$, which are the child demarcation tasks of $T_D$. $lT_6$ isn't equal to $T_D$, therefore $C_6$ should be included in $s_reduct$. Afterwards, continue to sort the segments of each subtasks in $lT_6$, forming the next cascade successive task:

| Ad | $C_1$ | $C_2$ | $C_3$ | $C_4$ | $C_5$ | $C_6$ | D |
|---|---|---|---|---|---|---|---|
| | 0 | 0 | 1 | 1 | 0 | 0 | 0 |
| | 1 | 0 | 1 | 1 | 0 | 0 | 0 |
| | 0 | 1 | 0 | 0 | 1 | 0 | 0 |
| | 0 | 1 | 1 | 0 | 1 | 0 | 0 |
| | 0 | 1 | 1 | 1 | 1 | 1 | 0 |
| | 1 | 2 | 0 | 0 | 1 | 0 | 1 |
| | 1 | 1 | 1 | 1 | 1 | 1 | 0 |
| | 0 | 2 | 0 | 0 | 1 | 0 | 1 |
| | 1 | 0 | 0 | 1 | 0 | 1 | 1 |
| | 0 | 1 | 0 | 1 | 0 | 1 | 1 |
| | 1 | 0 | 0 | 1 | 1 | 1 | 1 |
| | 1 | 0 | 0 | 1 | 1 | 1 | 1 |
| | 0 | 1 | 1 | 0 | 0 | 2 | 1 |
| | 1 | 0 | 0 | 0 | 0 | 2 | 1 |
| | 0 | 1 | 0 | 1 | 0 | 0 | 2 |
| | 1 | 1 | 0 | 1 | 0 | 0 | 2 |
| | 0 | 0 | 1 | 0 | 1 | 0 | 2 |
| | 1 | 0 | 1 | 0 | 1 | 0 | 2 |

(a) The original table

<table>
<tr><th>Ad</th><th>C<sub></sub>1</th><th>C2</th><th>C3</th><th>C4</th><th>C5</th><th>C6</th><th>D</th></tr>
<tr><td>1</td><td rowspan="18"></td><td rowspan="2"></td><td rowspan="2">1</td><td rowspan="2">1</td><td rowspan="2">0</td><td rowspan="4">0</td><td rowspan="6">0</td></tr>
<tr><td>2</td></tr>
<tr><td>3</td><td>1</td><td>0</td><td rowspan="2">0</td><td rowspan="2">1</td></tr>
<tr><td>4</td><td>1</td><td>1</td></tr>
<tr><td>5</td><td rowspan="2"></td><td rowspan="2">1</td><td rowspan="2">1</td><td rowspan="2">1</td><td rowspan="2">1</td></tr>
<tr><td>6</td></tr>
<tr><td>7</td><td rowspan="2">2</td><td rowspan="2">0</td><td rowspan="2">0</td><td rowspan="2">1</td><td rowspan="2">0</td><td rowspan="8">1</td></tr>
<tr><td>8</td></tr>
<tr><td>9</td><td rowspan="8"></td><td rowspan="2"></td><td rowspan="2"></td><td rowspan="2">0</td><td rowspan="4">1</td></tr>
<tr><td>10</td></tr>
<tr><td>11</td><td rowspan="2">0</td><td rowspan="2">1</td><td rowspan="2">1</td></tr>
<tr><td>12</td></tr>
<tr><td>13</td><td rowspan="2"></td><td rowspan="2"></td><td rowspan="2"></td><td rowspan="2">2</td></tr>
<tr><td>14</td></tr>
<tr><td>15</td><td rowspan="2">0</td><td rowspan="2">1</td><td rowspan="2">0</td><td rowspan="4">0</td><td rowspan="4">2</td></tr>
<tr><td>16</td></tr>
<tr><td>17</td><td rowspan="2">0</td><td rowspan="2">1</td><td rowspan="2">0</td><td rowspan="2">1</td></tr>
<tr><td>18</td></tr>
</table>

(b)A small example which illustrates the process of R2L-SQUEEZE

Table 2: A small example: Find a $s_reduct$

**Remark 1:** ***After R2L-SQUEEZE is performed, we should use an array "Ad" to record the relative address of all the objects in KS, as have shown in Table 2 (b).***

$lsT_5 = \{lT_5 : \{T_{60,50} : [1,2]_0, [15,16]_2; T_{60,51} : [3,4]_0, [7,8]_1, [17,18]_2; T_{61,51} : [5,6]_0, [11,12]_1\}; sT_5 : \{[9,10]_1\}\}$.

As have explained before, $C_5$ should be included in $s_reduct$. *From now on, I will only concern $lT_i$ and ignore $sT_i$, because $sT_i$ is not used at present.*

Follow on, we have:

$lT_4 = \{T_{60,50,41} : [1,2]_0, [15,16]_2; T_{60,51,40} : [3,4]_0, [7,8]_1, [17,18]_2; T_{61,51,41} : [5,6]_0, [11,12]_1\}$. ( $lT_4 = lT_5$, $C_4$ is redundant for it doesn't contribute to discriminating of different objects. Go on)

---

$sT_6 = \{[13,14]_1\}$. The algorithm forms a tree, the nodes of which are tasks. The root of the tree is $T_D$, and $T_{60}$ is a denotation of a child task, and $lsT_6$ is the denotation of all the child tasks and semi-tasks in level 1 of the tree, as shown in Figure 3.

$lT_3 = \{T_{60,51,40,30} : [3,3]_0, [7,8]_1; T_{60,51,40,31} : [4,4]_0, [17,18]_2\}$. $s_reduct \leftarrow s_reduct + \{C_3\}$.

$lT_2 = \phi$. $lT_2 \neq lT_3$, $\therefore s_reduct \leftarrow s_reduct + \{C_2\}$.

Therefore, at last we get the $s_reduct$: $\{C_2, C_3, C_5, C_6\}$.

From the instance, we can see that only the necessary values are sorted, and the need of sorting decrease gradually. Hence, it can be called "Taper-off-Sorting".

Figure 2 is the task tree formed in the run of algorithm.

It is easy to see from the prior proof that, after the demarcations offered by *s_reduct* are equal to that offered by decision attribute, *s_reduct* contains at least one reduct. Moreover, *s_reduct* has a good quality. That is, every attribute in *s_reduct* is independent to the attribute on its right side, which means that the "squeeze" gives *s_reduct* an order structure. Therefore, if we "squeeze" the *s_reduct* in a reverse order, we will obtain another *s_reduct*, every attribute in which is not only independent to the right side, but also independent to the left side. Furthermore, for any attribute, if the cascade task which is formed in its right side in algorithm R2L-SQUEEZE and the cascade task which is formed in its left side in algorithm L2R-SQUEEZE have some parts overlapped, then the attribute must be in the *reduct*. Otherwise, the attribute can be removed without violating the consistency of classifications.

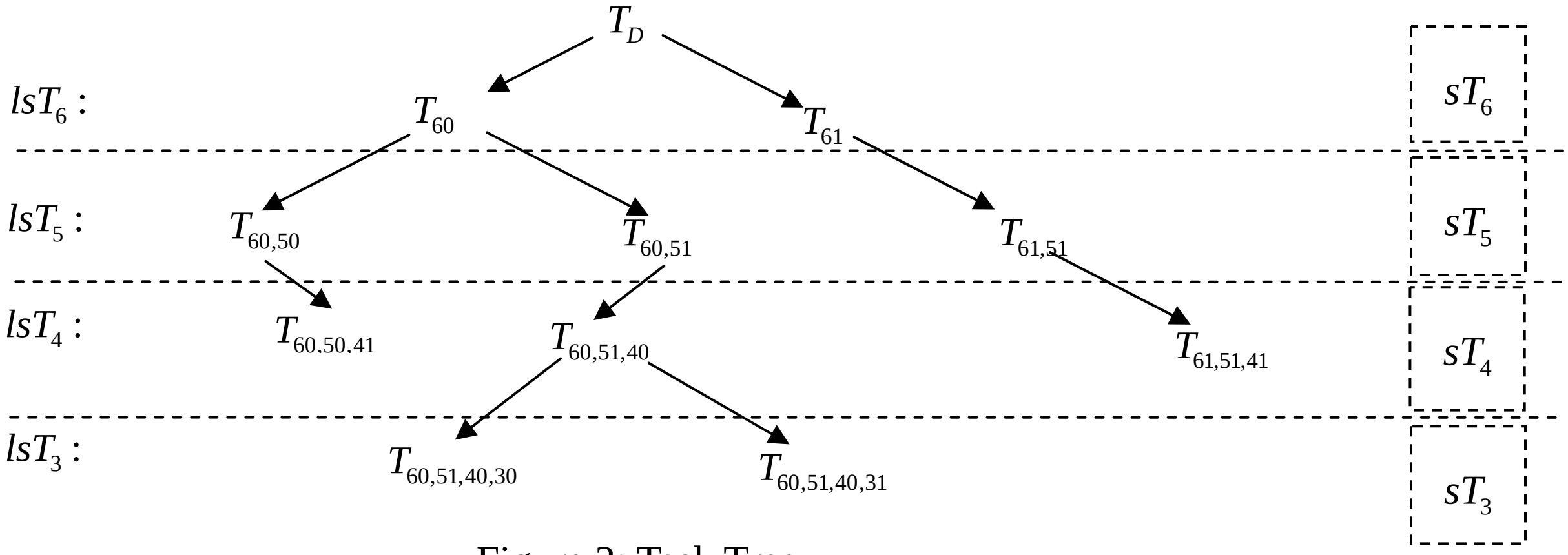

Figure 2: Task Tree

As for that small example, we have: $C_2$ and $C_6$ must be included in *reduct*.

$lT_2' = \{T_{20}' : [1,2]_0, \{[9,9],[11,12],[14,14]\}_1, [17,18]_2; T_{21}' : [3,4,5,6]_0, \{[10,10],[13,13]\}_1, [15,16]_2\}$.

$lT_2' \overset{+}{\cap} lT_5 = \phi$.[13] Therefore, $C_3$ is redundant to $\{C_2, C_5, C_6\}$.

Because $\{1,2\} \cap \{1,2,3,4\} = \{1,2\} \neq \phi$, and in the same time $\{17,18\} \cap \{15,16,17,18\} = \{17,18\} \neq \phi$, which means some part of demarcation task, $\{[1,2]_0, [17,18]_2\}$, is shared by $lT_2'$ and $lT_6$. So is $\{[5,6]_0, [10,10]_1\}$. Therefore, $lT_2' \overset{+}{\cap} lT_6 = \{[1,2]_0, [17,18]_2; [5,6]_0, [10,10]_1\} \neq \phi$. Hence, $C_5$ is independent to $\{C_2, C_6\}$, and it need to be included in *reduct*.

Therefore, *reduct* = $\{C_2, C_5, C_6\}$.

[13] For two "cascade tasks", if they share no part of demarcation task, then they are totally different, and their "task intersection" ($\overset{+}{\cap}$) is void.

Table 3 illustrates the process of L2R-SQUEEZE of Table 2(a). Table 2, Table 3 and their annotations show a complete process of algorithm TWI-SQUEEZE.

At last, I must bring forward two concepts, "cascade task intersection" and "potential cascade task intersection", which are very useful in incremental algorithm.

**Definition 2:** $\{a_1, a_2, \cdots, a_t\}$ is an attribute list, let $IST_{a_i} = lT'_{a_{i-1}} \overset{+}{\bigcap} lT_{a_{i+1}}$. $IST_i$ is called as "cascade task intersection" at layer (stage) $a_i$.

**Definition 3:** Let $PIST_{a_i} = lT'_{a_{i-1}} \bigcap lT_{a_{i+1}} - lT'_{a_{i-1}} \overset{+}{\bigcap} lT_{a_{i+1}}$. $PIST_i$ is called as "potential cascade task intersection" at layer $a_i$.

| Ad | **$C_6$** | **$C_5$** | **$C_3$** | **$C_2$** | D |
|---|---|---|---|---|---|
| 1 | 0 | 0 | 1 | 0 | 0 |
| 2 | 0 | 0 | 1 | 0 | 0 |
| 3 | 0 | 1 | 0 | 1 | 0 |
| 4 | 0 | 1 | 1 | 1 | 0 |
| 5 | 1 | 1 | 1 | 1 | 0 |
| 6 | 1 | 1 | 1 | 1 | 0 |
| 9 | 1 | 0 | 0 | 0 | 1 |
| 11 | 1 | 1 | 0 | 0 | 1 |
| 12 | 1 | 1 | 0 | 0 | 1 |
| 14 | 2 | 0 | 0 | 0 | 1 |
| 10 | 1 | 0 | 0 | 1 | 1 |
| 13 | 2 | 0 | 1 | 1 | 1 |
| 7 | 0 | 1 | 0 | 2 | 1 |
| 8 | 0 | 1 | 0 | 2 | 1 |
| 17 | 0 | 1 | 1 | 0 | 2 |
| 18 | 0 | 1 | 1 | 0 | 2 |
| 15 | 0 | 0 | 0 | 1 | 2 |
| 16 | 0 | 0 | 0 | 1 | 2 |

Table 3: A small example which illustrates the process of L2R-SQUEEZE

## 2. 3 Complexity of Algorithm TWI-SQUEEZE

The main part of time needed in algorithm is sorting[14]. Sorting needs no more than $O((t_C + t_D)N\ln N)$ steps and only one additional space unit, using heap sort. When the number of values is small and most of the objects are arranged in some order, binary sort is recommendable.

[14] We can use an array of pointers to store the address of objects (or recorders). Therefore we need only move pointers instead of recorders.

The other part of time needed in algorithm is what I call "conglutinative match", it need only $O(N)$ steps. Please refer to Appendix B for more details.

Therefore, the worst time complexity of algorithm TWI-SQUEEZE is: $O((t_C+t_D)N\ln N)$ or $O(t_C N\ln N)$, assuming $t_C$ is much larger than $t_D$, which is true in most practical applications. I'll ignore $t_D$ in later analysis as the other works do.

Generally speaking, the length of $rs_reduct$ is smaller than $t_C$. Assume $length(rs_reduct)=r\le t_C$, the worst time complexity of algorithm TWI-SQUEEZE is: $O(rN\ln N)$.

In the first squeeze, we can keep all demarcation tasks $lT_i$ $(C_i \in s_reduct)$ in memory, taking a memory-resident view. In the second squeeze, only one new demarcation task is needed. All these queues occupy $O(rN)$ units. These are additional space. In the same time, we need only put a small fraction of the total table, projection of one or more attributes of the table, into the memory. That is only $O(N)$ units are needed. Therefore the total space needed is $O(rN)$.[15] However, we can also do not maintain all demarcation tasks $lT_i$ in the memory, but write these tasks to disk. From this disk-resident view, the algorithm needs only $O(N)$ memory space, with the price of some read in and write out disk operations. Note that the price is acceptable because it needn't frequently run such disk operations, which can be ignored in analysis.

Which factors will influence the complexity? Either $r_D$ or $\overline{r_C}$ plays an important role in the complexity.[16] When they are large, the complexity is normally also large because they influence the complexity of finding intersection of two demarcation task. However, $r_D$ and $\overline{r_C}$ restrict each other because $r_D \cdot \overline{r_C} \le N$. Since $r_D$ and $\overline{r_C}$ have influence to the algorithm, it is necessary to combine the attribute-oriented induction method (Cai, Cercone, Han, 1991) or value discretization (there are a lot of works. See for example the tutorial written by (Komorowski, Pawlak, Polkowski, &Skowron, 1998, to recognize the works in rough set literature) in some case.

The initial order of condition attributes is also crucial for this algorithm. The complexity will increase if the size of demarcation task always decreases a very small bit. One example can cast some light on it. If the number of contiguous attributes which are similar in nature in classification ability is large, the algorithm is less efficient.

Let $\mu = lT_{i-1}/lT_i$ $(0 \le \mu \le 1)$ in procedure R2L-SQUEEZE, then the worst case time

[15] Usually it is O(N), except some unusual case.

[16] $\overline{r_C}$ represents the average number of different values of conditional attributes in a D-Region.

complexity of the algorithm is: $O\left( Min(t_C, \log_\mu \frac{2}{N}) \times N \ln N \right)$. We can analyze it more thoroughly. For example, if $\mu$ is not larger than 0.8 in most case (for example, 75%), which I think is not an uncommon case, the time complexity is:

$O\left(1.33 \times 5N \cdot \left(1 - 0.8^{5 \ln N}\right) \cdot \ln N\right) = O\left(N \ln N\right)$, and the space complexity is:

$O\left(1.33 \times 5N \cdot \left(1 - 0.8^{5 \ln N}\right)\right) = O\left(N\right)$, which implies that generally the algorithm works much better than its worst case.

### 2.4 Relative CORE (Attribute) and (Absolute) Irrelevant Attributes

The relative core (attribute) of $KS=(U, C \bigcup D)$ can be defined as: $CORE_D(C) = \bigcap reduct_D(C)$. Relative CORE is meaningful and can be exploited in knowledge discovery. In fact, it is just the attributes "strongly relevant (to the sample)" and the rest attributes in reduct is the attributes "weakly relevant (to the sample)". The reduct itself reveals the complexity of a function, and Algorithm TWI-SQUEEZE reveals the incremental usefulness, the fifth definition of relevance, according to Blum and Langley (1997). Relationship of reduct and relevance is also discussed by Komorowski, Pawlak, Polkowski, and Skowron (1998).

It is clear that when the parameter $rs_reduct$ of L2R-SQUEEZE is replaced by *C*, we get the core instead of a reduct. The time complexity is $O(t_C N \ln N)$, and space complexity is still $O(N)$ .

Up to now, we are talking about how to find relevant attributes (features). Can we find (absolute) irrelevant attributes, which do not appear in any reduct, efficiently? The answer is yes. And the solution is simple. We need only see whether an attribute cannot offer relative demarcations, it is an irrelevant attribute. The time complexity for finding all irrelevant attributes is $O(t_C N \ln N)$, and space complexity is $O(N)$.

It has been regarded as a problem in feature selection literature that many of the widely-used UCI data sets have few completely irrelevant attributes (Langley, Sage, 1994; Blum, Langley, 1997). But it not so serious a problem for algorithm TWI-SQUEEZE, because after an efficient pretreatment, there is no absolute irrelevant attribute left.

## 3. Parallel: Separate the Table

I've mentioned in complexity analysis that we can separate the table into columns arbitrarily to save space. In fact, Algorithm TWI-SQUEEZE has another good quality: we can divide the table easily and run the algorithm nearly in complete parallel.

How can we separate the table? One simple method is to divide the initial task into some queue sets, and assign each queue set to a distinct computer.

Can we separate the table more freely? The other thought is that instead of assigning queues, we assign subtasks which have the same or similar structure as the original cascade task, as possible as it could be.[17] This is realized by dividing D-Rs. Every D-R is divided to keep the structure, as shown in Figure 3. Different from Figure 1, to make it visual for understanding, I use length of line to denote the size of a D-R.

[17] Two tasks have same structure, if they have same queues and the corresponding queues have same nonempty

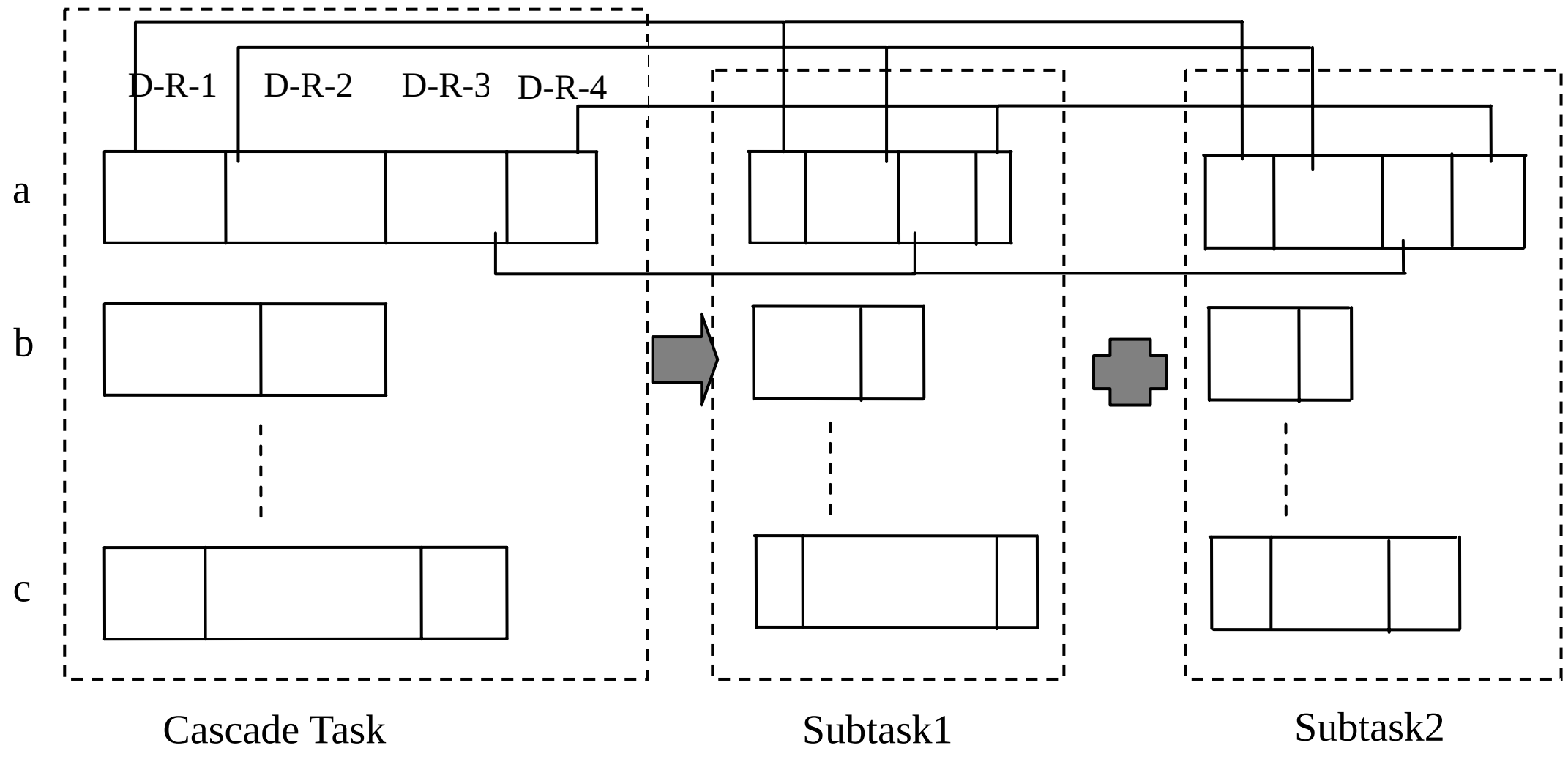

Figure 3: Separation of Cascade (Layered) Task

If a cascade task is linear parallel separable[18], we can run the algorithm in complete parallel. However, we must be cautious that a cascade task is not complete linear parallel separable, which is caused by the process check&remove. It can be nevertheless called as semilinear parallel separable. What can we do then? Parallel computing is just like assigning many people to paint a wall (Hillis, 1999). If the wall is divided into several regions and every one answers for one distinct region, then the task can be done in complete parallel. However, if the assignment is imperfect, although everyone takes charge of his distinct region, there will be some small areas that are left being painted. To solve this problem, a simple method is to assign another people to paint these left areas when they are found. Therefore, the painting task can be done in nearly complete parallel.

First of all, where are these "areas"? In our problem, after check&remove, some segments are removed from task $lT_i$ and stored in $sT_{i-1} = e(lT_i) - e(lT_{i-1})$. In fact, the thread to find "lost area" (task which has not been dealed with) hides just in the $sT_{i-1}$. We check if there are some semi-tasks in one computer which will transform to task when some semi-tasks in other computers are taken into account.

Therefore, we can separate the table arbitrarily as long as all the subtasks have the same or similar structure as the original one. Assign these subtasks to different computers, and computing the problem in complete parallel. In the same time, a special computer, act as the special painter who paints the ignored areas, receives the information (tasks and semi-tasks) from the other computer, integrate them and carries its duty. It works a bit slower than the other computers because it has to wait to receive and ingrate information frequently. Another

D-Regions.

[18] If a task can be divided into some subtasks, which can be finished in complete parallel without increase of time-space complexity, the task is called "linear parallel separable".

strategy it can adopt is to assign new computers to do the ignored tasks. Therefore, it plays an important role in the analysis of algorithm.

Although we can separate the tasks freely, a good assignment rule or heuristic policy must exist.

In the demarcation, we can also do not obey the requirement that subtasks should have the same or similar structure as the original one. We can separate one queue based on D-Region. For example, for the table in Table 2, the demarcation task can be divided into three subtasks: [1, 6], [7, 14]; [1, 6], [15, 18]; [7, 14], [15, 18]. It can also be divided into two subtasks: [1, 6], [7, 18]; [7, 14], [15, 18]. However, this kind of demarcation is not semilinear parallel separable. It can also be effective and efficient if the number of D-Region is small or there are many available computers[19].

I've introduced three methods which can realize parallel computing. There exists another method which can let us separate a table most freely. We need only divide the objects of table into some classes, without concern task trees of the table. Of course, the price of complete freedom is increase of complexity. I'll introduce the thoughts in section 4.1. Although in incremental algorithm, data are divided into two classes: old and new data, and we handle them in different time, we can also deal them in the same time for parallel computing, using the same ideas.

## 4. Adapt to the Changing World

The world is changing every moment. Therefore, an algorithm is effective only when it can match the changing world. As for our problem, the changes come from two sources. One is the change of objects. The other is change of attribute set.

### 4.1 Increment Version of Algorithm TWI-SQUEEZE

A good algorithm needs an efficient incremental version to make it better. Any increment version of attribute reduction algorithm must take the original table into account. So is mine. However, it uses a small part of old information instead of the whole table, and therefore the incremental version is very efficient.

As has pointed out in (Cheng, &Ye, 2002), there are two cases need to be concerned: 1.when new attributes added to reduct, it is possible that some old attributes become useless and therefore need to be removed from reduct; 2. when new objects are added, they may possibly make some old objects from discernible to indiscernible. And the attributes that are used to discern these old ones will also possibly become useless.

The problem seems difficult to solve efficiently. However, it is very simple if we observe it in a different viewpoint. In my algorithm TWI-SQUEEZE, $lT_k \ (k \in [1 \cdots t_C])$ can be divided into two parts. Each one corresponds to one of the two parts of the table. If we regard one of them as new data, the increment algorithm is at hand. Because different from all previous works, algorithm TWI-SQUEEZE is not based on compare of objects or segments, which makes it semilinear parallel separable.

[19] If one queue is separated into d parts based on D-Region, then at most 0.5d(d-1) computers are needed to compute these subtasks. Each computer is assigned a pair.

One simple idea is that new data correspond to a new task tree. In the process of sorting new data, when a task (or segment) is formed, insert it into appropriate position in old task tree. In the same time, we examine whether the semi-tasks of both new and old can be converted into new tasks. In this way we can merge new task tree and old task tree by inserting new task tree into old task tree gradually.

After algorithm R2L-SQUEEZE is performed, we obtain a united task tree. We can also run algorithm L2R-SQUEEZE to obtain another and compute cascade task intersection of these two tree layer by layer as before. However, a good incremental algorithm should fully exploit the old information obtained before as possible as it can. Can we expect a better way? It does exist. The crucial point is the recursive definition of cascade task (section 2.1, definition 1). And a simple idea is supported by a proposition:

**Proposition 1:** $\{C_k\}$ is a set of new attributes to the original reduct. In algorithm L2R-SQUEEZE, if the **intersection** of new "cascade task" induced by $\{C_k\}$ and "cascade task intersection" at layer *j* **is void** ($C_j$ belongs to the original reduct)**,** we can remove $C_j$ without losing necessary information when $\{C_k\}$ is added to reduct. In other words, $\{C_k\}$ can substitutes $C_j$.

**Proof:** Cascade task intersection at layer j contains all the demarcations that can only offered by $C_j$ in the original reduct. Let ($x$, $y$) be any demarcation that can be offered by $C_j$ but cannot be offered by other attributes in the original reduct. Therefore, ($x$, $y$) will be in $IST_j$. But $lT_k \cap IST_j = \phi$. Hence, ($x$, $y$) is not in $lT_k$ ($lT_k$ is induced by $\{C_k\}$), which means $\{C_k\}$ can offer the demarcation ($x$, $y$). Therefore, $\{C_k\}$ can substitutes $C_j$. ■

Below is the outline of the thought.

Input: New data $KS' = (U', C \cup D)$, $Oreduct, OlT_i, OsT_i, OIST_i, OPIST$ $(C_i \in reduct)$ and the original table.[20]

Output: $reduct$.

Step1. Rearrange *C* to make *Oreduct* in the rightmost side of the attribute list *C*.

Step2. Run algorithm R2L -SQUEEZE on *KS'* and the old cascade successive tasks and form a new united tree in the same time.[21] Now, we obtain a *s_reduct* that contains a *reduct*. If some new objects make some old objects indiscernible then cut out the inconsistent parts, and renew the task tree, and $Oreduct$ if necessary.

If $Oreduct \supseteq s_reduct$ then $reduct \leftarrow Oreduct$; output the *reduct* and return.

Else reverse $s_reduct$, and we obtain $rs_reduct$.

Step3. Run algorithm L2R-SQUEEZE ($KS', rs_reduct \cap oreduct$). In the process, some of the data in $KS$ is sorted and a new united task tree and united $IST_i, PIST_i \ (C_i \in reduct)$ is formed.

[20] To avoid confusion, I put O- as a prefix to the tasks produced before, which means the original one or old one.
[21] Some fractions of $KS$ that correspond to some $OsT_i$ will possibly be sorted in this process.

Step4. Let $\Delta red = s_reduct - Oreduct$. $reduct \leftarrow s_reduct$.

Run algorithm L2R-SQUEEZE $(KS \cup KS', \Delta red)$ to remove redundant attributes in $\Delta red$. Let $lT$ be the current cascade task. Then if $lT \overset{+}{\cap} IST_i = \phi$ then $reduct \leftarrow reduct - \{C_i\}$. At last we obtain a real $reduct$. In the process, we need renew task tree and $IST_i$, $PIST_i$ for future usage.

Here is an example.

| Ad | $C_3$ | $C_4$ | $C_1$ | $C_2$ | $C_5$ | $C_6$ | D |
|---|---|---|---|---|---|---|---|
| 19 | 0 | 1 | 0 | 0 | 0 | 0 | 0 |
| 20 | 0 | 1 | 1 | 0 | 0 | 0 | 0 |
| 21 | 1 | 0 | 0 | 0 | 1 | 1 | 0 |
| 22 | 0 | 1 | 1 | 1 | 1 | 0 | 1 |
| 23 | 1 | 1 | 1 | 1 | 1 | 0 | 1 |
| 24 | 1 | 0 | 1 | 0 | 1 | 1 | 1 |
| 25 | 0 | 0 | 1 | 0 | 1 | 1 | 1 |
| 26 | 0 | 0 | 0 | 0 | 0 | 1 | 2 |
| 27 | 0 | 0 | 0 | 0 | 0 | 1 | 2 |

Table 4: new data (step2, sorted)

The table shown in Table 4 is all the new added data at present, assuming Table 2(a) is the original table which has been dealed with. It is the result of running algorithm R2L-SQUEEZE. Readers need not concern why $C_3$ and $C_4$ are put on the rightmost side at present and can just skip to the next paragraph. $C_3$ and $C_4$ are tackled in this way because they are dependent to $\{C_2, C_5, C_6\}$, which means their contribution in taking on known demarcation tasks are relative small. Therefore I put them on the rightmost side. This policy can be adopted when the initial table is much larger than the incremental data.

$\{[19,20]_0, [22,23]_1\}, \{[21,21]_0, [24,25]_1, [26,27]_2\}$ are tasks at layer 6 of the new data, therefore, they should be merged into $lsT_6$:

$$lsT_6 = \{lT_6 : \{T_{60} : \{[1,4], [19,20]\}_0, \{[7,8], [22,23]\}_1, \{[15,18]\}_2; T_{61} : \{[5,6], [21,21]\}_0, \{[9,12], [24,25]\}_1, \{[26,27]\}_2\}; sT_6 : \{[13,14]\}\}.$$

Although $[19,20]_0, [22,23]_1, [26,27]_2$ are semi-tasks at layer 5 of the new data, they have counterparts in the old table. In the same time, we should also examine if the semi-tasks at layer 5 of the old table have counterparts in the new data. In this case, we find that the counterpart of $[9,10]_1$ is $[26,27]_2$. Therefore they should be merged into $lsT_5$:

$$lsT_5 = \{lT_5 : \{T_{60,50} : \{[1,2], [19,20]\}_0, \{[15,16]\}_2; T_{60,51} : \{[3,4]\}_0, \{[7,8], [22,23]\}_1, \{[17,18]\}_2;$$

$$T_{61,50} : \{[9,10]\}_1, \{[26,27]\}_2; T_{61,51} : \{[5,6], [21,21]\}_0, \{[11,12], [24,25]\}_1\}; sT_5 : \phi\}$$

In $lsT_5$, the new task segments which are absent in layer 5 task of old table, are $[19,20]_0, [22,23]_1, [26,27]_2, [21,21]_0, [24,25]_1$. We need examine if they have counterparts in the semi-tasks at layer 2 of the old table. Because in the old table, $OsT_2 = OlT_5 = \{OT_{60,50}:[1,2]_0,[15,16]_2;$ $OT_{60,51}:[3,4],[7,8],[17,18];OT_{61,51}:[5,6],[11,12]\}$, we have: $lT_2 = \{T_{60,51,21}:[3,4]_0, [22,23]_1; T_{60,51,20}:[9,9]_1, [26,27]_2;$ $T_{61,51,20}:[21,21]_0,\{[11,12],[24,25]\}_1$. And $sT_2 = \{\{[1,2],[5,6],[19,20]\}_0, \{[7,8],[10,10]\}_1, \{[15,16],[17,18]\}_2\}$.

We continue to compute $lT_1$, and find that $lT_1 = \phi$. Therefore, after step 2 is performed, we obtain a $s_reduct:\{C_1,C_2,C_5,C_6\}$. And $\Delta red = s_reduct - Oreduct = \{C_1\}$.

| Ad | $C_3$ | $C_4$ | $C_6$ | $C_5$ | $C_2$ | D | $C_1$ |
|---|---|---|---|---|---|---|---|
| 19 | 0 | 1 | 0 | 0 | 0 | 0 | |
| 20 | 0 | 1 | 0 | 0 | 0 | 0 | |
| 21 | 1 | 0 | 1 | 1 | 0 | 0 | |
| 22 | 0 | 1 | 0 | 1 | 1 | 1 | |
| 23 | 1 | 1 | 0 | 1 | 1 | 1 | |
| 26 | 0 | 0 | 1 | 0 | 0 | 2 | |
| 27 | 0 | 0 | 1 | 0 | 0 | 2 | |
| 24 | 1 | 0 | 1 | 1 | 0 | 1 | |
| 25 | 0 | 0 | 1 | 1 | 0 | 1 | |

Table 5: new data (step3, sorted)

| Ad | $C_6$ | $C_5$ | $C_2$ | D |
|---|---|---|---|---|
| 1 | 0 | 0 | 0 | 0 |
| 2 | 0 | 0 | 0 | 0 |
| 3 | 0 | 1 | 1 | 0 |
| 4 | 0 | 1 | 1 | 0 |
| 5 | 1 | 1 | 1 | 0 |
| 6 | 1 | 1 | 1 | 0 |
| 9 | 1 | 0 | 0 | 1 |
| 14 | 2 | 0 | 0 | 1 |
| 11 | 1 | 1 | 0 | 1 |
| 12 | 1 | 1 | 0 | 1 |
| 10 | 1 | 0 | 1 | 1 |
| 13 | 2 | 0 | 1 | 1 |
| 7 | 0 | 1 | 2 | 1 |
| 8 | 0 | 1 | 2 | 1 |
| 17 | 0 | 1 | 0 | 2 |
| 18 | 0 | 1 | 0 | 2 |
| 15 | 0 | 0 | 1 | 2 |
| 16 | 0 | 0 | 1 | 2 |

Table 6 : old data (part)

$lT_2^{'} = \{T_{20}^{'}:\{[1,2],[19,21]\}_0, \{[9,9],[11,12],[14,14],[24,25]\}_1, \{[17,18],[26,27]\}_2; T_{21}^{'}:[3,6]_0, \{[10,10],[13,13],[22,23]\}_1, [15,16]_2\}$.

$lT_5^{'} = \{T_{20,50}^{'}:\{[1,2],[19,20]\}_0, \{[9,9],[14,14]\}_1, [26,27]_2; T_{20,51}^{'}:[21,21]_0, \{[11,12],[24,25]\}_1, [17,18]_2; T_{21,50}^{'}:[10,13]_1, [15,16]_2; T_{21,51}^{'}:[3,6]_0, [22,23]_1\}$.

$lT_6^{'} = \{T_{20,50,61}^{'}:[9,9]_1, [26,27]_2; T_{20,51,61}^{'}:[21,21]_0, \{[11,12],[24,25]\}_1; T_{21,51,60}^{'}:[3,4]_0, [22,23]_1\}$.

When the number of objects which are contained in a segment is 1, we can use the labels of objects to replace a segment. For example, $[9, 9]_1$ can be replaced by "$\{9\}_1$".

$alT_1 = \{T_{10} : \{1,[3,5],\mathit{19,21}\}_0, \{8,10,13\}_1, \{15,17,\mathit{[26,27]}\}_2; T_{11} : \{2,6,\mathit{20}\}_0, \{7,9,[11,12],14,\mathit{[22,25]}\}_1, \{16,18\}_2\}.$

$C_1$ must be included in the reduct. $lT$=$alT_1$.

$lT \overset{+}{\cap} IST_2 = lT \overset{+}{\cap} lT_5 = \{task1:\{1,19\}_0,\{15\}_2; task2:\{2,20\}_0,\{16\}_2; task3:\{3,4\}_0,\{8\}_1,\{17\}_2; task4:\{10\}_1,[26,27]_2; task5:\{6\}_0,$

$\{11,12,25\}_1; task6:\{7,22,23\}_1,\{18\}_2\} \neq \phi.$ Therefore, $C_2$ should be included in the reduct.

$IST_5 = lT_2^{'} \overset{+}{\cap} lT_6 = \{task1:\{19,20\}_0,[17,18]_2; task2:\{21\}_0,\{9,[11,12],[24,25]\}_1, \mathit{[26,27]}_2; task3:[3,4]_0,[22,23]_1,[15,16]_2; task4:[5,6]_0,\{10\}_1\}.$

[22] $lT \overset{+}{\cap} IST_5 \neq \phi.$ Therefore, $C_5$ should be included in the reduct.

$IST_6 = lT_5^{'} \overset{+}{\cap} T_D = lT_5^{'}.$ $lT \overset{+}{\cap} IST_6 \neq \phi.$ Therefore, $C_6$ should be included in the reduct.

Consequently, the reduct is $\{C_1, C_2, C_5, C_6\}$.

Now let's analyze the algorithm. In the algorithm, we should renew two task trees, and we should also renew $IST_i$ and $PIST_i$. What price will it cost?

To renew two task trees, we need scan the old task trees at most one time. A cascade successive task is composed of cascade task (or layered task) and semi-tasks. They correspond to the values under some attribute. Assuming the average number of values of an attribute is $\bar{r}_C$, the average number of D-Regions of a queue (task or semi-task) is $\bar{r}_D$, and the length of $s_reduct$ is $\gamma$, the complexity of scanning a tree is $\gamma \bar{r}_C \bar{r}_D \left(\gamma \bar{r}_C \bar{r}_D < t_C N\right)$. In Step7, after running L2R-SQUEEZE ($KS \cup KS^{'}, \Delta$), we use $lT$ to renew the tree which is formed in step4. That is replacing $lT_i^{'}$ with $lT \overset{+}{\cap} lT_i^{'}$. Compute intersection is very efficient because both of $lT$ and $lT_i^{'}$ have an ascendant order in both tasks and D-Region.

Another price is that some segments in $S_i$ now should be sorted. Let $\varphi(lT)$ be the total data sorted before, then the time complexity of sorting the segments in $sT_i$ is less than $\gamma N - \varphi(lT)$.

There are two ways to renew $IST_i$ and $PIST_i$. One thought is very simple: after task trees are renewed, we compute their intersection as before. Assuming $|KS| = N, |KS^{'}| = m,$ it needs $O(N+m)$ steps.

The other idea is a bit complex. It is based on the fact that only the new part of the renewed task trees and $lT$ will influence $IST_i$ and $PIST_i$. The former will add some new elements to $IST_i$ and $PIST_i$, and the latter will reduce some fractions of old elements in $IST_i$ and $PIST_i$. This idea is expected to more efficient.

---

[22] Note that here the demarcations contained in $IST_5$ are more than that can be offered by attribute $C_5$.

Therefore, assuming $length(rs_reduct) = r, length(\Delta red) = l,$ the worst time complexity of this increment algorithm is: $O\left(l(N+m)\ln(N+m) + r \cdot m \log m + \gamma \bar{r}_C \bar{r}_D + \gamma N - \varphi(lT)\right)$. The worst space complexity is about: $O(lN + r \cdot m + \gamma \bar{r}_C \bar{r}_D)$, if tasks are memory-resident. Generally speaking, $\gamma \bar{r}_C \bar{r}_D$ is much smaller than $t_C N$. If tasks are disk-resident, space complexity is only $O(N)$.

Below is a visualization to illustrate why the incremental algorithm is more efficient than the method which simply applies the original algorithm to whole data.

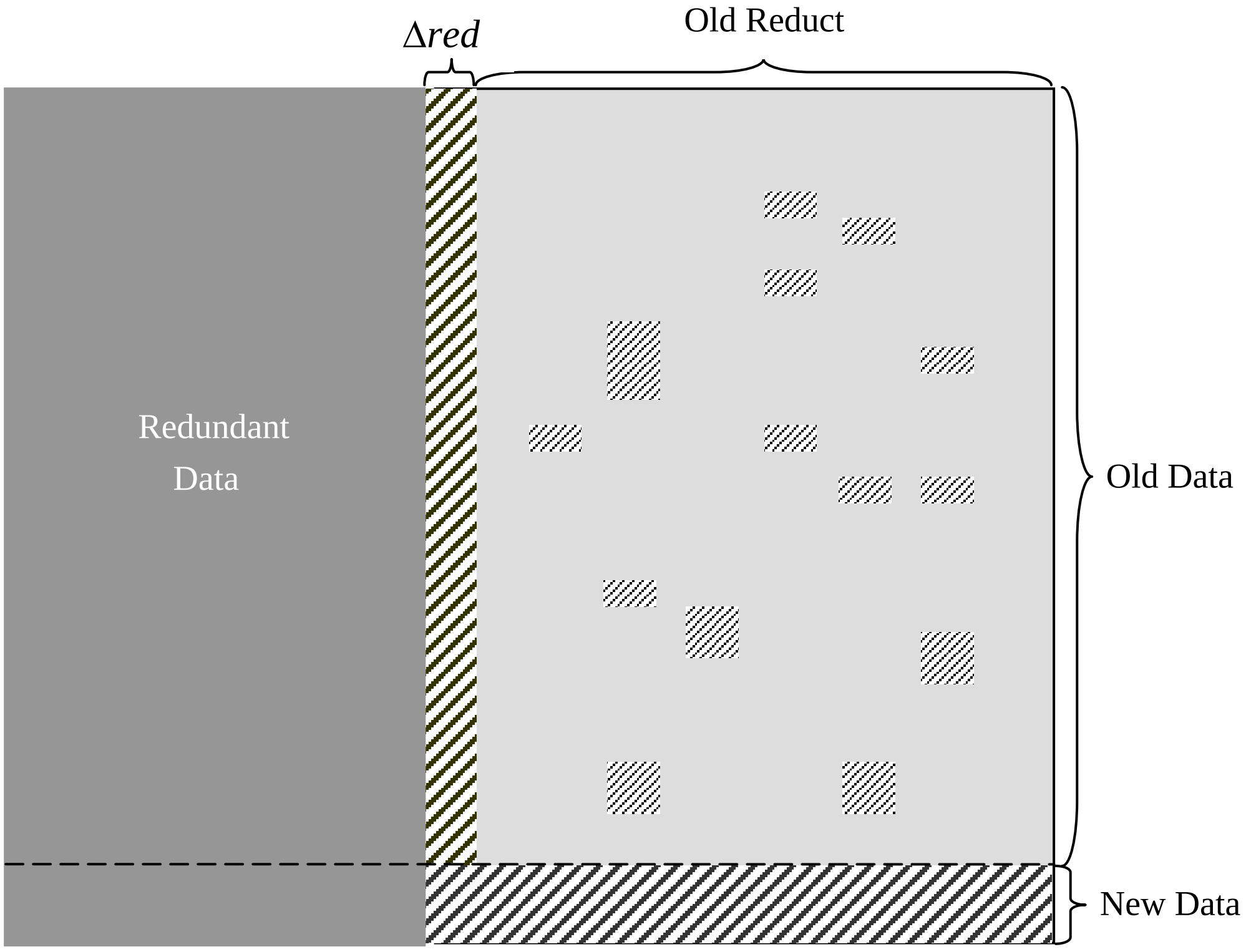

Figure 4: Effect Visualization of Incremental Algorithm

In Figure 4, the regions with diagonal are the regions that are sorted in incremental algorithm. As I have pointed out, sorting occupies a large part of computing time. Therefore, how to avoid repetitive sorting is a major problem for designing an efficient incremental version of algorithm TWI-SQUEEZE.

### 4.2 Decrease Version of Algorithm TWI-SQUEEZE

A doctor maintains a large decision table which contains all the diagnoses he has made. An object in this table is like this: if a patient has symptom 1 and symptom 2 and symptom 3 then give him some Rx. Here a symptom can be regarded as a condition attribute, and Rx can be seemed as decision attribute. To facilitate future work, he uses algorithm TWI-SQUEEZE to reduce the table.

Because the doctor wants to improve his leechcraft, he asks his patients for feedback on the therapies. And he finds that some of his prescriptions are wrong or useless. Therefore he decides to delete some objects in his table, but deletion of objects will cause some attributes useless in a reduct. Should he start from scratch to compute the reduct again?

Indeed there is no need to do so. He need only examine "cascade task intersections". If "cascade task intersection" at layer i has only two queues or sets, which means an object in one set is indiscernible from the objects in the other set, and if one set will be void after deletion, then he need remove attribute i from the redut. In this way, he can renew the reduct efficiently without repetitive computation.

### 4.3 Change of Condition Attribute Set

Condition attributes are the factors that will be taken into account for decision. Different users will prefer different subsets of condition attributes. And even for the same user, in different ear he will give different weight to the same attribute. A worse case is that some attributes which can be measured before now become immeasurable, or their prices are increased remarkably. Hence, a problem arises: if it is needed to add or remove one condition attribute to an existent reduct, can we renew the reduct with a little effort?

If one attribute is added to the reduct, it will cause some attributes in the reduct from indispensable to dispensable. How to find these attributes to be removed? A simple method is to use the same trick that is used in incremental algorithm. That is applying the proposition 1. Any attribute can be removed if the demarcation task must be done by it before can also done by the added attribute.

If one attribute is removed from the reduct, then we use the demarcation task that must be done by the attributes as the initial task, and use the thoughts in incremental algorithm to find a new reduct.

Therefore, a small change of condition attribute set can be handled efficiently, and no repetitive computing is needed to renew the reduct.

In a word, algorithm TWI-SQUEEZE is a scalable algorithm which can adapt to change of need efficiently.

## 5 Find a Short Reduct

### 5.1 Demarcation: Measure of Variety, Average Demarcation Information and Entropy, Information and Distinctiveness, Similarity and Difference

There are many kinds of information measures, among which Shannon entropy (or information entropy, communication entropy), is the most popular one and is applied everywhere. However, some people, including Shannon himself, are cautious about various applications of Shannon entropy (Shannon, 1956). Entropy is the measure of uncertainty[23]. In this sense, it can be a measure of information, which also increases certainty in effect. The basis of this measure is that we can obtain the probability distribution of a system.

---

[23] Strictly speaking, entropy is a measure of uncertainty of probability distribution in source alphabet. There are many kinds of uncertainty, such as uncertainty of dimensionality.

There are many extensions of Shannon entropy: (Rényi, 1959, 1976a, 1976b, 1976c), (Kullback, 1959, 1966), (Posner, 1967), (Arimoto, 1971), (Guiasu, 1971, 1977), (DeLuca, Termini, 1972, 1974), (Zhong, 1981a, 1981b, 1983), (Aczel, Forte, 1986), (Kapur, 1967, 2001), etc. Some of these generalized entropies, such as Rényi's $\alpha$ – entropy, Daroczy's $\beta$ – entropy (Arndt, 2001), Arimoto's $\gamma$ – entropy (1971), Havrda-Charvat's entropy (1967)[24], et cetera, are called parametric measures. Instead of measure of uncertainty, they are claimed to be measures of equality. These parametric measures are also useful[25]. They can reduce to Shannon entropy when parameters take some special value[26]. However all these measures either do not satisfy recursivity or both recursivity and additivity[27]. Recursivity is the most important property that distinguishes Shannon entropy from all sorts of generalized entropy measures.

Before we talk about information measures, one question has to be answered: what is information? It seems an obsolete question that has been answered completely. Uncertainty now becomes the alias of information. Measure of information is regarded as the measure of surprise. Why should I try to verbosely recall it again? Because I think there is some confusion on this topic.

The first question is whether there is only one sort of information? Must it measure observer's surprise?[28] If all the answers are yes, then what will cause "surprise"? An observer will surprise only when output fact is in inconsistent with expectation. When we see enough that event "X" happens frequently in the recent sequence of output, a rare output "Y" will surprise us, but "X" won't. Therefore, surprise is built on our previous experience, knowledge and expectation. Why should a rare event transfer more "information" than an ordinary event? It is because timeworn information is less valuable to an observer. Although the intuition of information comes from subjective "surprise" and "valuableness", "uncertainty" seems objective, so does the measure of uncertainty. But the perplexity is remained. Either "uncertainty" or "surprise" is built on repetitiveness, which is only meaningful to an observer. Therefore, measures of uncertainty or surprise can be called measures of equality (difference in distribution)[29] or *information with respect to observer*.[30]

---

[24] Havrda-Charvat's entropy: $\frac{1}{1-\alpha}\cdot\left[\sum_{k=1}^{n} p_k^{\alpha} - 1\right]$, $\alpha > 0$ and $\alpha \neq 1$. It is not recursive and additive. When $\alpha = 2$, Havrda-Charvat's entropy is a form of quadratic entropy (Arndt, 2001): $\sum_{k=1}^{n} p_k \cdot (1 - p_k) = 1 - \sum_{k=1}^{n} p_k^{\,2}$, which is in fact the *gini* index. Note the similarity and difference between quadratic entropy (*gini* index) and demarcation measure, when table is a complete and consistent non-decision table.

[25] "…the use of parametric measures of entropy gives us a great flexibility which we can use to our advantage." (Kapur, 2001)

[26] We can find infinite such functions, e.g. $H_\theta = \frac{1}{1-\theta}\sum_{i=1}^{n}\left(1 - p_k^{-\theta}\right)p_k^2$ or $H_\mu = \frac{1}{1-\mu}\sum_{i=1}^{n}\left(p_k^{\mu-1} - 1\right)p_k$ or $H_\eta = \frac{1}{1-\eta}\sum_{i=1}^{n}\left(1 - p_k^{\eta-1}\right)p_k^{\eta}$. However, not every such function satisfies necessary properties of equality measure.

[27] In fact, this so called additivity property reflects more "(statistic) independence" than "additivity".

[28] Some people don't think so, their view is that entropy is nothing to do with either content or form of information. "Statistic information … is nothing to do with content (of information). And it does not change with concrete form of information (e.g. translating words to binary codes). Hence it is also independent of form (of information). It is an abstract concept in statistics which reflects the statistic characteristic of expressive form of inform." (Zhu, 2000).

[29] Different from generalized entropies, Shannon entropy holds recursive property, one of the most important

If we adopt this opinion, then it is repetitiveness that makes sense to information measure of a system. That is, *repetitiveness* can influence the quantity character of information contained in a system, which in turn *influences what an observer can learn*. It seems find its support from our experiences. One white swan may make an observer think swan is white. It is clear that seeing more white swans will increase his belief. But “where is that process of reasoning, which from one instance draws a conclusion, so different from that which it infers from a hundred instances, that are no way different from that single instance?” It is the famous question asked by Hume. Keynes had given a clever answer to it. He pointed out that it is because of so called “Negative Analogy” (Keynes, 1921). In other words, it is the difference of similar objects that make sense to seemingly resemblant objects[31]. As a consequence, *absolute repetitiveness seems no use for learning in ideal situations*[32].

Perplexity seems not decrease, but is expanded to learning theory!

In my view, to dismiss the perplexity, we should make distinction between two sorts of information. One is *information with respect to observer*, and the other is *structural information, or variety*[33]*, contained in system*, which is nothing to do with repetitiveness. If two objects are identical, it means their difference is ignorable *in this system*. Keynes had stressed on the importance of variety in human cognition process: “The *variety* of the circumstances, in which the Newtonian generalization is fulfilled, rather than the number of them, is what seems to impress our reasonable faculties.” (Keynes, 1921) An object *itself* is informative in a system only when it is different from other objects in at least one aspect. Therefore, we should remove repetitive objects from universe because they do not add any information to the knowledge system and are insignificant from the view of variety of this

---

properties in measurement.

30 However, entropy is also objective in that it is nothing to do with content. And it reflects probability distribution of source alphabet objectively. Entropy and information measure are therefore regarded by some people two concepts with respect to sender and receiver, though their quantities are identical in ideal cases. What the author wants to stress on is that entropy is nothing to do with concrete systems, as concepts “cardinality” and “distribution” are nothing to do with content and structure of systems.

31 “Every new instance may diminish the unessential resemblances between instances and by introducing a new difference increase the Negative Analogy. For this reason, and for this reason only, new instances are valuable.” (Keynes, 1921)

32 However, redundancy, as a kind of constraint, can be used to restrain noise and correct errors. Therefore, redundancy in boundary region is useful in data mining, especially when some important attributes are forgot to be included in analysis.

What we have assumed in this paper is that, very value is deterministic. However, it is possible that values of some attributes are not certain and depend on some probability distribution. Fortunately, this kind of values can generally be transformed to deterministic values for decision making.

33 Some research also regards uncertainty the same thing as variety. For example, we can find the definition of uncertainty on http://pespmc1.vub.ac.be/ASC/UNCERTAINTY.html: a measure of variety such that uncertainty (H) is zero when all elements are in the same category. H increases with both the number of categories and their equiprobability. The uncertainty resulting from two or more sets of categories is the sum of the uncertainties of the sets of categories taken independently. H = the sum of P sub i times the log of P sub i, where P sub i is the probability of an element being in the Its category. Since the categories must be specified by an observer, the uncertainty of a system may be different as seen by different observers. (Here, H is Shannon entropy-author)

Ashby had given two measures for variety: in relation to a set of distinguishable elements, either (l) the number of distinct elements, or (2) the logarithm to the base 2 of that number, the context indicating the sense used. When variety is measured in the logarithmic form, its unit is the bit. (Ashby, 1956, see: http://pespmc1.vub.ac.be /ASC/VARIETY.html). The second measure happens to be a special case of Shannon entropy, although the ideas are different. However, *in the case of equal probability*, Shannon entropy holds monotonicity property of source alphabet. It does can reflect some aspects of variety of system in this sense.

system[34]. Or the information conveyed by them is negligible in this system. This is a main difference between measures of variety (information of system), and entropy (information for observer), when they are applied to data mining and knowledge discovery. For entropy, it is repetitiveness, not difference of objects themselves, that makes sense in the measurement, which is closely related to coding.

The second question is how to measure variety? The three key properties, i.e. "monotonicity of source alphabet", "continuity", and "recursivity", which should be held by any measure of uncertainty, must also be included in consideration of any measure of variety. However, when measure is defined on $Z^+$ instead of $R^+$, requirement of continuity can be loosed a bit. Another requirement is that variety measure should not rely on probability distribution. And in my view, a good candidate should also reflect difference between any two objects. It is difference that underpins all kinds of measures for uncertainty and variety[35]. And any definition of information should take difference into account. In my view, *information is the thing that reflects difference or any relation built on difference.* The relation of uncertainty and difference will be discussed in Appendix E. I'm not lonely in taking this view. G. Longo had also pointed out long before that information should be contained in the difference of objects (Longo, 1975)[36]. I think structural information of system can be measured directly on the basis of differences between distinctive objects. At last, any measures of variety must satisfy that output variety cannot exceed variety of original system at any time[37].

Now, I'll show that a novel information measure based on so called demarcation information has the necessary key properties of measure for variety. Although demarcation information can be defined on *a cover of universe*, I'll only discuss some key properties of them on the basis of *a partition of universe*, because these properties are comparable to that of Shannon entropy.

Notice that Although variety measure has nothing to do with repetition and distribution, in a multivariable system (table) that has no repetitive objects, the partition induced by one variable maybe contains repetition, if we want to use the precious recursivity property, the properties of variety measure must be proved based on some distribution.

Let $F=(X_1,X_2,\cdots,X_n)$ be a partition of universe, then the demarcation information of $F$

---

[34] However, this *strategy* cannot reflect equality of distribution of source alphabet, which is closely related to measure of surprise and information. And repetitiveness can be used as a tool that helps reconcile inconsistence in knowledge system because noise is inevitable in practical application, even if we take this strategy.

[35] In fact, entropy measure can be regarded as measure of difference in distribution, or equality.

[36] "Once you realize that it is information that triggers actions and energy, and that information is contained in the differences between objects and energy and not in the objects themselves, you also become aware that the actual objects used for communicating, i.e. for conveying information, are unimportant. Only differences (i.e. relations) matter. This is actually the starting point of any theory of information." (Longo, 1975)

Some people perhaps will do not agree with it. Mere repetition sometimes can also convey information, such as stress or conveying emotion information. But, why can repetition function in this way? It is only because repetition itself is *different* from normal expression, which can convey some specially meanings. Here, repetition is used in coding, and is not redundant (or repetitive!) *in the context*.

If the analysis is to disclose the attitudes of candidates, then repetition seems also significance. If candidate A holds all the opinions of B, we say they share the same viewpoint. However, in this context, candidate himself is involved in the analysis and therefore is an additional dimension. Hence, there is no repetition in this example. Less difference between two objects only means the candidates share more opinions.

Notice that, here we only talk "difference" among objects. We can also talk about difference inside an object. And we can also talk about difference between different eras of one object.

[37] Any entropy measures, including *gini* index, do not satisfy this property.

is $S(F)=\{(x,y)\mid x\in X_i \wedge y\in X_j \wedge i\neq j\}$;[38] the demarcation information measure of *F*, or variety of system, is $\varpi(F)=Card(S(F))=\sum_{\substack{i,j\in[1,n],\\ i\neq j}} Card(X_i)\cdot Card(X_j)=2\sum_{i<j} Card(X_i)\cdot Card(X_j)$.[39, 40] We can use some *k*-ary unit to code the difference, i.e. $log_k(\varpi(F))$. It is not an average amount of information conveyed by an information source, except that it is divided by cardinality of universe[41].

Note that this definition doesn't take some metric to measure the actual distance between two objects because our interest in this paper is mainly on classification. In this context, we only need know whether two objects are different or not. However, a more general consideration is to take some distance into account. For example, we can use the hamming distance between two objects. That is, it counts the difference of two objects on how many attributes they are different.

Let $S(X_i,X_j)=\{(x,y)\mid x\in X_i-X_j \wedge y\in X_j-X_i \text{ or } y\in X_i-X_j \wedge x\in X_j-X_i\}$. For $\forall x\in U$, $\forall s\subseteq P(F)$, $\varphi(x\in s)$ is the demarcation information function that reflects amount of demarcation Information conveyed by "$x\in s$" in given partition $F$, if $\varphi(x\in s)=Card(S(\{x\},\bar{s}))=Card(\bar{s})$.[42]

Assume $F=(X_1,X_2,\cdots,X_n)$ and $F'=(X_1',X_2',\cdots,X_n')$ are two partitions of universe. Then, it is easy to prove that:

$$\varphi(\{x\mid x\in X_i \wedge x\in X_j'\})=Card\left(S(\{x\},\overline{X_i})\cup S(\{x\},X_i\cap\overline{X_j'})\right)$$

$$=Card\left(S(\{x\},\overline{X_j'})\cup S(\{x\},X_j'\cap\overline{X_i})\right)=Card\left(\overline{X_i\cap X_j'}\right).$$

Assume $Card(X_i)=x_i$. $Card(U)=N$. We have some key properties about function $\varphi$ and $\varpi$, as shown in Appendix D.

[38] If there are many missing values in table, then there is no complete partition on universe. However, we can still define demarcation information by taking only the known demarcations into account.

[39] For the partition induced by all attributes, *Card* ($X_i$)=1. But for partition induced by some subset of allattributes, *Card* ($X_i$) $\geq 1$.

[40] If we are only talking about measure of difference and information, $\varpi(F)=0.5\times Card(S(F))=0.5\times \sum_{\substack{i,j\in[1,n],\\ i\neq j}} Card(X_i)\cdot Card(X_j)=\sum_{i<j} Card(X_i)\cdot Card(X_j)$ is a better choice. However, if symmetry is concerned, we need not scale down the quantity to half. It is well known that symmetry is beautiful. Hence, it is more an aesthetic preference than technique cause. And in some special cases, such as when reduction that maintains $\leq$ relation in a system, $\varpi(F)=Card(S(F))$ is a better choice. From now on, I will ignore 0.5 everywhere, although retaining it will help understand their true meanings of definitions.

[41] It is comparable to Shannon entropy. Because in a given universe, the cardinality of it is fixed, we can also compare measure of demarcation information instead of its average measure to Shannon entropy.

[42] From the view of rough set theory, $F$ is knowledge about the universe. And $S$ is a concept in knowledge $F$. This measure is a simplified case of a more complex measure. In that measurement, we should first give distance of two objects in certain attribute, and compute the average distance between a concept and its complement concept. The cardinality of its complement concept divided by average distance is the amount of information. It is intuitive that demarcation of two objects with less distance will bring more information. But here, we assume all the distance is the same, and therefore the average distance is assigned 1.

Assume $s, e \subseteq U$, $\varphi(x \in s | x \in e) = \varphi(x \in s \cap e) - \varphi(x \in e)$ can be regarded as the measure of conditional demarcation information which tells us how much information an observer can obtain from massage "$x \in s$" when he already knows "$x \in e$". Though it also tells us the amount of information an observer should try to get before he know the fact that "$x \in s$", assuming he already knows "$x \in e$". $\varphi(x \in s; x \in e) = \varphi(x \in s) - \varphi(x \in s | x \in e) = \varphi(x \in s) \cap \varphi(x \in e)$ can be called mutual or shared demarcation information function that measures the demarcation information contained in both concepts $s$ and $e$. It also tells us that knowing $x \in e$ will bring how much useful (demarcation) information to an observer, when $x \in s$.

If $F_1 = (X_1, X_2, \cdots, X_n)$ and $F_2 = (Y_1, Y_2, \cdots, Y_m)$ are two partitions of universe, then let $\varpi(F_1 | F_2) = \varpi(F_1) - \varpi(F_2)$, $\varpi(F_2 | F_1) = \varpi(F_2) - \varpi(F_1)$. $\varpi(F_1 | F_2)$ can be called conditional variety measure of $F_1$ with respect to $F_2$. It is a measure of *difference of variety* of a system induced by different partitions. $\varpi(F_1) - \varpi(F_1 | F_2) = \varpi(F_2) - \varpi(F_2 | F_1) = \varpi(F_1) \cap \varpi(F_2)$ is mutual or shared variety measure of two different partitions of a system. It is easy to prove that: $\varpi(F_1), \varpi(F_2) \geq \varpi(F_1) - \varpi(F_1 | F_2) = \varpi(F_2) - \varpi(F_2 | F_1) \geq 0$.

From the proofs in Appendix D, we can see that all the key properties of Shannon entropy can find their natural counterparts in variety, or demarcation information, measure.

Now we will discuss an instance. The universe is a universe of bird with 4 *distinct* objects, which were owned to one of my friend. There are two attributes defined on this universe. That is {Speckle, Color}. The corresponding table is shown below.

| Bird | Speckle | Color |
|---|---|---|
| Tom | Y | White |
| Susan | N | White |
| Joan | Y | Black |
| Jack | N | Black |

Table 7: Four Birds

One day, my friend phoned me that one bird was dead. I want to know which one and therefore I asked: "Is it Tom (a white bird with speckles in its wings)?" "No." answered my friend. Note that in this message, I only knew the dead bird is not Tom. It is one unit of demarcation information.

"Is it Susan (a white bird without speckles in its wings)?" "No. You have the last guess", said my friend.

"Is it Jack (a black bird without speckles in its wings)?" "Yes!" said my friend.

Note that in this process, the message is identical to tell me the dead bird is not Joan, a black bird with speckles in its wings. Now there was no *uncertainty* anymore. In this process, I received three units of demarcation information.

In fact, I can also ask my friend in a different way.

"Is it White?" "No."

In this process, I receive two units of demarcation information. That is, this bird is not Tom or Susan.

"Does it have speckled wings?" "No."

"Aha, it is Jack!" I know it because I've received all three units of necessary demarcation information. Indeed, before I asked my friend, I've already know that to be informative, three units of demarcation information are necessary and sufficient, because any bird needs three units to be discerned from the others in this universe.

Still, I can guess in another way. That is, first I guess its wings are speckled. Then I guess it is white.

Now, let's analyze these processes. In the second and third process, I first ask whether the bird is white, followed by the other question. From the view of Shannon entropy, the uncertainty in these two questions is identical, because the two attributes are clearly independent, and their independency is absolute. But from the view of demarcation information, their independency is relative with respect to amount of new information they convey.

We can also look at the previous example in another point of view. That is, two demarcations are now discriminated by how many attributes are counted respectively. For instance, there are one unit of demarcation information between Tom and Jack in the previous viewpoint. But now we say there are two unit of demarcation information between these two birds. One unit discriminate their Color, the other unit discriminate their Speckle.

For example, when I know the dead bird is not White, I receive two units of demarcation information (it is not Tom, Susan). When I know the dead bird is not Speckled, I also receive two units of demarcation information (it is not Tom, Joan). When I know the unfortunate bird is not White and not Speckled, I receive 4 units of demarcation information (one unit for distinguishing Jack and Joan; one unit for Jack and Susan; two units for Jack and Tom).

This view makes sense only when attributes are "independent" and their influences in discrimination are addible. Note that this "independence" is nothing to do with that "independence" in probability.

In conclusion, *"independence" is relative. It depends on what view we take.* When I know the bird is not white, I get one bit information in the context of the whole universe. When I know its wings are not speckled, I still receive one bit information, but in the context of half universe, that is {Joan, Jack}. Their norms are in direct proportion with the cardinality of respective contexts. This is from the view of Shannon theory. But from the view of demarcation information theory, one unit of information has the same context, which is the whole universe. Therefore, the case that is "independent" in Shannon theory is no longer true in demarcation information theory, and vice versa. *The meaning of "independence" in demarcation information theory is the same as that in rough set theory.* That is, when removal of an attribute will change positive region, then the attribute is called "independent" to other attributes. Clearly, *they are two kinds of different viewpoints of "independence"*.

What we've heard is a somewhat sad story. Therefore, we change the scenario a little:

That day, my friend phoned me that he would free one bird.

"That's a great idea." I said, "How will you do it? I mean which bird will you free? Do you choose it arbitrarily?"

"I do it by dicing." My friend said, "If the dot is 1, I will choose Tom; if the dot is 2, Susan will be chosen; if it is 3, I'll say goodbye to Joan; but if it is 4, 5, or 6, My Jack will be free."

After two days, I phoned my friend.

"Now tell me which bird is free." I said, "Um, wait a moment. Let me guess it. It must be Jack."

"Yes, you've got it!" answered my friend.

Now, let's analyze this event. In the guessing, what I need know is not the information about universe of bird, but that about universe of gambling. This universe is made of some sets of numbers: {{1}, {2}, {3}, {4, 5, 6}}. The purpose of Shannon entropy is to evaluate at least how much guesses or queries should be made averagely, which determines the cost in coding and transmitting the answers (information). It need not know what the system is really like. It is the real reason why Shannon entropy can successfully exclude so called semantic information. But if I query based on the information of the specific system, I should know the attributes defined on the universe and hence *can* also ask "Is it white?" And two guesses are needed. Therefore, the purpose of these two measures and universe they are dealing with are totally different. The aim of demarcation information measure is to measure system information, or model the information complexity of system; and the purpose of Shannon entropy is to measure and model information complexity based on observer[43].

As for data mining and knowledge discover, on one hand we should study systems themselves because what we want to extract is the information hides in these systems; but on the other hand we should also evaluate the extracted information or knowledge subjectively. Making these two process distinction is important. If repetitiveness itself is of interests (that is information is contained in the distribution density itself), then we should use Shannon entropy. Otherwise we should adopt variety measures.

Because variety measure has nothing to do with probability distribution, the additivity property which reflects statistic independency becomes meaningless to variety measure.

By comparing with Shannon entropy, we can find some advantages of demarcation information measure, although they are used for different purpose.

1) We need not know the probabilities distribution of source alphabet.

Although subjective probability or empirical probability distribution can be adopted to handle this problem (Zhong, 1996), sometimes even empirical probability is still not easy to be obtained. Moreover, if signal isn't random, there is no corresponding statistic model (Zhu, 2000). In this case, any statistic information measure is not exact[44].

2) Usually, there are some repetitive objects in a table. These redundant objects, which will not add information to system, will influence the probabilities distribution, and therefore Shannon entropy. But they will not influence $\varpi(F)$. It is indeed the difference between equality measure and variety measure.

[43] However, we can also look at Shannon entropy in a different view. See Appendix E for some details.

[44] In coding theory, there are some adaptive coding algorithms which can tackle this problem to some extent.

3) $\varpi$ (*F*) are suitable for incomplete system. Although Rényi entropy[45] can handle incomplete probabilities distribution problem, it is impossible for Rényi entropy to handle a table with many missing values.
4) $\varpi$ (*F*) can be defined on "cover", which seems out of reach of Shannon entropy.
5) Using demarcation information, we can not only measure true information, but also can measure false information. It will be discussed in Appendix G.
6) Nowadays, data is usually dynamically increasing. We have to compute Shannon entropy repetitively to adapt to new data. But for $\varpi$ (*F*), there is no need to start from scratch computing it, which is ensured by two properties.

**Property 5.1:** $F_1 = (X_1, X_2, \cdots, X_n)$ is a partition of $U$ induced by attribute set $C$. $F_2 = (Y_1, Y_2, \cdots, Y_n)$ is a partition of new data $U'$ induced by the same attribute set $C$. Here, the objects in $X_i$ are identical to that in $Y_i$ under $C$. $F$ is a partition of new data $U \cup U'$ induced by $C$. Then we have: $\varpi(F) = \varpi(F_1) + \varpi(F_2) + \sum_{i \neq j} x_i y_j$.

**Proof:** $\varpi(F) = \sum_{i \neq j} (x_i + y_i)(x_j + y_j)$

$$= \sum_{i \neq j} (x_i x_j + y_i y_j + x_i y_j + x_j y_i)$$

$$= \varpi(F_1) + \varpi(F_2) + \sum_{i \neq j} x_i y_j. \quad \blacksquare$$

Note that *U'* is generally much smaller than *U*. And so is $y_i$, compared with $x_i$. Thus the computation time of $\varpi(F_2) + \sum_{i \neq j} x_i y_j$ is also relative small, compared with the time that is used to compute $\varpi$ ($F_1$).

If the complete value set of attribute is unknown before the whole data is at hand, we can utilize another method to compute $\varpi$ incrementally.

**Property 5.2:** $F_1 = (X_1, X_2, \cdots, X_n)$ is a partition of $U$ induced by attribute set $C$, and $X_i \neq \phi$. $F_2 = (Y_1, Y_2, \cdots, Y_n, \cdots, Y_m)$ is a partition of new data $U'$ induced by the same attribute set $C$. $n \le m$, and $Y_j \neq \phi$ $(j \in [n+1, m])$. Here, the objects in $X_i$ are identical to that in $Y_i$ under $C$ $(i \in [1, n])$. $F$ is a partition of new data $U \cup U'$ induced by $C$. Let $F_2' = (Y_1, Y_2, \cdots, Y_n)$, $F_2'' = (Y_{n+1}, Y_{n+2}, \cdots, Y_m)$. Then we have:

---

[45] For an incomplete probabilities distribution that satisfies: $\begin{cases} 0 \le p_i \le 1 & i = 1, \cdots, n \\ \sum_{i=1}^{n} p_i < 1 \end{cases}$

Rényi entropy is : $H_R(p_1, \cdots, p_n) = -\sum_{i=1}^{n} p_i \log p_i \Big/ \sum_{i=1}^{n} p_i$.

Later Rényi proposed his $\alpha$–entropy: $\frac{1}{1-\alpha} \cdot \ln \sum_{k=1}^{n} p_k^{\alpha}$, $\alpha > 0$ and $\alpha \neq 1$. When $\alpha = 2$, it is Rényi's quadratic entropy.

$$\varpi(F)=\varpi(F_1)+\varpi(F_2)+2\sum_{\substack{i\in[1,n]\\ j\in[1,m]\\ i\neq j}} x_i y_j \cdot$$

**Proof:** $F=\left(X_1\cup Y_1, X_2\cup Y_2,\cdots,X_n\cup Y_n,X_{n+1},\cdots,X_m\right)$.

$$\varpi(F)=\sum_{\substack{i,j\in[1,n]\\ i\neq j}}(x_i+y_i)(x_j+y_j)+\sum_{\substack{i,j\in[n+1,m]\\ i\neq j}} y_i y_j+2\sum_{\substack{i\in[1,n]\\ j\in[n+1,m]}}(x_i+y_i)y_j$$

$$=\sum_{\substack{i,j\in[1,n]\\ i\neq j}}\left(x_i x_j+y_i y_j+x_i y_j+x_j y_i\right)++\sum_{\substack{i,j\in[n+1,m]\\ i\neq j}} y_i y_j+2\sum_{\substack{i\in[1,n]\\ j\in[n+1,m]}}(x_i+y_i)y_j$$

$$=\sum_{\substack{i,j\in[1,n]\\ i\neq j}} x_i x_j+2\sum_{\substack{i\in[1,n]\\ j\in[n+1,m]}} y_i y_j+2\sum_{\substack{i\in[1,n]\\ j\in[1,m]\\ i\neq j}} x_i y_j$$

$$=\varpi(F_1)+\varpi(F_2)+2\sum_{\substack{i\in[1,n]\\ j\in[1,m]\\ i\neq j}} x_i y_j \cdot \quad \blacksquare$$

7) Up to now, we only discussed about demarcation of two values semantic. In this semantics, two objects either form a demarcation or not. However, it is well known that there are different semantic distance between concepts and objects. Fortunately, it is *natural* and *agreeable* to talk about demarcation with semantic distance of two objects by add the difference to demarcation[46]. See Appendix F for some details. Clearly, it has expected applications in data mining and knowledge discovery.

8) Now we can not only talk about measure of information, but also the meaning of information itself.

Assume the semantic distance of a demarcation ($x_i$,$y_j$) is $w_{ij}$, $\varpi$ ($F$)= $\sum_{\substack{i,j\in[1,n],\\ i\neq j}} w_{ij} Card(X_i)\cdot Card(X_j)$ is the total difference contained in a system. And $Aw(F)=\frac{\sum_{\substack{i,j\in[1,n],\\ i\neq j}} w_{ij} Card(X_i)\cdot Card(X_j)}{\sum_{\substack{i,j\in[1,n],\\ i\neq j}} Card(X_i)\cdot Card(X_j)}$ is the average demarcation distance of the system, which can be regarded as the unit of demarcation information in this system. $A\varpi(F)=\frac{\sum_{\substack{i,j\in[1,n],\\ i\neq j}} w_{ij} Card(X_i)\cdot Card(X_j)}{Card(U)(Card(U)-1)}$ is the average semantic distance of the system.

Note that Shannon entropy has nothing to do with semantic element[47]. Therefore, we

[46] However, we can talk about semantic distance of two objects only when they are comparable in some aspects, including so called “nonalignable differences”. For example, the steering wheel of a car is comparable to the handlebar of a motorcycle, because they are both the direction control part of these two sorts of vehicles, although their difference is called nonalignable difference (Best, 1998).

[47] Entropy is nothing to do with content. For example, assume there are 500*600 pixels on a video display. Every point has a value selected form ten candidates randomly. Then one picture can convey about $10^6$ bits information, according to Shannon theory. Assume there are one thousand words in an article, and every word is selected from a

can talk about Shannon entropy without some semantic distance. But, for demarcation information, we can talk about it only when some semantic distance is given. For instance, given a semantic distance function and a threshold, an arbitrary partition of universe may be contains zero demarcation information. But it can have nonzero Shannon entropy. Hence, in my view, demarcation information is more suitable to guide unsupervised learning, in the background of some distance and threshold.

But we should be more cautious that not everything is relevant and comparable. We can talk about the difference of two objects only when they are comparable *in the context.* The other point that must be mentioned is that information is relative. There is no absolute information. What we can talk about is information in *a certain universe*. An example can cast some light on it, which can be called "the world of white". In this universe, all the objects are white. According the definition, the proposition "all cats are white" contains no demarcation knowledge. In other words, knowing this fact will not help the preys (such as mouse) surviving in this world. Color has no practical significance *in this universe*. In a word, what we can talk about is knowledge *in certain universe*. And the truth in knowledge is relative and makes sense in a context. The knowledge human can discover is also the knowledge of *our universe*. Whether it can extend to other universes is a thing that should be scrutinized carefully. This fact also tells us that, to discover useful knowledge in data, good techniques are and only are one aspect that leads to our objective, the other possibly more important aspect is what data we are mining on, because it determines the quality of knowledge we can discover. We can "discover" any "knowledge" if we do not select data properly.

I've argued the right of demarcation information and its measure by comparing it to Shannon entropy. We can find its applications in research fields such as classification and pattern recognition. In this section, I try to justify that there exist some other information measures which reflect different aspects of information. Every measure is not sufficient to reflect all aspects involving not only system but also observers. In cognitive psychology, concepts have different levels. It is said that basic level concepts have high differentiation, which includes both the measure of informativeness and distinctiveness (Best, 1998). My paper discloses that the measure of variety also has the qualities of information measure. Since variety measure can be used to measure distinctiveness, therefore distinctiveness measure and the so called "information" measure can be unified by variety measure.

Is combination of Shannon entropy and demarcation information measure sufficient for measure of every aspects of information? The answer is no. There exist other different measures that are complements to these two measures and can be used in knowledge discovery, such as descriptive complexity or Kolmogorov-Chaitin complexity (Sipser, 1997)[48]. However, it is not to be discussed in this paper to avoid turning off our topics.

---

dictionary of $10^5$ words randomly, then such an article can convey about $1.66*10^4$ bits information. Hence, a picture with low resolution and very limited colors can convey much more information than a moderate size articles. (Zhou, 2002, with small change).

In fact, excluding semantic element is more an achievement than shortage *in communication theory*.

[48] Kolmogorov (1965) had pointed long before that there are more than one method to define the concept of information quantity.

## 5.2 Heuristic Information and Greedy Strategy

The nature of algorithm TWI-SQUEEZE is local, and the efficiency of the algorithm just originates from this feature. Therefore, the algorithm cannot ensure finding a short reduct by itself. To find a short reduct, we must combine something else.

If we relax the limitations and aim to find an approximate reduct, it is possible to obtain a much shorter reduct. It will be discussed later.

We can also use some small tricks. For example, when the margin or difference of two successive cascade tasks is smaller than a given parameter (it can be adapted in computing), we skip the attribute which only affords a negligible fraction of task.

There are other more attractive methods to find a short reduct. One thought is to utilize global feature of data. Another thought, which is mostly adopted, is to use heuristic information to guide the algorithm. In this paper, the latter will be discussed.

Interesting enough, algorithm TWI-SQUEEZE can easily combines with any heuristic information, because attribute order of the initial table is just the reflection of heuristic information. It means that we can talk about heuristic part and reduction part independently.

Up to now, there are mainly three kinds of heuristic information that have been used in various heuristic algorithms: 1. occurrence frequency of an attribute in discernibility matrix; 2. conditional entropy or mutual information; 3. definitions of attribute significance based on rough set theory[49]. The second and third sorts of heuristic information can be regarded as relative significance of attributes. However, we should be cautious to accept relative significance of attribute because it is easy to prove that attributes depend on each other to form their "significance". And as a result, it is easy to prove that if some bad attributes are included in the analysis or some important attributes are not included, the results of "significance of attributes" will cheat us.

As for algorithm TWI-SQUEEZE, a good choice is to combine the first kind of heuristic information. And the heuristic information can be computed efficiently in a similar way as Hoa and Son (1996) has done to a non-decision table. It will be introduced after some concepts are put forward. In Appendix C, I will give a small example that computes the heuristic information.

As we have introduced before, *xRy* is called a unit of (demarcation/indiscernible) information, if x can (be/not be) discerned from y by some attribute. A cover based definition is given in appendix A. Definitions listed below are based on an assumption that there is no redundancy in the universe.

**Definition 4:** $A$ is a (demarcation/equivalence) relation, then the amount of (demarcation /indiscernible) information of $A$ is $Card(A)$.[50] We use $S(A)$ to denotes demarcation

---

[49] There are mainly two kinds of definitions, and they are not identical, although they are similar. One is: $\sigma_{(C,D)}(a)=1-\frac{\gamma(C-\{a\},D)}{\gamma(C,D)}$ or $\sigma_{(C,D)}(a)=\gamma(C,D)-\gamma(C-\{a\},D)$ where $\gamma(P,D)=\sum_{X\in U/D}\frac{Card(\underline{P}(X))}{Card(U)}$. The other is: $\sigma_{(C,D)}(a)=1-\frac{\gamma(C,D)}{\gamma(C\cup\{a\},D)}$ or $\sigma_{(C,D)}(a)=\gamma(C\cup\{a\},D)-\gamma(C,D)$. A definition in rough analysis literature is introduced by Guan and Bell (1998).

[50] In algorithm TWI-SQUEEZE, there is an invariable of demarcation information which is the footstone of my thought: $IND(D)$ and its size $Card(D)$. Find a reduct can be regarded as finding a subset of conditional attribute set which still maintains the invariable, and in the same time its proper subset cannot do so.

information and use $I(A)$ to denotes indiscernible information of $A$.

Because an attribute set induces a relation, we also use $S(P)$ to denotes demarcations induced by attribute set $P$, and $I(P)$ denotes indiscernible information induced by $P$.

**Definition 5:** $KS=(U,C\bigcup D)$ is a decision table. The pair *xRy* is called a unit of relative (demarcation/indiscernible) information to attribute *D*, if *x* can (be/not be) discerned from *y* by *R*, and in the same time, *x* can be discerned from *y* by *D*. We use $S(C,D)=S(C)\bigcap S(D)$ to denotes all the relative demarcations of $KS$.

A set $U$ and equivalence relation $ER$; $Card(U)=N$. $U/ER=\left\{X_k \,|\, U=\bigcup X_k\left(X_i\bigcap X_j=\phi, i\neq j\right)\right\}$, $Card(U/ER)=n$. $DR$ is the complementary relation of $ER$. Then we have:

**Proposition 2:** $Card(DR)=\sum_{i\neq j} Card(X_i)\cdot Card(X_j)=2\sum_{i<j} Card(X_i)\cdot Card(X_j)$.

**Proof:** From definition we have it. ■

**Proposition 3:** $Card(DR)\leq \frac{n-1}{n}\cdot N^2$. Here $n$ is the number of equivalence classes of $IND(D)$.

This proposition is the same as property 8 of demarcation information function which has been proved in section 5.1.

If (*x*, *y*) is a unit of demarcation information, then (*y*, *x*) is also a unit of demarcation information. But when we perform attribute reduction, we can regard (*x*, *y*) and (*y*, *x*) as two sides of one demarcation, and as a consequence count them only once. ***It is easy to prove that the result of attribute reduction in this way is the same as the result of attribute reduction which discriminates (x, y) and (y, x).***

**Remark 2:** proposition 3 means that the maximal number produced in the process of computing heuristic information is less than $0.5N^2$ *if we consider* (*x*, *y*) *and* (*y*, *x*) *as one.* When *N* is very large, we can use some tricks, such as approximation, using *log*(*Card*(*DR*)) or simply to divide all the results by *N*.

**Definition 6:** Let $I(C_i,lT)$ be the amount of relative indiscernible information of $C_i$ relative to $lT$; $S(C_i,lT)$ be the amount of relative demarcation information of $C_i$ relative to $lT$. $lT$ is cascade task. Assuming the indiscernible information contained in $lT$ is $I(lT)$, then we have: $I(C_i,lT)=IND(C_i)\bigcap I\left(lT\right)$. $S(C_i,lT)=I\left(lT\right)-I(C_i,lT)$ [51].

The thought is that every time we select an attribute which can render the maximal relative demarcation information that is needed. However, in the algorithm we use relative indiscernible information instead because it is easier to compute and the numbers produced is smaller. How to compute indiscernible information? Assuming *T* is the current cascade task formed at attribute *c*, which has *k* queues and *a* D-Regions. $x_{ij}$ is length of a segment, $i\in[1,k]$, $j\in[1,a]$. Then relative indiscernible information of $c$ relative to $lT$ can be computed by:

---

[51] It is indeed the difference of indiscernible information contained in two successive cascade tasks.

$$Card\left(I(c,lT)\right)=\sum_i\left\{\left(\sum_j x_{ij}\right)^2-\sum_j x_{ij}^2\right\}.$$

But still, we can compute $0.5\times Card(I\ (c,T))$ instead since ($x$, $y$) and ($y$, $x$) can be regarded as one. Note that when there is no repetitive object in the universe, this kind of heuristic information has the same effect as occurrence frequency of an attribute in discernibility matrix when attribute reduction is concerned.

If we use information entropy instead as heuristic information, the entropy can be computed by:

$$p_i=\sum_j x_{ij}\Big/\sum_i\sum_j x_{ij};$$
$$E(c,lT)=-\sum_i p_i\log_2 p_i.$$

It has a neater form than conditional information entropy (Yu, Yang, Wu, &Li, 2001; Wang, Yu &Yang, 2002). However, I will use indiscernible information as heuristic information because it seems can be combined with algorithm TWI-SQUEEZE more tightly. Moreover, it is more convenient to count in an incomplete system.

Below is the outline of the heuristic algorithm.

Input: A decision table $KS=(U,C\bigcup D)$, which is described before.

Output: $reduct$.

Step1. $s_reduct\leftarrow\phi.\quad C^{'}\leftarrow C.$

Step2. Sort the table under $D$, and remove repetitive objects (assume the reduced universe is $U^{'}$). We obtain the invariable of demarcation information: $0.5\sum_{i,j}Card(X_i)\cdot Card(X_j)\cdot\quad\left(X_k\in U^{'}/D\right)$

Step3. Sort the initial task $lT$ under condition attributes separately. Select $c\in C$' s.t.

$Card\left(I(c,lT)\right)=Min_{j\in C^{'}}Card\left(I(C_j,lT)\right).$ And form new cascade task. Let it be the initial task $lT$. $s_reduct\leftarrow s_reduct+\{c\}.\quad C^{'}\leftarrow C^{'}-\{c\}.$

Step4. Repeat Step3 until current cascade task is void.

Step5. Run algorithm L2R-SQUEEZE and obtain a reduct.

The soundness is ensured by theorems in appendix A, which also tell us that no relative demarcation will be lost in reduction.

All the heuristic algorithms mentioned in this section can be regarded as the counterparts of sequential suboptimal search methods in feature selection literature. In my view, both literatures can benefit each other a lot because they share a similar problem.

I've discussed how to find a short reduct. But how can we find a shortest or minimal reduct? It is a much more difficult problem which has been proved to be NP-hard, although greedy strategy can often "come across" a minimal reduct in practice as some works have reported[52]. My algorithm cannot ensure finding a minimal reduct too, because all these heuristic algorithms belong to one class, the prominent feature of which is greedy, in nature.

[52] "…they can obtain the minimal reduction in plenty of experiments." (Wang &Miao, 1998). However, in this paper, Wang Jue et al also pointed out that all algorithms are not complete in the strict sense of "minimal reduct".

### 5.3 Replacement

In section 5.1 I've introduced how to find a short reduct using heuristic information. In this section I'll introduce another method that can help finding a short reduct.

Let's first see a special case. If we obtain a reduct, which contains all the condition attributes except one, for example $C_i$, then we can check whether $C_i$ can replace any attributes in the reduct or not. Using definition 4.1, this kind of checking is very fast, and we can deem it as one unit polynomial complexity operation. If $C_i$ can replace more than one attribute, than we can obtain a new reduct which is shorter than the original one. Furthermore, if some attributes in the original reduct can replaced by less attributes, I call this kind of replacement as shrink replacement or s-replacement, and the original reduct is s-replaced.

Below is a relevant proposition:

**Proposition 4: (1) If a reduct is minimal, then it cannot be s-replaced by any combination of other attributes that are not in the reduct.**

**(2) If** $Card(C) = n, Card(reduct) = m, Card(\min reduct - reduct) = l,$ **then after at most** $C_{n-m}^{l}$ **replacements, we can obtain a minimal reduct**[53]**.**

**(3) For a given positive integer** $a,$ **we can find any reduct,** $reduct^{'},$ **that satisfies** $Card(reduct^{'} - reduct) \le a,$ **in** $C_{n-m}^{\min(a, n-m)}$ **replacements.**

**Proof:** (1) From definition we have it. However, its inverse proposition is not true.

**(2)** If $l = 0$, the reduct obtained is a minimal or shortest reduct.

If $l \neq 0$, assume $s = \min reduct - reduct \neq \min reduct,$ then $s \bigcup reduct$ contains one (it is assured by definition) and only one reduct that includes $s$, which is a minimal reduct.

It can be found efficiently by algorithm TWI-SQUEEZE.

If not, assume there exists another reduct which includes $s$, it must include some attribute, e.g. attribute b, that is not in the minimal reduct that is contained in $s \bigcup reduct$. From definition of reduct, attribute b can offer demarcations that cannot be offered by both $s$ and $reduct - \min reduct.$ Therefore, if such reduct exists, the minimal reduct is not a reduct. Contradiction.

If $l \neq 0,$ and $s = \min reduct - reduct = \min reduct,$ then $s$ will s-replace all the attributes in $reduct.$ As a consequence, $s$ will be found by the algorithm as a reduct.

Therefore, if $s$ is used in replacement, the algorithm can find a minimal reduct. And we know, $s$ will be used in replacements in at most $C_{n-m}^{l}$ replacements. Hence, the conclusion is true.

(3) For a given positive integer $a,$ assume there is a reduct, $reduct^{'},$ it satisfies $Card(reduct^{'} - reduct) \le a.$ Let $s = reduct^{'} - reduct.$ Then similar to the proof of (2), there is only one reduct that contains $s$, which can be found efficiently by algorithm TWI-SQUEEZE.

---

[53] But generally it will take much more time to know this fact. That is, although we've find a minimal reduct in some times of replacements, we cannot know it without more tests because $l$ is uncertain.

Therefore, in $C_{n-m}^{\min(a,n-m)}$ replacements, we can find any reduct, $reduct^{'}$, that satisfies $Card(reduct^{'} - reduct) \le a$. ■

**Remark 3:** Proposition 4 (2) tells us that if some attributes are favorite to users and will be included in reduct definitely, and if there exists a reduct that doesn't includes any such attribute, we can find a minimal attribute set that includes these attributes efficiently. Proposition 4 (3) tells us that we can find a slightly variant reduct of a given reduct efficiently. Replacement is not necessary a s-replacement. But s-replacement is favorable.

After a reduct is obtained by some heuristic algorithm, we can perform s-replacements and obtain a possible shorter reduct. However, it is possible time consuming even when we want a moderate number of new attributes to be inserted into original reduct. Therefore, it can find a minimal reduct only when the input reduct is sufficiently similar with respect to the minimal reduct.

## 6. Knowledge Reduction: a Complete Version

### 6.1 Find an Irreducible System---a Complete Knowledge Reduction

A complete knowledge reduction is composed of attribute reduction and value reduction. There are many works in attribute reduction. On the contrary, there are fewer results in value reduction. Wang (2001) collected several representative algorithms in his monograph.

Fortunately, there already exists a very efficient method for this problem. The secret hides just in the algorithm TWI-SQUEEZE!

Assume we have obtained a reduct. But there still possibly exists some redundant values in the table. It is caused by that the demarcations induced by the left attributes not only offer the necessary demarcations that cannot be offered by the right attributes, but also offer some demarcations that can be offered by the right ones, which will made some of the values of right ones repetitious.

One solution is that we perform L2R-SQUEEZE once more, as shown in Table 8 (a), replacing some values with "*" and removing repetitious objects (by sorting). The table dealed with is just the same as Table 2(a). Afterwards, perform algorithm R2L-SQUEEZE the second time. In the process, we only retain the values appearing in cascade task intersection and replace all the other values with "*". This process can not only ensure that the values of $C_i$ offer all the necessary demarcations that cannot offer by the values of attributes on its both sides, but most of all it can also retain all the necessary demarcations for the whole system. It deserves to say again that we must replace the values not in cascade task intersection with "*" layer by layer in the running of R2L-SQUEEZE, which is different from attribute reduction. Otherwise new inconsistency will possibly occur if we only retain the values in cascade task intersection. I will give an example to illustrate how to perform value reduction.

**Remark 4**: In the process, we obtain $sT_i$ $(i \in [1 \cdots length(reduct)])$. Replace all the values that are on the left side of $sT_i$ with "*", and we obtain a reduction, $LS$, of the original table.

**Remark 5**: In Table 8 (a), the "Ad" is rearranged, hence we use segment instead of normal set to represent the basic elements in task tree. But when we perform algorithm R2L-SQUEEZE, we use set instead of segment because "*"s are in the table and sortings make the numbers in array Ad disordered.

| Ad | $C_6$ | $C_5$ | $C_2$ | D |
|---|---|---|---|---|
| 1 | 0 | 0 | 0 | 0 |
| 2 | 0 | 0 | 0 | 0 |
| 3 | * | 1 | 1 | 0 |
| 4 | * | 1 | 1 | 0 |
| 5 | * | 1 | 1 | 0 |
| 6 | * | 1 | 1 | 0 |
| 7 | 1 | 0 | 0 | 1 |
| 8 | 2 | 0 | 0 | 1 |
| 9 | 1 | 1 | 0 | 1 |
| 10 | 1 | 1 | 0 | 1 |
| 11 | 1 | 0 | 1 | 1 |
| 12 | 2 | 0 | 1 | 1 |
| 13 | * | * | 2 | 1 |
| 14 | * | * | 2 | 1 |
| 15 | 0 | 1 | 0 | 2 |
| 16 | 0 | 1 | 0 | 2 |
| 17 | 0 | 0 | 1 | 2 |
| 18 | 0 | 0 | 1 | 2 |

Table 8 (a)

| Ad | $C_2$ | $C_5$ | $C_6$ | D |
|---|---|---|---|---|
| 1(2) | 0 | 0 | 0 | 0 |
| 3(4) | 1 | 1 | * | 0 |
| 7 | 0 | 0 | 1 | 1 |
| 11 | 1 | 0 | 1 | 1 |
| 9(2) | 0 | 1 | 1 | 1 |
| 8 | 0 | 0 | 2 | 1 |
| 12 | 1 | 0 | 2 | 1 |
| 13(2) | 2 | * | * | 1 |
| 17(2) | 1 | 0 | 0 | 2 |
| 15(2) | 0 | 1 | 0 | 2 |

Table 8 (b)

**Remark 6**: In this example, we remove repetitious objects when possible, to illustrate the idea clearly. But effect of this step needs to be verified in practice.

After finding a reduct {$C_2$, $C_5$, $C_6$}, we perform algorithm L2R-SQUEEZE the second time. And we obtain new cascade tasks as shown below:

$IT_2=\{T_{20}:[1,2]_0,[7,10]_1,[15,16]_2;T_{21}:[3,6]_0,[11,12]_1,[17,18]_2\}$.

$IT_5=\{T_{20,50}:[1,2]_0,[7,8]_1;T_{20,51}:[9,10]_1,[15,16]_2;T_{21,50}:[11,12]_1,[17,18]_2\}$. $IT_6=\phi$.

Remove repetitious objects. The result is shown in Table 8 (b). For object 13, "$C_2$=2" is sufficient for this rule. Afterwards, perform algorithm R2L-SQUEEZE the second time, and we can obtain new cascade tasks.

$IT_6^{'}=\{T_{60}^{'}:\{1,3\}_0,\{13\}_1,\{17,15\}_2;T_{61}^{'}:\{3\}_0,\{7,11,9,13\}_1;T_{62}^{'}:\{3\}_0,\{8,12,13\}_1\}$.

$IT_2\overset{+}{\cap}IT_6^{'}=\{\{1\}_0,\{15\}_2;\{3\}_0,\{17\}_2;\{3\}_0,\{11\}_1;\{3\}_0,\{12\}_1\}$. As shown in Table 8(c), the core values of $C_5$ are colored with red. And in Table 8 (d), other values, except core values, are replaced with "*".

$IT_5^{'}=\{T_{60,50}^{'}:\{1\}_0,\{13\}_1,\{17\}_2;T_{60,51}^{'}:\{3\}_0,\{13\}_1,\{15\}_2;T_{61,51}^{'}:\{3\}_0,\{7,9,13\}_1;T_{62,51}^{'}:\{3\}_0,\{8,13\}_1\}$.

$IT_2^{'}=\phi$. As shown in Table 8 (e), the core values of $C_2$ are colored with red. And in Table 8 (f), other values, except core values, are replaced with "*".

At last, we need remove repetitious values of the reduced table. It is shown in Table 8 (g).

| Ad | $C_2$ | $C_5$ | $C_6$ | D |
|---|---|---|---|---|
| 1(2) | 0 | 0 | 0 | 0 |
| 3(4) | 1 | 1 | * | 0 |
| 7 | 0 | 0 | 1 | 1 |
| 11 | 1 | 0 | 1 | 1 |
| 9(2) | 0 | 1 | 1 | 1 |
| 8 | 0 | 0 | 2 | 1 |
| 12 | 1 | 0 | 2 | 1 |
| 13(2) | 2 | * | * | 1 |
| 17(2) | 1 | 0 | 0 | 2 |
| 15(2) | 0 | 1 | 0 | 2 |

Table 8 (c)

**Remark 7:** *Red color is the core values in cascade task intersection.*

| Ad | $C_2$ | $C_5$ | $C_6$ | D |
|---|---|---|---|---|
| 1(2) | 0 | 0 | 0 | 0 |
| 3(4) | 1 | 1 | * | 0 |
| 7 | 0 | * | 1 | 1 |
| 11 | 1 | 0 | 1 | 1 |
| 9(2) | 0 | * | 1 | 1 |
| 8 | 0 | * | 2 | 1 |
| 12 | 1 | 0 | 2 | 1 |
| 13(2) | 2 | * | * | 1 |
| 17(2) | 1 | 0 | 0 | 2 |
| 15(2) | 0 | 1 | 0 | 2 |

Table 8 (d)

| Ad | $C_2$ | $C_5$ | $C_6$ | D |
|---|---|---|---|---|
| 1(2) | 0 | 0 | 0 | 0 |
| 3(4) | 1 | 1 | * | 0 |
| 7 | 0 | * | 1 | 1 |
| 11 | 1 | 0 | 1 | 1 |
| 9(2) | 0 | * | 1 | 1 |
| 8 | 0 | * | 2 | 1 |
| 12 | 1 | 0 | 2 | 1 |
| 13(2) | 2 | * | * | 1 |
| 17(2) | 1 | 0 | 0 | 2 |
| 15(2) | 0 | 1 | 0 | 2 |

Table 8 (e)

| Ad | $C_2$ | $C_5$ | $C_6$ | D |
|---|---|---|---|---|
| 1(2) | 0 | 0 | 0 | 0 |
| 3(4) | 1 | 1 | * | 0 |
| 7 | 0 | * | 1 | 1 |
| 11 | * | 0 | 1 | 1 |
| 9(2) | 0 | * | 1 | 1 |
| 8 | 0 | * | 2 | 1 |
| 12 | * | 0 | 2 | 1 |
| 13(2) | 2 | * | * | 1 |
| 17(2) | 1 | 0 | 0 | 2 |
| 15(2) | 0 | 1 | 0 | 2 |

Table 8 (f)

| Ad | $C_2$ | $C_5$ | $C_6$ | D |
|---|---|---|---|---|
| 1(2) | 0 | 0 | 0 | 0 |
| 3(4) | 1 | 1 | * | 0 |
| 7(3) | 0 | * | 1 | 1 |
| 11 | * | 0 | 1 | 1 |
| 8 | 0 | * | 2 | 1 |
| 12 | * | 0 | 2 | 1 |
| 13(2) | 2 | * | * | 1 |
| 17(2) | 1 | 0 | 0 | 2 |
| 15(2) | 0 | 1 | 0 | 2 |

Table 8 (g)

Table 8 (g) can be rewritten as a rule system[54]:

If $(C_2=0)\wedge(C_5=0)\wedge(C_6=0)$ Then $D=0$.(this rule match 2 cases in the original table)

If $(C_2=1)\wedge(C_5=1)$ Then $D=0$.(this rule match 4 cases in the original table)

……,

If $(C_2=0)\wedge(C_5=1)\wedge(C_6=0)$ Then $D=2$.(this rule match 2 cases in the original table)

The goal of knowledge reduction is to find a compact system which can not be further reduced. Below is an outline of an algorithm for knowledge reduction.

Input: A decision table $KS=(U,C\cup D)$, which is described before.

Output: An irreducible decision table[55] *mLS* =(*mL*, *reduct* $\cup$ *D*) (*mL* $\subseteq$ *U*). Some values in *mLS* are "*".

Step1. Perform algorithm TWI-SQUEEZE, remove repetitious objects by sorting and obtain a reduced decision table *KS'*=(*U'*, *reduct* $\cup$ *D*).

(Step1 is a process of attribute reduction)

Step2. Perform L2R-SQUEEZE.[56]

Some values in the objects, which appear in $sT_i$, are redundant for distinguishing the objects[57]. Therefore, replace these values with "*".

Step3. Remove repetitious objects.

Step4. Perform R2L-SQUEEZE.

In the process, we only retain the values appearing in cascade task intersection and replace all other values by "*" layer by layer.

Step5. Remove repetitious objects. At last we obtain a minimal reduced decision table of the original one, which corresponds to a rule system.

**Proposition 5**: *mLS*, which is the result of above knowledge reduction, can not further be reduced.[58]

**Proof Sketch:** First, the attribute set cannot further be reduced. It is ensured by the definition of reduct. Second, this kind of reduction won't lose necessary classification information, or relative demarcations. Before value reduction, all necessary demarcations are at hand, which is ensured by the proofs in appendix A. The algorithm also ensures all relative demarcations be retained in the process of value reduction. A value can be substituted by "*" only when it won't cause loss of relative demarcations. Third, and the last, every attribute and its corresponding value in a rule (or object) cannot further be removed because each one appears in at least one cascade task intersections, which means the removing will surely cause new

---

[54] See section 7.8.3 for other kind of rules that maybe draw from reduced table.

Although some researchers classify rules as characteristic rule, discrimination rule, classification rule, and decision rules (Han, 1992, Hu, 1995), these rules are same from the viewpoint of algorithm and demarcation information. In other words, they can be extracted from decision table using the same algorithm.

[55] Note that, here "irreducible" is limited to original data representation. That is, no new representation of values is permitted.

[56] However, we can also perform R2L-SQUEEZE first. It doesn't matter if we have no bias on these attributes. But if the right side attributes are more important for a user, performing R2L-SQUEEZE first is more preferable.

[57] Their corresponding attributes are on the right side of $C_i$, i.e. $C_{i+1}$, $C_{i+2}$, etc.

[58] Here Assume decision attributes need not be reduced.

inconsistency in the table. Therefore, the proposition is true. And *mLS* is irreducible. ■

There is no "water" (redundancy) in a rule system reduced by the amended version of algorithm TWI-SQUEEZE. Let $T(mLS)$ be the amount of values retained in *mLS*. Then $R = \frac{T(mLS)}{T(LS)}$ can be called compression ratio of knowledge reduction. However, although *mLS* is irreducible, it is very possible that *mLS* is not a minimal irreducible rule system. Finding a minimal irreducible rule system is a NP-hard problem.

### 6.2 Deal with Incompleteness Efficiently

As we have seen that in an incomplete table created in value reduction, there are many redundant values in cascade tasks, which significantly improved computational complexity, especially space complexity. Figure 5(a) show it clearly by color the objects with red, which have a "*" value under certain attribute.

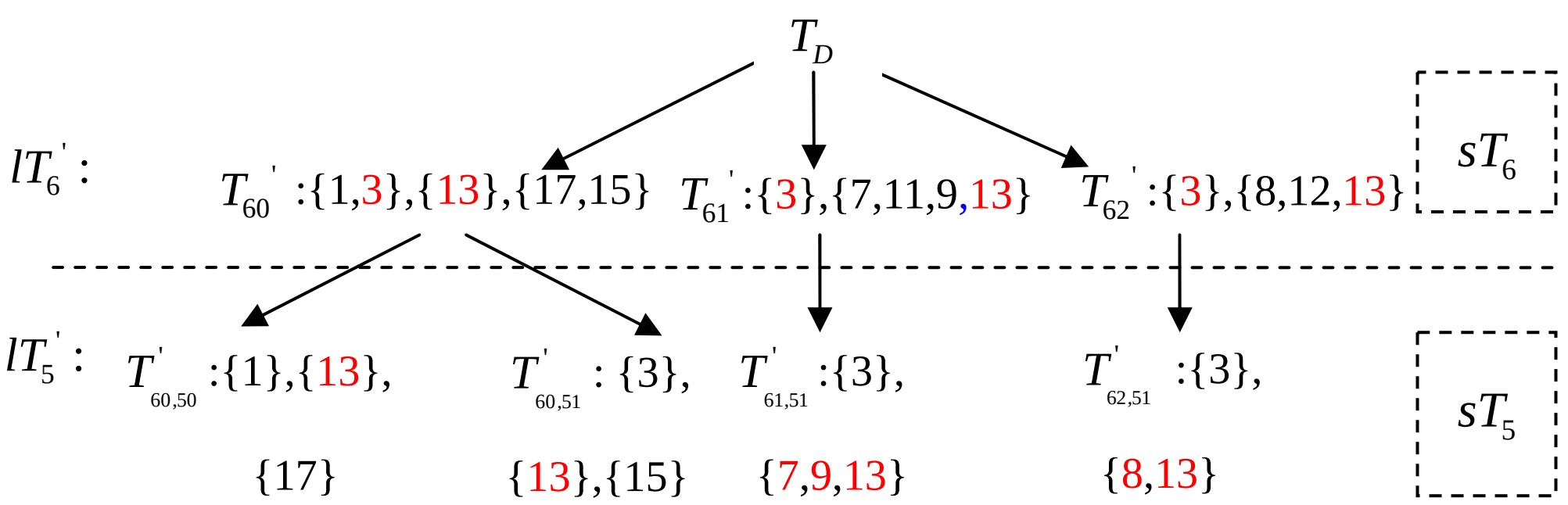

Figure 5 (a): Task Tree produced in the 2rd R2L-SQUEEZE

**Remark 8:** In Figure 5 (a), $sT_6 = \phi$. $sT_5 = \{11,12\}$. $sT_2$ includes all the objects.

Therefore, we need a better solution, which is shown in Figure 5 (b).

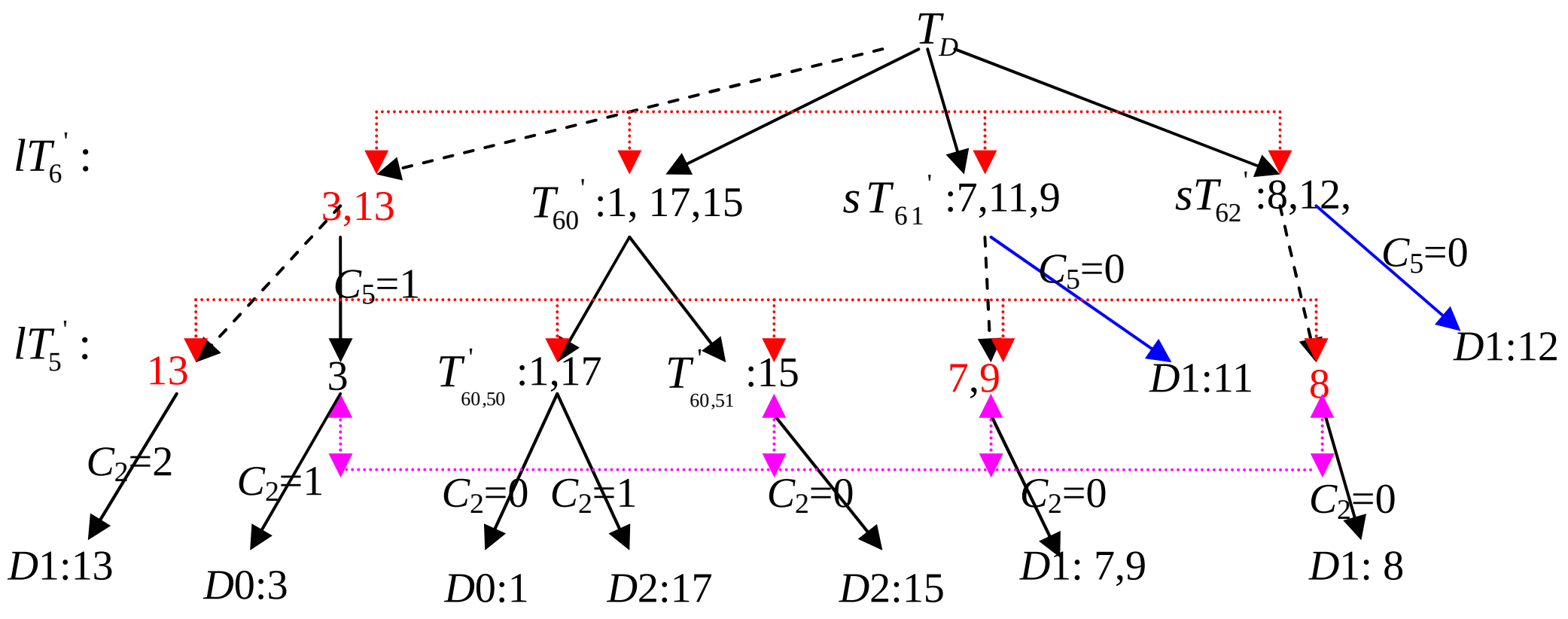

Figure 5 (b): Task Tree produced in the 2rd R2L-SQUEEZE

The main difference between the new solution and the old method is that:

1. Objects with "*" in a task is separated from other objects, and form a set. We can call it star set (*set). All the objects in *set are further grouped according to their D-Region values.

2. This time the semi-task $sT_i$ never combines with the objects in star set to form a task. Therefore, in the old method, $sT_6=\phi$ because the objects in it have combined with the objects in star set to form tasks. And in new method, $sT_6=\left\{sT_{61}^{'}:\{7,11,9\},sT_{62}^{'}:\{8,12\}\right\}$.
3. An "established task" is the task without "*" elements in it. In contrast, a semi-task can also be called "latent task", which is also without "*".
4. A task in $lsT_i$ is divided into three parts. One is star set (left side in Figure 5 (b)); one is established task (middle side in Figure); one is semi-tasks (right side in Figure), which is a complement to tasks in the same level with respect to their father task. Note that in the original algorithm TWI-SQUEEZE, single segments are all grouped together to form one semi-tasks $sT_i$. This time, they are grouped according to their D-Region values.
5. Some pointers are needed to earmark the areas which corresponding elements of *set can influence.

New method follows below rules, which are a bit different from the original value reduction method, but the original main framework is remained:

1. The propagation of *set and semi-tasks are independent to established task. They propagate in the same way as decision tree. But any element of *set need to keep track of the areas it can influence by some pointers, as the red and pink pointers in Figure 5 (b).
2. Objects in semi-tasks can be removed away forever if and only if there is no element in *set that has a different decision value. E.g., object 11, 12 are cast off in Figure 5 (b).
3. When forming cascade task intersection, we first match the objects in *set, and search the areas which are influenced by these objects.

For example, $lsT_6^{'}=\left\{*:\{3\}_0,\{13\}_1;T_{60}^{'}:\{1\}_0,\{17,15\}_2;sT_{61}^{'}:\{7,11,9\}_1;T_{62}^{'}:\{8,12\}_1\right\}$. And $lT_2=\left\{T_{20}:[1,2]_0,[7,10]_1,[15,16]_2;T_{21}:[3,6]_0,[11,12]_1,[17,18]_2\right\}$. In $lsT_6^{'}$, the areas influenced by *set are $\left\{T_{60}^{'}:\{1\}_0,\{17,15\}_2;sT_{61}^{'}:\{7,11,9\}_1;T_{62}^{'}:\{8,12\}_1\right\}$. Because in $lT_2$, object 3 and $[11,12]_1,[17,18]_2$ form a task, and object 3 is in *set of $lsT_6^{'}$, we search $\left\{T_{60}^{'}:\{1\}_0,\{17,15\}_2;sT_{61}^{'}:\{7,11,9\}_1;T_{62}^{'}:\{8,12\}_1\right\}$ and find object 11,12,17 in these areas. Therefore, these objects should be retained in table. Afterwards, we compute $lT_2\overset{+}{\cap}T_{60}^{'}$, object 1 and 15 are matched. Therefore, $lT_2\overset{+}{\cap}lT_6^{'}=\left\{\{1\}_0,\{15\}_2;\{3\}_0,\{17\}_2;\{3\}_0,\{11\}_1;\{3\}_0,\{12\}_1\right\}$.

Note that this time we can compute cascade task intersection directly from the task tree shown in Figure 5 (b).

The time complexity of value reduction is $O(t_C N\ln N)$; and space complexity $O(N)$.

### 6.3 Computational Complexity of Knowledge Reduction

To perform a complete knowledge reduction, we need four scans of database. Using the techniques in section 6.2, the worst time complexity of this algorithm is $O(t_C N\ln N)$ and total space complexity $O(N)$.

## 6.4 Incomplete and Approximation Value Reduction

In section 6.1 and 6.2, two complete value reduction algorithms are proposed. After complete value reduction, no value can be further removed without loss of accuracy of classification. In this section, we discuss about incomplete value reduction and approximation value reduction.

Incomplete value reduction is also a reduction that can ensure the accuracy of classification. But it is still possible that there are some redundant values after reduction. Assume we have performed attribute reduction. To carry out incomplete value reduction, we need only run algorithm L2R-SQUEEZE or R2L-SQUEEZE once more. Replacing some values with "*" and removing repetitious objects (by sorting), we get a reduced table. Table 9 (a) shows the reduced table after perform algorithm R2L-SQUEEZE on table 2 (a) the second time; Table 9 (b) shows the reduced table after perform algorithm L2R-SQUEEZE the second time. Compare with complete value reduction, incomplete value reduction in this way can achieve a sufficient concise result with less computational effort, and as a consequence can be a choice in practical application.

Approximation value reduction is a sort of approximation reduction that probably results in loss of accuracy of classification. Generally, the rule system produced in this way is smaller

| Ad | $C_2$ | $C_5$ | $C_6$ | D |
|---|---|---|---|---|
| 1(2) | 0 | 0 | 0 | 0 |
| 3(2) | 1 | 1 | 0 | 0 |
| 5(2) | 1 | 1 | 1 | 0 |
| 13(2) | 2 | 1 | 0 | 1 |
| 7(2) | * | 0 | 1 | 1 |
| 9(2) | 0 | 1 | 1 | 1 |
| 8(2) | * | * | 2 | 1 |
| 17(2) | 1 | 0 | 0 | 2 |
| 15(2) | 0 | 1 | 0 | 2 |

Table 9 (a): after 2nd R2L-SQUEEZE

| Ad | $C_2$ | $C_5$ | $C_6$ | D |
|---|---|---|---|---|
| 1(2) | 0 | 0 | 0 | 0 |
| 3(4) | 1 | 1 | * | 0 |
| 7(1) | 0 | 0 | 1 | 1 |
| 8(1) | 0 | 0 | 2 | 1 |
| 9(2) | 0 | 1 | 1 | 1 |
| 11(1) | 1 | 0 | 1 | 1 |
| 12(1) | 1 | 0 | 2 | 1 |
| 13(2) | 2 | * | * | 1 |
| 15(2) | 0 | 1 | 0 | 2 |
| 17(2) | 1 | 0 | 0 | 2 |

Table 9 (b): after 2nd L2R-SQUEEZE

| Ad | $C_6$ | $C_5$ | $C_2$ | D |
|---|---|---|---|---|
| 1 | 0 | 0 | 0 | 0 |
| 3 | | 1 | 1 | 0 |
| 13 | | | 2 | 1 |
| 7 | 1 | 0 | | 1 |
| 9 | 1 | 1 | 0 | 1 |
| 8 | 2 | | | 1 |
| 17 | 0 | 0 | 1 | 2 |
| 15 | 0 | 1 | 0 | 2 |

Table 9 (c): An instance of approximation value reduction than other kinds of reduction. For example, after attribution reduction on table 2 (a), we first perform incomplete value reduction by algorithm L2R-SQUEEZE, and then followed by incomplete value reduction by algorithm R2L-SQUEEZE. At last, we obtain a reduced table as shown in Table 9 (c).

Good approximation value reduction can subdue noise in the original table and therefore can be regarded as a way to avoid overfitting.

### 6.5 Find Core Values

Core value is the value that cannot be omitted in the table. If it is absent after value reduction, it will absolutely cause new inconsistence. One of the significance of finding core value is that core values are sensitive to noise, which means that a reduction is robust only if these values are ensure to be exact enough. Therefore, if we want to be assured about the robust of reduced rule system or classifier, we should check these values carefully. One problem rise, how can we find core values efficiently? Or can we find them with little sweat?

The answer lies in attribute reduction by algorithm TWI-SQUEEZE. In the process of attribute reduction, we will compute cascade task intersection. It is interesting that the core values are just the values appearing in cascade task intersections, which is ensured by proposition 6. Therefore, when we find a reduct, we also find all the core values in the same time, the distribution of which is also an interest itself.

**Proposition 6:** All core values are just the values appearing in cascade task intersections produced in the process of attribute reduction.

**Proof Sketch:** First, the values appearing in cascade task intersections are core values. It is because the values of certain attribute appearing in cascade task intersections are the values that are used to offer some necessary relative demarcations, which cannot be offered by other attributes. Therefore, in an object, loss of such value will definitely cause new inconsistency in classification, which means any such value is a core value.

Second, if a value is a core value, it must appear in some cascade task intersection. A value is a core value in an object if and only if removal of it will cause inconsistency between this object and some other objects. It also implies that other attributes cannot offer necessary relative demarcations to discern these objects. Therefore, core values must appear in cascade task intersections, from the definition of cascade task intersection. ■

## 7. Further Discussions

### 7.1. Beyond Decision Table

Reduction of non-decision table, or information system, received less concern in rough set literature (Gou, J., &Ye, D., 2002). However, non-decision table can be transferred to decision table efficiently, which is proved by Yang et al. (2002). After sorted, the table is divided into some equivalence classes. These equivalence classes can be regarded as decision classes and correspond to D-Regions. And concrete values of decision attributes are not important for algorithm TWI-SQUEEZE when it runs on a decision table. We need only know the D-Regions instead. The time complexity of transferring a non-decision table to a decision table is $O(t_C N \ln N)$, and the space complexity is $O(N)$.[59]

[59] In fact, this sorting can be utilized by the latter reduction.

### 7.2 User Preference

There are possibly many reducts in a decision table. Different user will favor different reduct because the significance of an attribute is different for different people, or only because the prices in obtaining values are varied much. It is possible that some prices cannot be accepted by one user can be easily accepted by another. A user will possibly try to keep the attributes he deems as the most important in a "reduct" even if it will impair the strict definition of "reduct". The crucial point is that everyone has his rational preference, and a good algorithm should be able to take this kind of subjective will into consideration.

Indeed, this problem is easy to be tackled. In rough set theory, the special attributes that is the user's preference can be put into "reduct" firstly and directly (Hu, Pao, Yu, 2002). As for algorithm TWI-SQUEEZE, one straightforward strategy is that these special attributes should be move to the rightmost side of attribute list, similar to Wang's algorithm (2001). Combined with á-priori, the algorithm can find a better reduct in shorter time.

However, this strategy cannot ensure the users to find a small reduct. Can the user find a small attribute set that includes the required attributes? As has proved in section 5.3, if some attributes, a set $s$, are favorite to users and will be included in reduct definitely, and if there exists a reduct, *reduct*, we can find a minimal attribute subset of $s \cup reduct$ that includes these attributes very efficiently, that is in $O(t_C N\log N)$. Hence, in this case, there exists a better strategy: first find a arbitrary reduct (the longer the better) that excludes favorite attributes as possible as it could, then use the favourite attributes to renew (s-replacement is not necessary) the reduct. In this way, we can find a user favourite and at the same time local minimal attribute set very efficiently.

### 7.3 Approximate Reduction

#### 7.3.1 Variable Precision Model: Non-monotonicity and A Solution

W.Ziarko has proposed an extended model of classical rough set theory, which is called "variable precision rough set model" (Ziarko, 1993). It is introduced by Zhang in his monograph (Zhang, W. et al., 2001). Pawlak's rough set model is a special case of Ziarko's variable precision rough set model.

The main difference between classical rough set model and variable precision model is that in Pawlak model a set must belong or not belong to another set definitely, but in variable precision model a set can belong to another set to some extent.

$\nu: \mathrm{P}(U)\times \mathrm{P}(U)\times[0,0.5)\to[0,1]$ is a rough inclusion function, which satisfies:

$\nu(X,Y,\beta)=f_\beta(t)$, where $t=Card(X\cap Y)/Card(X)$, for any $\phi\neq X,Y\subseteq U$ and $0\le\beta<0.5$.

Generally,

$$f_\beta(t)=\begin{cases}0 & if \quad 0\le t\le\beta\\ (t-\beta)/(1-2\beta) & if \quad \beta<t<1-\beta\\ 1 & if \quad 1-\beta\le t\le 1\end{cases}$$

Majority inclusion relation can be defined as $Y\overset{\beta}{\supseteq}X \Leftrightarrow f_\beta(t)=1$.

W.Ziarko's model is also based on the most useful equivalence relation.

($U$, $R$) is an approximation space. $U$ is a non-empty, finite set. $R$ is an equivalence relation on *U*. $U/R=\{E_1,E_2,\cdots,E_n\}$ is a partition of *U*.

The $\beta$ lower approximation of a set $X$ is: $\underline{R}_{\beta}X=\bigcup\left\{E\in U/R\mid X\overset{\beta}{\supseteq}E\right\}$.

Let $KS=(U,C\bigcup D)$. $D$ depends on $C$ in a degree $k$ $(0\le k\le 1)$ with precision $1-\beta$ if $k=\gamma(C,D,\beta)=\dfrac{Card(POS(C,D,\beta))}{Card(U)}$, where $POS(C,D,\beta)=\bigcup_{Y\in U/D}\underline{IND(C)}_{\beta}(Y)$ is called a positive region of the partition $U/D$ with respect to $C$ and with precision $1-\beta$, is the set of all elements of *U* that can be classified to blocks of the partition $U/D$ by means of $C$ with precision $1-\beta$. When $\beta=0$, it is Pawlak's classical model.

An approximation reduct or $\beta$ reduct of $C$ with respect to $D$ is a subset of $C$, which can be denoted as $red(C,D,\beta)$. It satisfies:

(1): $\gamma(C,D,\beta)=\gamma(red(C,D,\beta),D,\beta)$;

(2): (1) will not be satisfied if any attribute is removed from $red(C,D,\beta)$.

This definition is based on expected monotonicity of change of positive region in reduction. However, when $\beta$ is larger than 0, reduction is not monotonic. For example, Table 10 (1) is a small table to be reduced with $\beta=0.4$. $C=\{C_1,C_2\}$.

| | $C_1$ | $C_2$ | D |
|---|---|---|---|
| 1 | 0 | 1 | 0 |
| 2 | 1 | 1 | 0 |
| 3 | 1 | 1 | 1 |

Table 10 (1)

$POS(C,D,\beta)=\{1\}$. However, $POS(\{C_2\},D,\beta)=\{1,2,3\}$. It seems removing of attributes can enlarge positive region. Therefore, the definition cannot tell us how to find reducts rightly although it is clear that in this small instance the reduct should be $\{C_1\}$.

How can we do something about it? In fact, the nature of variable precision model is to offer a mechanism to deal with boundary region. We can use a metaphor to disclose its implication. In voting, different people perhaps have different preference to an objective or decision. Opinion of majority has priority in decision, a popular way that can be called majority rule. When only a handful people below a given ratio hold a different opinion, their decision will be ignored. Minority will have to change either their decision or their topic.

Variable precision model works just like that. Because the number of condition attributes is generally much larger than decision attributes, errors occur in condition attribute values are more possible than errors occur in decision attribute values. Therefore, we *can* change some values in condition attribute values to keep majority rule. Sometimes we *have to* change condition attribute value temporary instead of decision attributes values, especially in tolerance relation based model. For example, as shown in Figure 6, three tolerance sets are formed by seven objects in a table, using labels to denote them are *I* (1)={1, 2, 3, 4, 6, 7}, *I* (2) ={1, 2, 5} and *I* (5)= {3, 4, 5, 6, 7} respectively. That is, objects 2, 3, 4, 6, 7 are similar to object 1, and objects 1, 5 are similar to object 2, and objects 3, 4, 6, 7 are similar to object 5. If $\beta = 0.4$, what decision value should objects 1 and 2 take? And what side should object 5 take? It seems majority rule will be definitely in dilemma. The sole way out seems either to announce *I* (x) is not reasonable or to change condition attribute values of some objects temporary.

However, even if *I* (x) is reasonable, it still will cause problem.

As shown in Figure 7, object set *I* (1)={1, 2, 3, 4} is tolerance set with respect to attribute set {$C_1$, $C_2$} and *I* (1)={1, 2, 3, 4, 5} with respect to attribute set {$C_2$}. If $\beta = 0.4$, object 1 belongs to positive region with respect to {$C_2$}, but doesn't belong to positive region with respect to {$C_1$, $C_2$}. It means non-monotonicity is a problem that lies in the attribute reduction, using the original definitions.

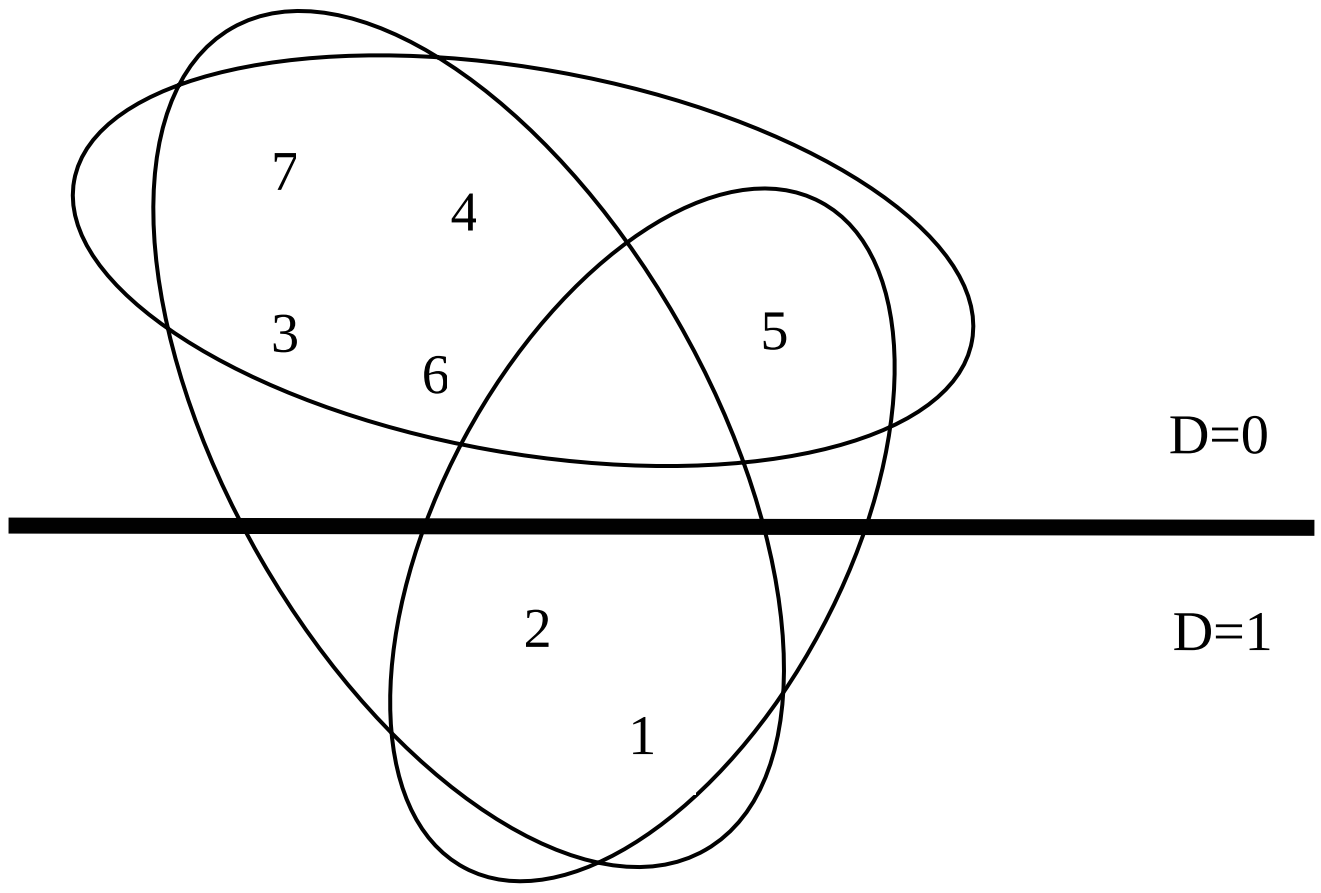

Figure 6

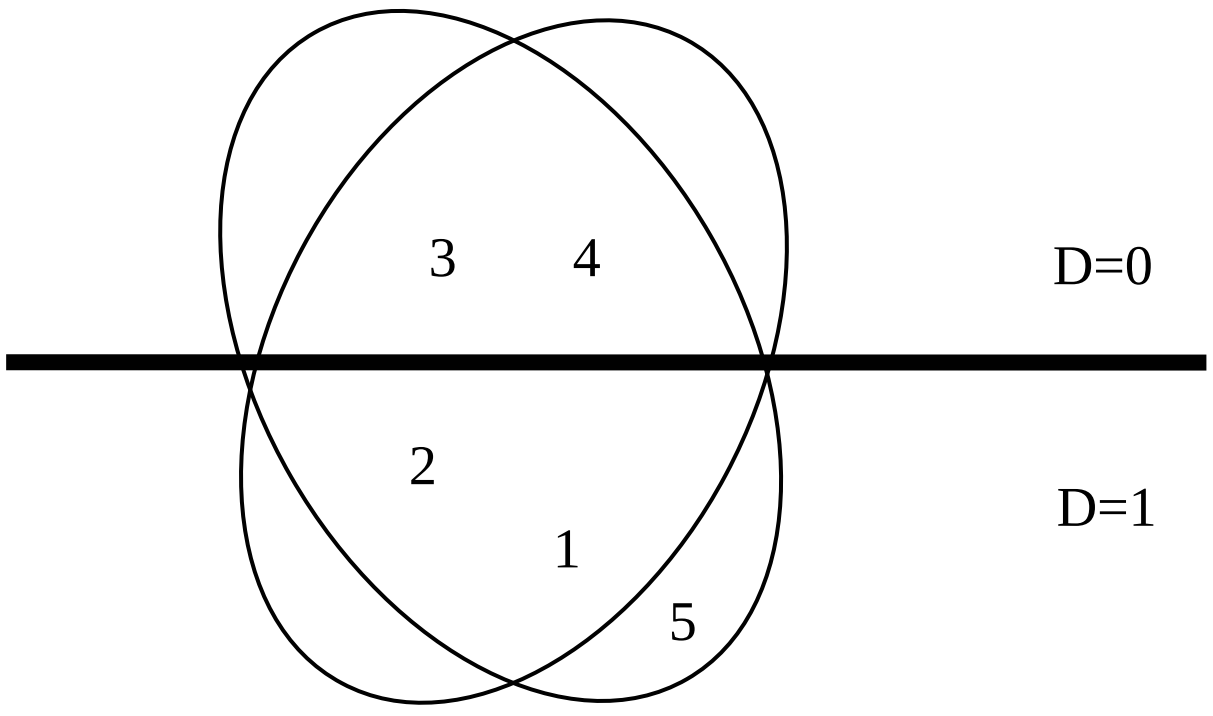

Figure 7

Therefore, we have to adapt the latter choice. That is to change condition attribute values of some objects temporary. But sometimes we can also change decision values.

In the case of Table 10 (1), the table can first be changed to Table 10 (2). Therefore, we have a reduct, which is $\{C_2\}$ now.

| | $C_1$ | $C_2$ | D |
|---|---|---|---|
| 1 | 0 | 1 | 0 |
| 2 | 1 | 1 | 0 |
| 3 | 1 | 0 | 1 |

Table 10 (2)

The reduced table is Table 10 (3).

| | $C_2$ | D |
|---|---|---|
| 1 | 1 | 0 |
| 3 | 0 | 1 |

Table 10 (3)

These rules, as the result of reduction, can classify objects in ***present*** table with some precision.

Note that, the value of $C_2$ of object 3 should be recovered after reduction.

After change some values temporarily, variable precision model can be transformed to Pawlak's rough set model, reduction problem of which can be tackled safely by algorithm TWI-SQUEEZE.

In this paper, algorithms and ideas have been proposed to extract knowledge from decision table. An important topic is how to deal with boundary region? The methods adopted in this paper are the same as rough set theory, i.e. simply separate the objects in boundary region from that in positive region, or use majority rule that underlies variable precision rough set model. However, it is possible to use other ideas to treat it in a more elaborate way. In this paper, we have assumed that all the necessary attributes are at hand in analysis, and therefore remove redundant objects directly. However, information, including attributes themselves, are often incomplete in practical applications. Redundancy is therefore possibly able to tell some truth of data in boundary region. Hence, how to utilize redundancy appropriately and fully in boundary region analysis is still an interesting topic.

### 7.3.2 Graded Rough Set Model

Yao & Lin introduced a graded rough set model in parallel to graded modal logic (Yao, Lin, 1996). Given the universe U and an equivalence relation R on U, a family of graded rough set operators are defined as:

$$\underline{R}_k(A) = \{x \,\|\, R_s(x)| - |A \cap R_s(x)| \le k\}, \overline{R}_k(A) = \{x \,\|\, A \cap R_s(x)| > k\}.$$

Attribute reduction in graded rough set model based on equivalence relation is also very easy. We need only check if there are some short segments in current cascade task, which satisfy that their length are not larger than $k$. If such short segments exist, remove them from current cascade task. Algorithm TWI-SQUEEZE need not change any more and in this model it is especially efficient when *k* is sufficient large, because it can probably find a smaller approximate reduct than a genuine reduct.

However, it should be pointed out that the graded rough set model given by Y.Y.Yao is a more general model in that it is based on a general binary relation. It still needs to discuss the practical meaning and significance of reduction in such binary relation.

### 7.3.3 Other approximation

There are other possible methods that can achieve approximate reduction. However, no matter what method is adopted, one criterion or measure is needed to control the error ratio of classification. Rough set theory has given a criterion function that does the work. It will be discussed in section 7.10.1.

In conclusion, approximation has two purposes. First, it is the necessary way through which we can conquer overfitting. Second, it is also a way to obtain concise rule set. The lost information, which is the result of approximation, can be regarded as exceptions. For example, after attribute reduction on a consistent table based on variable precision model, the exception rules are just the objects in boundary region.

## 7.4 Incomplete System

In a practical application, a decision table is often incomplete, which means many values are missing in the table. In fact, when value reduction is applied on a table, the table will have many values missed. Therefore, the absence of values, which are replaced by “*”, has at least three cases needed to be considered. The first case is that “*” represents all the possible values in a finite domain. In fact, the semantics of this case is not “missing”, but is “all”. In the second case “*” only represent a part of the possible values in the domain. In the third case we have no idea with the “*” except that it represent an unknown value. It can be the result of high price in obtaining the values, or an out-of-range value or simply the values cannot be measured anymore. It seems that different cases need different strategies to handle when potential inconsistence emerges.

A method based on rough set theory was put forward by Kryszkiewicz to tackle incomplete problem (1998). Instead of equivalence relation, a binary relation so called “similarity relation” is the footstone of the algorithm, combined with discernibility matrix and discernibility function. The main problem of this method is also the main problem of all the algorithms that use discernibility matrix. That is, when the table is large, the computational complexity rises quickly to an unacceptable degree.

In fact, algorithm TWI-SQUEEZE can handle incomplete system with little adaptation, as long as we adopt the idea temporarily that “*” cannot be discerned from any other values. The process of value reduction discussed before is a paradigm that deals with incomplete system.
It is possible that at last some objects cannot be discerned by any subset of condition attribute set, and some objects are inconsistent clearly. If some objects are definitely inconsistent, it means there are no missing values in these objects and they can be taken out and handled in a special way. If there are some missing values in these objects, we have to adopt different strategies to handle different cases as have been mentioned. In the first semantics case, the objects that cannot be discerned are also these objects that are inconsistent. Therefore, it can be handled in the same way as the case there is no missing values. Kryszkiewicz regards the semantics of missing values in this way too (1998). In the second case, there are two possibilities. One is that there is no inconsistency; the other is on the contrary. We can simply take both possibilities into account. For the third case, we can also have two attitudes as the second case. In fact, all these three cases can be reduced to two cases: the objects that cannot be discerned are regarded to be consistent or not. For example, there are two objects that contain potential inconsistence, assuming the domain of attribute $C_1$ is {0, 1, 2}. These objects can be written in the form of rules:

Object 1: (If) $C_1$=0 and $C_2$=0 and $C_3$=1 (Then) $D$=0.

Object 2: (If) $C_1$=* and $C_2$=0 and $C_3$=1 (Then) $D$=1.

If object 1 and 2 are regarded to be inconsistent, they can be simply taken out from the original decision table. Otherwise, we can use object 1' and 2' to replace object 1 and 2:

Object 1': (If) $C_1$=0 and $C_2$=0 and $C_3$=1 (Then) $D$=0.

Object 2': (If) $C_1 \neq 0$ and $C_2$=0 and $C_3$=1 (Then) $D$=1.

Note that in this case, "$\neq$" is imported as building brick of rules. Hence, the development of the original table has two histories now. When new data are added, we can choose one history or consider both.

### 7.5 Data Preprocessing

It will never immoderate to stress the importance of data preprocessing, especially for algorithm TWI-SQUEEZE to be effective in practical applications.

Data preprocessing is done by data transformation. Data transformation is important itself. It includes many approaches, such as smoothing, discretization, concept hierarchy generation, etc. Smoothing includes binning, clustering, regression, etc. It offers solutions for conquering noisy data.

Discretiztion is mainly a technique that handles continuous attributes. Discretization is important in reduction not only because it can reduce the number of values for a given continuous attribute, but also because it can remove meaningless demarcations. And it can also help determine what generalization level a rule system will be as a whole. There are many works have done to this problem, including the contributions from rough set community (Komorowski, Pawlak, Polkowski, &Skowron, 1998).

Discretization is also crucial for conquering overfitting. We have to decide what granule we are comparing on or in other words, we have to choose a criterion to decide that two values are really different or not. If we fail to do so, it is very possible that some attribute happens to partition the examples, despite being unrelated to the actual target function. That is, not all differences are helpful in learning. How to discern them will be one of my future works.

When background concept hierarchies are at hand, we can achieve better reduction by concept hierarchy generation. It is a necessary step before we perform reduction. It is well known that basic level concept is more preferable to superordinate level and subordinate level concepts to describe an event (Best, 1998). However, superordinate level concepts are generally preferred in data mining and knowledge discovery literature[60]. The goal of knowledge discovery is to find short rules which have strong generalization. But, we should be more cautious before adopting this idea. When concepts are generalized, many demarcations are also lost in this reduction process. It means longer reducts (though fewer rules) tend to be found, which seems to contradict with intuition. It implies either more complex combinations of attributes, or, more and even meaningless rules are needed to represent function dependency if concept level is not selected properly.

Concept hierarchies can be provided by knowledge engineers or domain experts, and sometimes can be discovered automatically or semi-automatically. The method is called attribute-oriented induction, which is brought forwarded or introduced by Yandong Cai et al.(Cai, Cercone, Han, 1991), Jiawei Han et al. (Han, Cai, Cercone, 1992; Han, 2000), Xiaohua Hu et al (Hu, Shan, Cercone, Ziarko, 1994; Hu, 1995, Hu, 2003). The key to their approach is the attribute-oriented concept tree ascension technique, which is useful for any reduction algorithm.

Therefore, discretizatoin and generalization of values is very important for knowledge reduction by algorithm TWI-SQUEEZE. In many cases, the results of reduction will be very bad without proper preprocessing of data, especially when there are some continuous-valued attributes.

On the other contrary, algorithm TWI-SQUEEZE can build in these data transform technologies. Difference is relative to the resolution or coarseness in observation and what aspects compare is based on. Two objects are different in high resolution or some aspects whereas equal in lower resolution or other aspects. Therefore, demarcation based on difference is also relative.

### 7.6 Dynamic Reduct

The thought of dynamic reduct is combining some attribute reduction algorithm with sampling techniques (Bazan, Skowron, Synak, 1994a, 1994b; Bazan, 1996). Generally, this method is computational intensive. However, this kind of method is useful if noise is not ignorable and if our aim is to predict instead of description. In both cases, approximation is the necessary way to lead us to our claim.

The standard means to find dynamic reduct is to apply reduction algorithm on both the whole and parts of the universe. And what we find at last are still reducts of the whole universe. We can also combine the ideas of approximation and sampling. As the increasing large collections of data is used in practical applications, sampling becomes a promising method for data mining or knowledge discovery: instead of doing complicated discovery process on all the data, one first takes a small sample, finds the regularities in it, and then possibly validates these on the whole data (Kivinen, Mannila, 1994). In a word, sampling can not only lessen the prohibitive computation requirement in large database, but also can be adopted to test hypothesis and avoid overfitting. Generally, sampling is performed on universe. But it is also possible that the attributes are not appropriately selected, which will greatly influence the result of reduction. It seems sampling not only on objects but also on attributes deserves consideration. As for algorithm TWI-SQUEEZE, it means reordering of attributes when applied to different samples.

[60] In some cases, such as fraud detection, outliers and abnormal patterns are more of interests.

It is possible that for any $P \subseteq C$ ($C$ is the whole conditional attribute set), which satisfies $\frac{\left|\{B \in F : P \in RED(B,d)\}\right|}{|F|} \geq 1-\varepsilon$ ( $F \subseteq \{K' \mid K' = (U', C \cup D), U' \subseteq U\}$ ), $Card(P)$ is too small to contain a reduct or an approximate one (Bazan, 1996). Therefore, instead of trying to find a (approximate) reduct in this way, we can find an attribute subset that includes most attributes which appear in *RED* (*B*, *d*) above a given frequency $1-\varepsilon$, and in the same time sustains given accuracy of classification. Combined with idea of approximation reduct, sampling techniques will make the algorithm better for prediction, rather than a precise description of current data.

### 7.7 Find More Reducts

Algorithm TWI-SQUEEZE can find a reduct efficiently, which has been discussed before. Can it find more reduct?

It is clear that with replacement (section 5.3) we are probable to find a new reduct and in the same time control the variance to some extent. But it is usually time consuming.

If the reduct found is relative small to the whole condition attribute set, we can move the attributes in reduct to the left end of the attribute list. Run algorithm TWI-SQUEEZE again and we are mostly come across a very different reduct. Furthermore, if the reduct does exist, it is easy to prove that it has maximum different attributes to the original one.

It is easy to find more reduct. The crucial point is what order of the conditional set algorithm TWI-SQUEEZE is running on. We can also control which attributes to be included into the reduct, if they are indispensable to the system by put them in the right side of conditional attribute list.

### 7.8 Closely Relevant Topics

#### 7.8.1 Feature Selection

Finding a reduct is a process of dimensionality reduction, which can lessen the curse of dimensionality. It is closely related to feature selection, which can be regarded as finding a specified length approximate reduct with an acceptable "error of reduct approximation" $\varepsilon_{(C,D)}(B)$ ( Komorowski, Pawlak, Polkowski, &Skowron, 1998). Here, $B$ is a subset of $C$. In rough set theory, $\varepsilon_{(C,D)}(B) = \frac{(\gamma(C,D)-\gamma(B,D))}{\gamma(C,D)} = 1-\frac{\gamma(B,D)}{\gamma(C,D)}$, $\left(\gamma(P,D) = \sum_{X \in U/D} \frac{|\underline{P}(X)|}{|U|}\right)$. It is easy to compute if we only concern about current data because we need only add up the lengths of all the segments in current cascade task $lT$. Afterwards, divide the sum by *Card*(*U*). For example, in Table 2(a), if $B=\{C_5,C_6\}$, $\varepsilon_{(C,D)}(B) = \frac{14}{18}$; if $B=\{C_3,C_4,C_5,C_6\}$, $\varepsilon_{(C,D)}(B) = \frac{6}{18}$.

$\varepsilon_{(C,D)}(B)$ has a good quality. That is for any $B' \subseteq B$, $\varepsilon_{(C,D)}(B) \leq \varepsilon_{(C,D)}(B')$. But it does not mean that for any $B' \subset B$, $\varepsilon_{(C,D)}(B) < \varepsilon_{(C,D)}(B')$.

What we should pay attention to such kind of measurement is that we should partition the data into two parts, which are training set (current data) and test set, using methods such as k-fold cross-validation. Therefore, we need adjust $\varepsilon_{(C,D)}(B)$ when test set is taken into account. Let the original $\varepsilon_{(C,D)}(B) = \frac{a}{b}$ ($b$ is the cardinality of training set), and there are $c$ wrong cases form $d$ test samples, the new $\varepsilon_{(C,D)}(B) = \frac{a+c}{b+d}$.

We can perform FS in a similar way as we perform attribute reduction because they share a similar problem. The algorithm framework introduced in this paper with heuristic information can be regarded as modified sequential suboptimal search methods. But this time the algorithm R2L-SQUEEZE terminates as long as the specified number of attributes are found. The feature selection criterion function is at hand. That is $J(B) = 1 - \varepsilon_{(C,D)}(B)$. After L2R-SQUEEZE is performed, it is possible that some attributes are removed. Then R2L-SQUEEZE will run on the reduced approximate reduct, followed by L2R-SQUEEZE. The process is repeated until the specified number of attributes are found and no attribute is dispensable to other attributes.

This method also suffers from nesting effect. It is a tradeoff between optimality and computational efficiency. Its performance seems between classical sequential suboptimal search methods and sequential floating search methods when effectiveness is concerned, although some experiments should be done to test it.

We can also combine the subprogram of algorithm TWO-SQUEEZE and floating search methods (Pudil, Novovicova, & Kittler, 1994). For example, we can perform SFFS algorithm by applying one step of algorithm R2L-SQUEEZE first, then conditionally exclude one feature (attribute) by applying one step of algorithm L2R-SQUEEZE. Afterwards, examine if the reduced feature set is the best subset so far to determine leaving out the conditionally excluded feature or not.

It is said that adaptive sequential forward floating search method (ASFFS) can find an approximate reduct which is close to the optimal one (Somol, Pudil, Novovicova, & Paclik, 1999). Hence, we can combine ASFFS and replacement introduced in section 5.2 to increase the possibility of finding an optimal one.

### 7.8.2 Classifier

#### 7.8.2.1 Classifier Tree

The reduced table, which can be regarded a rule system[61], is a classifier. From reduced system we can quickly construct a classifier tree which is similar to a decision tree. For example, the

[61] Strictly speaking, there is a gap between reduced table and rule system. However, if we adopt the restriction that any rule has the form “If A=? And B=? Then C=?”, then these two system is identical. There is no transform is needed. But if we want to extend the relational operator to “$\neq$” and “Or”, as has discussed in section 7.8.3, we need to pay attention to the gap.

reduced system of Table 8 (g) can be represented as:

Remember that most rules cover more than one case in the original table. I use one object to represent the other for neater graph.

The classifier tree in Figure 8 is a bit special which need to be explained. Different from decision tree, branches of every level have an order. That is, when we classify an object, we match its condition from left to right in any level. If it matches one branch, it will not search the right hand branches anymore, but go down to continue the match. In Figure 8, every branch has a label like: $C_k=a$. In the Figure, if the $i^{th}$ level branch (the branches ramified by root is the first level branch) has a label like: $C_j=b$, here $C_j$ is the $i^{th}$ attribute in reduct {$C_6$, $C_5$, $C_2$}, we use "*b*" to label the branch for simple. We put such branch at the right side of the $i^{th}$ level. Any match goes from top to down and from left to right.

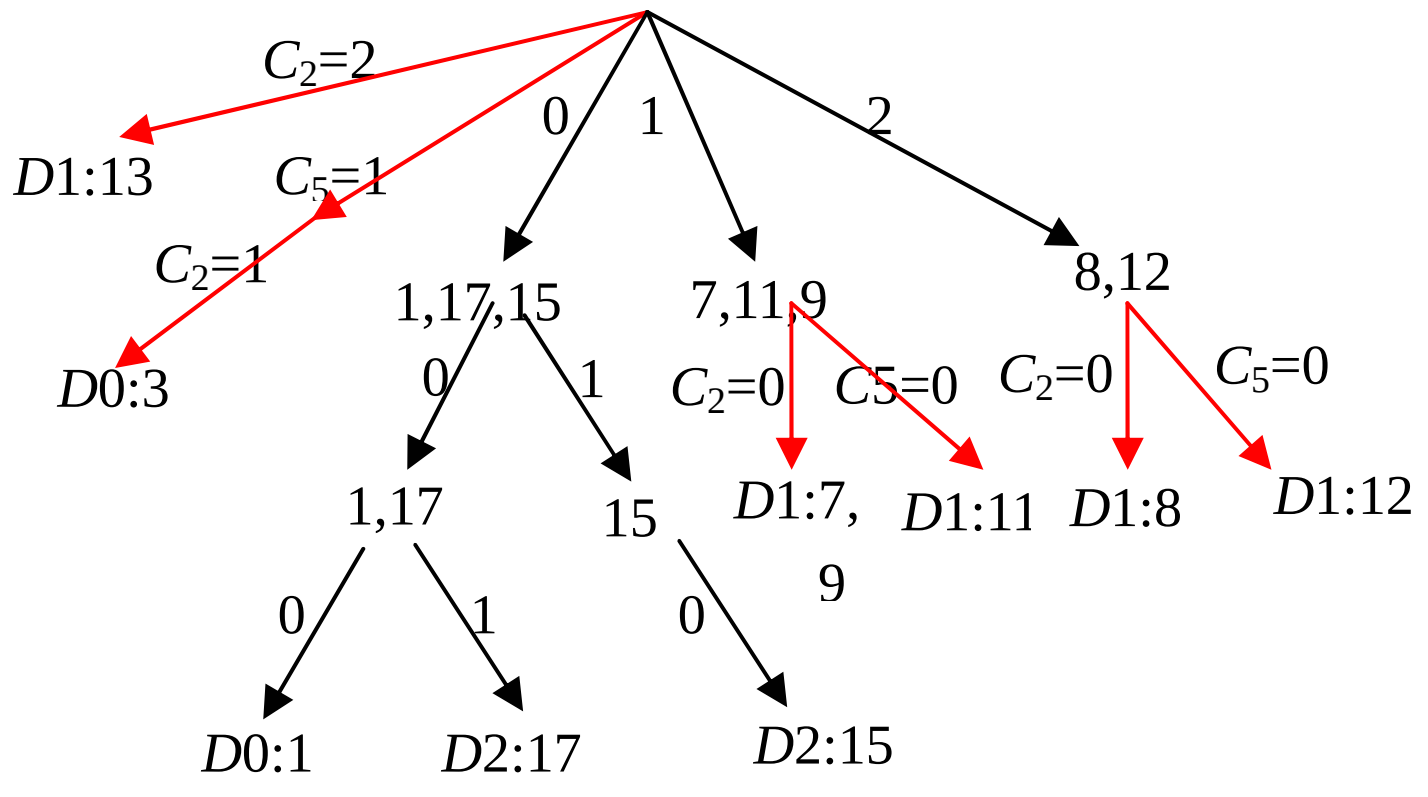

Figure 8: A Classifier Tree

Another difference is that every object to be classified will be assigned a set of all the possible target concepts first. The size of the set will shrink to one after tests of some attributes of the object. For example, in the first level of classifier tree in Figure 8, the node of the third branch is {1, 17, 15}, which has assigned a target concept set: {*D*0, *D*2}. It means that any element of {1, 17, 15} can be classified to *D*0 or *D*2. The rule can be such: If $C_6=0$ Then $D=0$ with a probability of 0.33 or $D=2$ with a probability of 0.67. At last, the possible target set of its grandchild node {1} is shrunk to {*D*0}, which means object 1 should be classified to *D*0 class or concept.

If the reduced table has *n* objects, and *t* attributes, the time complexity of transforming it to classifier tree is $O(nt)$. Note however, we should amend the example a bit: when sorting a column of table, we always put the objects with a "*" value on the top. On the contrary, in the example of Table 8 (a) ~ (g), these objects are put on the bottom.

Although this kind of tree is more complex to be understood than classical decision tree, it solves the problem faced by decision tree (Cendrowska, 1988), which is also introduced by Wang et al (2001). That is "its (decision tree output of ID3) use in expert systems frequently demands irrelevant information to be supplied" (Cendrowska, 1988).

Algorithm L2R-SQUEEZE (the fourth scan) forms a task tree, which is shown in Figure 5 (b). It is interesting that if we ignore the broken line in the tree, it is just the classifier tree shown in Figure 8, which means the task tree is also a classifier.

Therefore, after knowledge reduction, we have two identical classifiers: one is a rule system and the other is a tree. As far as efficiency of classification is concerned, tree is favorable. But rules are more understandable to people.

In a decision table, we can draw useful rules if condition values and decision values are both at hand. Compared with the tree created by ID3 (Quinlan, 1986), the classifier tree created in this paper is shorter if use heuristic information such as information entropy to find a suboptimal reduct----that is as fewer attributes as possible, which means it is one choice to relieve the curse of dimension. It is more suitable for incomplete system because we need not replace the missing value with an assumed value. It can also avoid fragmentation problem involved in popular decision tree algorithms (Han&Kamber, 2000). And most of all, in my view, one distinguished feature of this kind of classifier is that we can draw conclusions at any level of the tree, because we have elements of both condition values and decision values at every level. It means we can possibly obtain more general rules and classifications. For example, we can find rules like “If $C_1=x$ and $C_2=y$ Then $D=a$ or $D=b$”. If *a, b* have a corresponding superordinate level concept, we can draw a more general rule by attribute oriented induction. Concepts in disjunction form are useful in life. For example, “parent” is the disjunction of “father” and “mother”. And we can draw conclusions such as “Any swan has only two alternative colors: that is white or black”, etc.

The classifiers (rule system and classifier tree) built by algorithm TWI-SQUEEZE also has their shortages. One of them is that although they are fit for dealing redundant data, they are sensitive to irrelevant information. Algorithm TWI-SQUEEZE doesn’t build in valid solutions for removing meaningless attributes itself. It is leaved to preprocessing phase.

Another possible shortage has to be mentioned is that it is sensitive to noise. Generally speaking, some redundancy is needed to restrain noise. Therefore, terseness seems both its advantage and shortage that need to be verified in practice. Approximate reduction (approximate reduct and approximate value reduction) and data transformation seems to be helpful to this problem. Finding a robust and scalable version is a challenging open problem.

In conclusion, this kind classifier can be regarded as a complement of other classifiers, such as decision tree algorithm ID3 and C4.5. No single method is superior. There are many criterions, such as accuracy, training time, robustness, interpretability, scalability, etc. All of them should be considered to evaluate a classifier. Nowadays, databases are increasingly large, and as a result, efficiency and scalability have an increasingly weight in building classifiers. Classifier created by algorithm TWI-SQUEEZE is proved to be efficient. It seems also exist a good scalable version. In this view, it is one choice in practical applications.

#### 7.8.2.2 Order-Deterministic Classifier Tree

Must a rule system be consistent to be useful? Generally speaking, it is true. However, it is also common that when we make decisions, we will have more than one choice, given incomplete information. Generally, information is always incomplete. What we should do is to select one choice which is deemed the best or just randomly select one if few information can tell us what to do. Time is possibly more important to our concern. Therefore, some unusual classifier should be built to fit to the case.

Let’s see the classifier tree in Figure 8, or Figure 9 (a). It is consistent in any view, even if we do not follow the rules recommended. Can it be pruned a bit more? The answer is no if consistency is commanded. But if we are restricted to follow the rules given before, we do can prune a bit more to generalize the original classifier further. And it can also ensure classifying the training samples rightly.

As we can see clearly from Figure 9 (b), this tree is more difficult to be understood, but it can classify more objects, including some unseen objects. For instance, an object with $C_6$=0, $C_5$=1, $C_2$=0, or an object with $C_6$=3, $C_5$=0, $C_2$=1, etc.

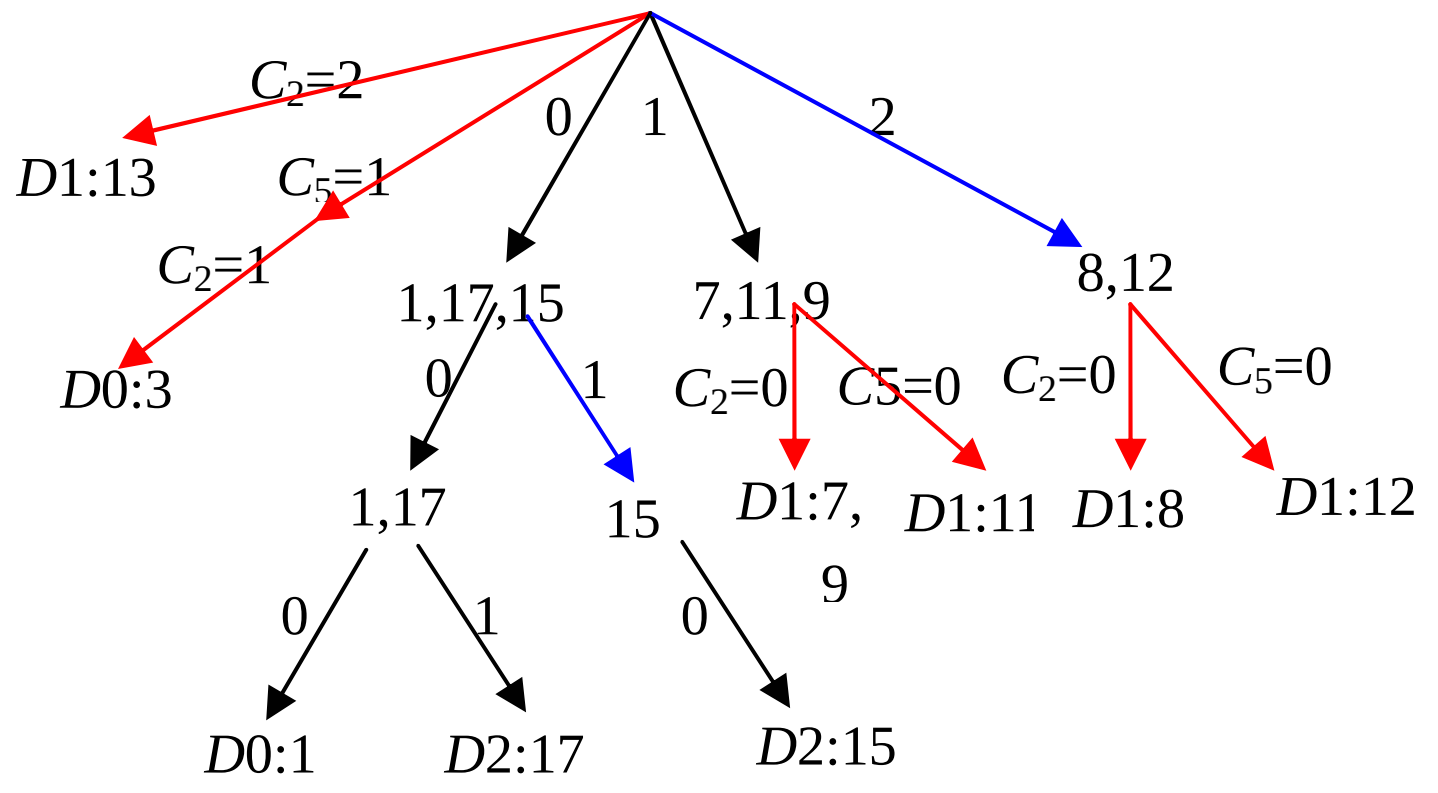

Figure 9 (a): A Classifier Tree

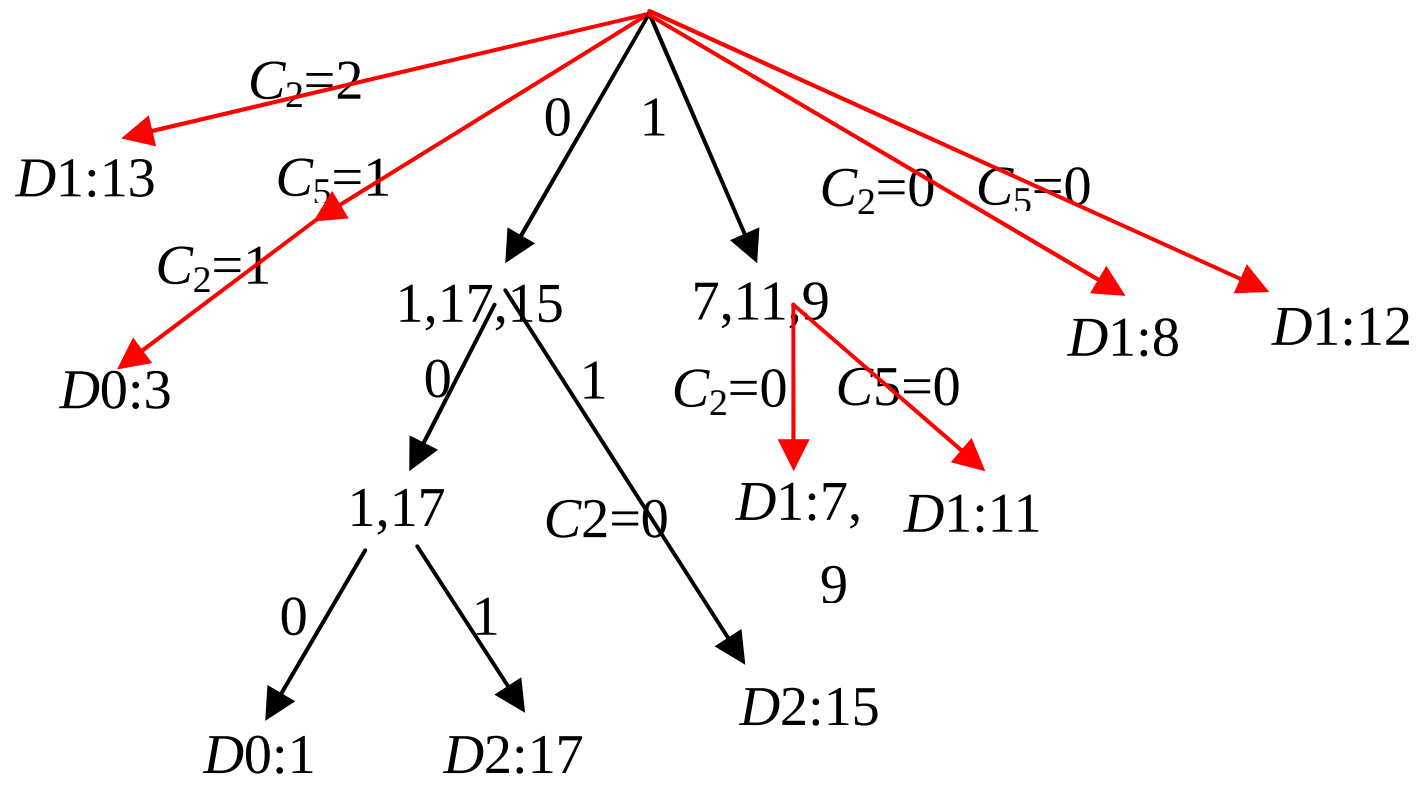

Figure 9 (b): An Order-Deterministic Classifier Tree

It seems that order-deterministic classifier tree has a mechanism which can handle exception flexibly. Indeed, its counterpart is *rules with exception*. In Figure 9 (b), the right branches can be regarded as the exceptions of left branches.

### 7.8.3 Reduction as Inductive Machine Learning: a Casual Discussion

It is too dread a task to discuss even some aspects of machine learning here. Therefore, I'll just select two topics more and discuss them casually.

From the previous discussion, it is clear that knowledge reduction is a kind of selective inductive machine learning. But it seems as if a table's classification ability doesn't change after knowledge reduction. Then, if knowledge reduction is a kind of selective inductive machine learning, where does the generalization comes from?

Generalization seems coming from loss of demarcation information in the reduction process. It means some attributes are no longer needed in discerning some pairs of objects, which also implies shrinking of intension of condition, and increasing of possibility of condition match for new data. If we unfold reduced system, i.e. to list all the objects that match the rules explicitly, then we will find that generalization doesn't come from loss of demarcation information. On the contrary, it comes from increase of potential demarcation information, or variety in the system. When the variety reaches its maximum, the system becomes not generalizable, or cannot further be learned, as had been firstly argued by Ashby. Therefore, generalization comes from loss of actual demarcations, and increase of potential demarcations.

It is clear that value reduction is a kind of generalization. Whether attribute reduction is a kind of generalization or not is decided by the result of the reduction. If it doesn't change the demarcation relation, in my view it cannot be called inductive machine learning in current universe, although it possibly does reduce some redundant attributes. However, because a table is generally a sample of the whole universe, attribute reduction does can be regarded as one kind of generalization process in this sense, because the intension of conditions of rules is shrunk.

The role of exception in rule system has been discussed for long. Wang jue, et al. reported through experiments that if some exceptions are disparted from original table, we can obtain a much more concise rule system (Wang, 1997, Wang, 1998), which proved the claim of Nosofsky et al.(1994,1997).

In my view, the role of exception can be explained in a similar way. If some exceptions are disparted from original table, some actual demarcations will be lost, which means a more concise description of original system and a stronger generalization of samples.

Like ID3 (Qinlan, 1986), the inductive bias of heuristic algorithm introduced in section 5 is that shorter trees are preferred over longer trees[62]. Still like ID3, its bias is preference bias. After reduction, we can transform the reduced table to rules. So far, the rules we've talked about have only the form "If A=? Then B=?". "=" is a weak restriction in the forms of hypotheses. If we can use "≠" and "or", more concise and general rules can possibly be obtained. Therefore, it deserves to try assimilating other ideas, such as AQ algorithms (Michalski, 1969, 1986). That is, to maximize the generalization of rules, we can use "A≠{*b*}" instead of to let "A=*a*".

For example, a rule set (or a table) is:

(If) $a$=0 and $b$=0 (Then) $d$=0;
(If) $a$=1 and $b$=1 (Then) $d$=0;
(If) $a$=0 and $b$=1 (Then) $d$=1;
(If) $a$=0 and $b$=2 (Then) $d$=1;

Using the original algorithm, this rule set can be reduced to:

| | | |
|---|---|---|
| (If) $a$=* and $b$=0 (Then) $d$=0; | | (If) $a$=* and $b$=0 (Then) $d$=0; |
| (If) $a$=1 and $b$=* (Then) $d$=0; | or | (If) $a$=1 and $b$=1 (Then) $d$=0; |

[62] Occam's razor has found its empirical supports in many psychological domains. Its counterpart in psychology is known as simplicity principle, which claims that human tend to choose the pattern that provides the briefest representation of the available information (Chater, 1999).

(If) $a$=0 and $b$=1 (Then) $d$=1;　　(If) $a$=0 and $b$=1 (Then) $d$=1;
(If) $a$=0 and $b$=2 (Then) $d$=1;　　(If) $a$=* and $b$=2 (Then) $d$=1;

Using "≠", the rule set can be reduced to:

(If) $a$=* and $b \neq$ {1, 2} (Then) $d$=0;
(If) $a \neq 0$ and $b$=* (Then) $d$=0;
(If) $a \neq 1$ and $b \neq 0$ (Then) $d$=1;

or

(If) $a$=* and $b \neq$ {1, 2} (Then) $d$=0;
(If) $a \neq 0$ and $b \neq 2$ (Then) $d$=0;
(If) $a \neq 1$ and $b \neq 0$ (Then) $d$=1;
(If) $a$=* and $b \neq$ 0, 1 (Then) $d$=1;

However, such "maximal generalized" rules will cause inconsistency when classifying unseen objects, if the united domain of attributes of sample doesn't equal to that of the universe. For example, object "(If) $a$=0 and $b$=3 (Then) $d$=0" matches both rule "(If) $a$=* and $b \neq$ {1, 2} (Then) $d$=0" and rule "(If) $a \neq 1$ and $b \neq 0$ (Then) $d$=1", which means this object can not classified explicitly.

Although we can adopt some strategies to settle inconsistency, a more prudent scheme is not to use "≠" but "or", if do not concern noise. For example, the rules "(If) $a$=0 and $b$=1 (Then) $d$=1" and "(If) $a$=0 and $b$=2 (Then) $d$=1" can be rewritten as: "(If) $a$=0 and $b$=1 or 2 (Then) $d$=1".

What has been discussed above seems only a small amendment. But in some cases, it will influence the results much and therefore deserves to adjust relational operator in this way.

### 7.9. Tolerance Relation Based Model

There are some works that extends classical equivalence relation based rough set theory to tolerance relation based rough set theory (Komorowski, Pawlak, & Skowron, 1998). I've given a still more general framework, in which various models can be embedded in. It is based on cover because we cannot represent knowledge without concepts (subsets of universe). That's why I think no needs to further generalize the framework, at least in most cases. The elements of cover can be regarded as concepts. These elements needn't have a core, or they can have several unequal cores. I think it is reasonable because although centrality is a wide phenomenon in natural concepts, it is possible that there is more than one prototype in a concept. For example, orange and apple are both the prototypes of fruit (the points are 1.07 and 1.08 respectively (Rosch, 1975)).

There will be too many concepts, which are probably overlapped much, if we use tolerance relation to induce concepts. Instead, human incline to use economy principle in using concepts and terms to describe events. It is one reason why I use cover to represent a concept and leave the construction of concrete covers to be fulfilled later[63]. The other reason is that it is easier to extend to other models if structural element is abstract concept as cover.

Skowron and Stepaniuk (1996) have proposed an algorithm that is concerned of finding relative tolerance reduct. An example is used to cast some light on it. I've proved my thoughts based on tolerance relation in Appendix A. Hence, I will also give a solution to this example in two ways. The second way is recommended.

[63] For example, constraints can be used to restrict covers, like: 1) the maximum distance of two objects in a cover should below a given threshold; 2) average distance between two cover should larger than another threshold. Condition 1) can generally lead covers to tolerant sets.

In this example, two objects are regarded as discernible if their hamming distance is at least 2. And $\beta = 0$. The table can be transformed to a table like Table 11(1).

The point is what should be retained in cascade task. This time, difference in one attributes no longer make two objects distinct. It is clear that nothing can be discerned at layer Eyes. Therefore, $lT_{Eyes}$={{1, 2, 3, 6, 8}, {4, 5, 7}}.

From Table 11(2), we can see: $lT_{Hair}$={{1, 2, 6, 8}, {5, 7}; {1, 2}, {4}}.

| | Height | Weight | Hair | Eyes | d |
|---|---|---|---|---|---|
| 1 | 0 | 0 | 0 | 1 | 1 |
| 2 | 1 | 1 | 0 | 1 | 1 |
| 3 | 1 | 1 | 0 | 0 | 1 |
| 4 | 1 | 1 | 2 | 1 | 2 |
| 5 | 0 | 0 | 1 | 1 | 2 |
| 6 | 1 | 1 | 1 | 0 | 1 |
| 7 | 1 | 1 | 1 | 1 | 2 |
| 8 | 0 | 0 | 1 | 0 | 1 |

Table 11(1): An example (Skowron,et al, 1996)

| | Height | Weight | Hair | Eyes | d |
|---|---|---|---|---|---|
| 3 | 1 | 1 | 0 | 0 | 1 |
| 1 | 0 | 0 | 0 | 1 | 1 |
| 2 | 1 | 1 | 0 | 1 | 1 |
| 6 | 1 | 1 | 1 | 0 | 1 |
| 8 | 0 | 0 | 1 | 0 | 1 |
| 5 | 0 | 0 | 1 | 1 | 2 |
| 7 | 1 | 1 | 1 | 1 | 2 |
| 4 | 1 | 1 | 2 | 1 | 2 |

Table 11(2): An example (Skowron,et al, 1996)

| | Height | Weight | Hair | Eyes | d |
|---|---|---|---|---|---|
| 1 | 0 | 0 | 0 | 1 | 1 |
| 8 | 0 | 0 | 1 | 0 | 1 |
| 3 | 1 | 1 | 0 | 0 | 1 |
| 2 | 1 | 1 | 0 | 1 | 1 |
| 6 | 1 | 1 | 1 | 0 | 1 |
| 5 | 0 | 0 | 1 | 1 | 2 |
| 7 | 1 | 1 | 1 | 1 | 2 |
| 4 | 1 | 1 | 2 | 1 | 2 |

Table 11 (3): An example (Skowron,et al, 1996)

From Table 11 (3), we can see: $lT_{Weight}$={{1, 8}, {5}; {2, 6}, {7}; {2}, {4}}.

From Table 11 (3), we can also see: $lT_{Height}$={{1, 8}, {5}; {2, 6}, {7}; {2}, {4}}. $lT_{Weight}$= $lT_{Height}$. Therefore, attribute “Height” can be removed and cut objects 1,8,2,6,5,7,4 from original table. Renew cascade tasks. That is cut the objects 1,8,2,6,5,7,4 from cascade tasks, which results in: $lT_{Eyes} = lT_{Hair} = lT_{Weight} = \phi$ . Now we get a new consistent table:

| | Weight | Hair | Eyes | d |
|---|---|---|---|---|
| 3 | 1 | 0 | 0 | 1 |

Table 11 (4): An example (Skowron,et al, 1996)

There is only one deterministic object or rule in the table! It means no attribute is needed to keep *required* classification ability of the table with given similarity measure and precision, because even if all attributes are retained, only object 3 can be discriminated from others. It is strange. But it is a sound result to our requirement.

It is clear the similarity measure and precision are not a good choice for this table and the value “2” of attribute “d” is lost in the reduction. We can therefore introduce an object, for example object 4, to complement the losing value. And we can get a new reduced table as Table 11 (5). But the better method is to adjust either similarity measure or $\beta$.

| | Hair | Eyes | d |
|---|---|---|---|
| 3 | 0 | 0 | 1 |
| 4 | 2 | 1 | 2 |

Table 11 (5): An example (Skowron,et al, 1996)

Note that the corresponding reduct in the paper of Skowron and Stepaniuk (1996) is {Height, Hair, Eyes}. The reduced table is Table 12. In fact, it is an inconsistent table. For instance, object 1 and object 5 cannot be discerned by condition attributes with respect of hamming distance threshold, but can be discerned by decision attribute. In a word, the reduced table in Table 11 (4) or (5) retains the positive region and *required* classification ability of original table and is enough to serve our objective.

| | Height | Hair | Eyes | d |
|---|---|---|---|---|
| 1 | 0 | 0 | 1 | 1 |
| 2 | 1 | 0 | 1 | 1 |
| 3 | 1 | 0 | 0 | 1 |
| 4 | 1 | 2 | 1 | 2 |
| 5 | 0 | 1 | 1 | 2 |
| 6 | 1 | 1 | 0 | 1 |
| 7 | 1 | 1 | 1 | 2 |
| 8 | 0 | 1 | 0 | 1 |

Table 12: A reduced table (Skowron,et al, 1996)

The objects that are cut from original table are the objects in boundary region. They are not useless in most case, and cannot be discarded simply. How to extract knowledge from boundary region? Indeed the main function of variable precision rough set model is just a try to solve the problem. There are other possible means to deal with it. However, it is not my concern in this paper and therefore I leave it to be solved.

From the example, we can see that it cost more time in sorting for information system based on tolerance relation than that based on equivalence relation. One way to lessen this problem is to scan part of the database first and move those attributes that can offer more relative demarcations on the right side. For example, from Table 11 (2), we can see the total demarcations induced by attribute Height is: $S\left(IND(Height), d\right) = 2\times 2 + 3\times 1 = 7$. It is also easy to compute that of other attributes: $S\left(IND(Weight), d\right) = 2\times 2 + 3\times 1 = 7$; $S\left(IND(Hair), d\right) = 3\times 3 + 2\times 1 = 11$; $S\left(IND(Eyes), d\right) = 3\times 3 = 9$. Therefore, we should put "Hair" and "Eyes" on the right side, and put "Height" and "Weight" on the left side.

A better method, which can avoid repeated sorting, is to sort one column every time as before. However, some rules should be followed:

1. Give every segment a mark initiated with 0. When a segment is to be removed if in the case of Pawlak's model, add its mark with 1. A segment can be removed from cascade task if and only if its mark is 2. In fact, the mark denotes how many times an object in the segment can be discerned from other objects. If mark equal to 2, it means the object can be discerned from others by given similarity measure based on hamming distance.

2. In the process of forming subtasks, two objects to be discerned should consider both their values and marks before renewing their marks. Let $\delta(x, y) = 1$ if values of object $x$ and $y$ are different, else $\delta(x, y) = 0$. Then it is clear that if $\delta(x, y) + mark(x) + mark(y) < 2$, these two objects need to be discerned by other attributes and therefore should be put in the subtasks. Otherwise, these two objects are discerned.

Table 13 (1) and 18 (2) show the reduction process. The subscript is the mark of corresponding object.

| | Height | Weight | Hair | Eyes | d |
|---|---|---|---|---|---|
| 3 | 1 | 1 | $0_2$ | $0_1$ | 1 |
| 8 | $0_1$ | $0_1$ | $1_1$ | $0_1$ | 1 |
| 6 | $1_1$ | $1_1$ | $1_1$ | $0_1$ | 1 |
| 1 | $0_1$ | $0_1$ | $0_1$ | $1_0$ | 1 |
| 2 | $1_1$ | $1_1$ | $0_1$ | $1_0$ | 1 |
| 5 | $0_0$ | $0_0$ | $1_0$ | $1_0$ | 2 |
| 7 | $1_0$ | $1_0$ | $1_0$ | $1_0$ | 2 |
| 4 | $1_1$ | $1_1$ | $2_1$ | $1_0$ | 2 |

Table 13 (1):An example ( Skowron,et al, 1996)

From Table 13 (1), we have:

$lT_{Eyes}$: $\{\{3_1, 8_1, 6_1, 1_0, 2_0\}, \{5_0, 7_0, 4_0\}\}$; $lT_{Hair}$: $\{\{8_1, 6_1\}, \{5_0, 7_0\}; \{1_1, 2_1\}, \{5_0, 7_0, 4_1\}\}$;

$lT_{Weight}$: $\{\{8_1\}, \{5_0\}; \{6_1\}, \{7_0\}; \{1_1\}, \{5_0\}; \{2_1\}, \{7_0, 4_1\}\}$;

$lT_{Height}$: $\{\{8_1\}, \{5_0\}; \{6_1\}, \{7_0\}; \{1_1\}, \{5_0\}; \{2_1\}, \{7_0, 4_1\}\}$;

We can see that still only the object 3 is the deterministic rule in the table with given requirement. And the algorithm is as efficient as it is applied to Pawlak's model. Now let us adjust $\beta$ from 0 to 0.4 and see what will happen.

Let's observe $lT_{Weight}$ on the above. The subtask $\{\{2_1\}, \{7_0, 4_1\}\}$ now can be solved with attribute "Weight" because $\frac{Card(\{2_1\})}{Card(\{7_0, 4_1\})} \approx 0.33$. Therefore, we have a new $lT_{Weight}$ : $\{\{8_1\}, \{5_0\}; \{6_1\}, \{7_0\}; \{1_1\}, \{5_0\}\}$. Now $lT_{Height}$ is $\{\{8_1\}, \{5_0\}; \{6_1\}, \{7_0\}; \{1_1\}, \{5_0\}\}$. Since $lT_{Weight} = lT_{Height}$, attribute "Height" should be removed. Remove objects 8, 5, 6, 7, 1 from the original table and renew cascade tasks.

| | Weight | Hair | Eyes | d |
|---|---|---|---|---|
| 3 | 1 | $0_2$ | $0_1$ | 1 |
| 2 | $1_1$ | $0_1$ | $1_0$ | 1 |
| 4 | $1_1$ | $2_1$ | $1_0$ | 2 |

Table 13 (2):An example (Skowron,et al, 1996)

$lT_{Eyes}$: $\{\{3_1, 2_0\}, \{4_0\}\}$; $lT_{Hair}$: $\{\{2_1\}, \{4_1\}\}$;

$lT_{Weight} = lT_{Height} = \{\{2_1\}, \{4_1\}\}$. Therefore, attribute "Weight" should be removed.

| | Hair | Eyes | d |
|---|---|---|---|
| 3 | $0_2$ | $0_1$ | 1 |
| 2 | $0_1$ | $1_0$ | 1 |
| 4 | $2_1$ | $1_0$ | 2 |

Table 13 (3):An example (Skowron,et al, 1996)

Skowron and Stepaniuk have proposed a tolerance approximation space model long before(Skowron & Stepaniuk, 1996). Discernibility relation is used to define tolerance reduct of non-decision table. I also find independently that demarcation relation or discernibility relation can be used in finding reducts efficiently (He, 2003), not only in decision table but also in non-decision table, not only in consistent table but also in inconsistent table. In this paper, a tolerance approximation space model, which is similar to that of Skowron and Stepaniuk's, is introduced to prove the soundness of algorithm TWI-SQUEEZE (See Appendix A).

However, there is a restriction in the models. That is, if two objects cannot be discerned by a set of attributes, they also cannot be discerned by any subset of these attributes. An identical restriction is: if two objects can be discerned by a subset of attribute set *C*, they can also be discerned by attribute set *C*. This restriction seems not reasonable. For example, we can give such a similarity measure: $\frac{H(x,y)}{\|x\|}$. Here $H(x,y)$ is the Hamming distance, and $\|x\|$ is the number of attributes that are used to describe object $x$. Two objects are similar if $\frac{H(x,y)}{\|x\|}$ exceeds a given threshold, e.g. $\alpha$. It's a common sense that if two objects share enough attribute values (exceed a given ratio), they can be called similar to some extent. Or else they cannot be called similar.

Does it be in contradiction with our restriction? Not really. The crucial point is that we can use absolute scale to measure similarity. In this instance, we can use $H(x,y)-(1-\alpha)t_C$ instead to determine whether two objects are similar or not. If $H(x,y)-(1-\alpha)t_C<0$, these two objects are similar, or else they are different and form a demarcation.

## 8. Compare Computational Complexity of Various Algorithms

In this section, some attribute reduction algorithms will be compared by their computational complexity. I'll only discuss algorithms that are complete.

Discernibility Matrix and Discernibility Fuction (Skowron, &Rauszer, 1992) is a method that is suitable to find all reducts in moderate scale table. It can also find a redut in $O\left(t^2N^2\right)$ with space complexity $O\left(tN^2\right)$. In my view, discernibility matrix is fit for the case when $\frac{Card(core_attributes)}{Card(reduct)}$ is relative large, or there are a large ratio of missing values in table (therefore it is a sparse matrix), or it can be reduced to a low dimension matrix.

The efficient algorithm proposed by Hoa and Son (1996) is a method in rough set literature, which is based on primitive definition in rough set theory. It is a good choice in practical applications[64]. Assume there are $t_C$ condition attributes and one decision attribute; the cardinality of $U$ is $N$. Its time complexity is $O\left(t_C{}^2N\ln N\right)$, and total space complexity is $O\left(t_CN\right)$, with additional space complexity only $O\left(N\right)$. It is easy to prove that the best case time complexity is half of the worst case time complexity. It can find a suboptimal reduct using Johnson's strategy, which is based on occurrence frequency of an attribute in discernibility matrix. The heuristic information can be computed in $O\left(t_C{}^2 N \ln N\right)$.

[64] Indeed, when I finished my thesis (He, 2003), I think it is a significant improvement to the previous works. But after reading Hoa and Son's paper (1996), I realized that my work is only a minor improvement in the sense of worst case computational complexity. Although I still believe my algorithms have better performance in the sense of average complexity by some theorectical analysises, it is difficult to be analyzed thoroughly. The same is the algorithm in this paper.

As has analyzed in section 2.3, algorithm TWI-SQUEEZE can find a reduct in $O\left(t_C N \ln N\right)$, with $O(N)$ memory space. The improvement seems not remarkable. But we should notice that its best case time complexity is only $O(1)$. And I have also shown that in many ordinary cases, its time complexity is small. Further more, this algorithm has more advantages than efficiency, which has been discussed thoroughly in the paper.

I've discussed the heuristic version of algorithm TWI-SQUEEZE in section 5. Demarcation information can be computed in $O\left(t_C{}^2 N \ln N\right)$. There are many heuristic algorithms, either complete or incomplete. All the algorithms based on genetic algorithms are ignored here because their computational complexity cannot be analyzed thoroughly. We can only talk about the complexity of fitness functions. But it is possible that they have good performance in applications.

Wang et al. (2002) has compared time complexities of three heuristic algorithms based on conditional entropy and mutual information. MIBARK (Miao & Hu, 1999) and CEBARKCC have a $O\left(t_C N^2\right)+O\left(N^3\right)$ time complexity; CEBARKNC has a $O\left(t_C{}^2 N\right)+O\left(t_C N^3\right)$ time complexity.

Some algorithms use various sorts of attribute significance, which are based on rough set theory, to find a reduct (Pawlak, 1991; Hu, Pao, Yu, 2002). These algorithms are characterized by computing positive regions with respect to different condition attribute subsets repetitively. Therefore, their performance won't exceed algorithm of Hoa and Son (1996).

As for non-decision table, or information system, Guan and Bell (1998) has given an algorithm which can find one reduct in $O\left(t_C{}^3 N^2\right)$, based on significance of attribute. There are other specially designed but incomplete algorithms. Algorithm TWI-SQUEEZE is complete and it can find a reduct in $O\left(t_C N \ln N\right)$. As has proved that non-decision table can be transferred to a consistent decision table (Yang et al., 2002), other algorithms that are designed for decision table, such as Hoa and Son's, are also applicable to non-decision table.

Value reduction has received less concern previously. Several representative algorithms can be found in the monograph of Wang (2001). Some algorithms are incomplete, e.g. the frequently cited heuristic value reduction algorithm (Chang et al., 1999; Wang, 2001). Its time complexity is $O(t_C N^2 \ln N)$, and space complexity is $O\left(t_C N\right)$. Another example can be regarded as the counterpart of forward selection method in attribute reduction. The thought is to add more values to precomputed core values. Soundness of this sort of algorithms is also tend to be violated if we ignore the fact that adding one more value will probably cause more values redundant. The classical method, which was introduced by Pawlak (1991), Liu (2001a),

Wang (2001) et al., is complete. But it is not efficient, with time complexity of $O(t_C{}^2 N^2 \ln N)$ and space complexity of $O(t_C N)$. Decision matrices proposed by Ziarko and Shan (1993) can be used to find all "maximal generalized rules" (Hu, 1995), which means the algorithm of Ziarko et al. (Wang, 2001) is complete and can find all possible reduction of value. Generally, finding a maximal generalized rule of corresponding object is as difficult as finding all maximal generalized rules of this object because a conjunction can be verified to be a maximal generalized rule only when it is assured not to be absorbed by shorter conjunctions. Therefore, just like attribute reduction based on discernibility matrix method, value reduction based on decision matrices is also suitable for moderate scale database and useful in theoretical proofs, but finds its limitation in large scale database because of its time and space consuming (NP-hard) feature.

As has been discussed in section 6, algorithm TWI-SQUEEZE can be used for value reduction. It is complete and efficient in different cases, such as incomplete, inconsistent decision table, with time complexity of $O(t_C N \ln N)$, and space complexity of $O(N)$. And it is even much more efficient than those incomplete algorithms.

It is clear that the worst case time complexity of any algorithms, which is fundamentally based on sorting, is at least $O(t_C NlnN)$. Whether this bound is also applicable to all algorithms based on Turing machine model is still to be proved.

## 9. Conclusions

To my best knowledge, Discernibility Matrix (Skowron &Rauszer, 1992) first use demarcations of objects in finding reducts[65]. I find that we can unify Dissimilarity Matrix and Discernibility Matrix, and the demarcation information defined on the basis of them can be regarded as a measure of information and knowledge of a system, which is simple and neat (see section 5.1), and most of all satisfies key properties of any information measure, such as monotonicity of source alphabet, continuity, and recursivity (however it is not defined on the basis of probability distribution). And algorithm in this paper takes full advantage of an invariable of demarcation information (of a given table), which makes it very efficient.

Shannon entropy is the measure of uncertainty of a given system. Demarcation information is the measure of variety or difference contained in a given system. In my view, they reflect two different aspects of information, and have their suitable application areas respectively. The relationship between uncertainty and difference is explored. In cognitive psychology, distinctiveness and so called informativeness are two elements that make up the "differentiation" of a concept. In this paper, I've proved that demarcation information can not only used to measure distinctiveness, but also can be used to measure information. Two measures, which were regarded as different before, are unified into one form now. And it has many applications, in my view, among which, algorithm TWI-SQUEEZE is an illustrative example. It can also used in other applications, such as clustering, association mining, knowledge discovery, pattern recognition, inductive machine learning, etc.

[65] Extension Matrix (Hong, 1997) first uses difference compare of two objects to maintain necessary information in partitioning positive samples and negative samples.

Algorithm TWI-SQUEEZE has many remarkable properties. It is the fastest algorithm at present; we can run the algorithm nearly in complete parallel; it is scalable and can adapt to the changing world efficiently; perform this algorithm twice, both complete attribute reduction and value reduction are at hand; and the result of knowledge reduction are two different kinds of classifiers; it is easy to combine with any heuristic information (but demarcation information measure is recommended); it can integrate user preference to find a local optimal attribute subset; it can also be used in feature selection and knowledge reduction and it can also be regarded as a process of selective inductive machine learning.

Another distinguished feature of algorithm TWI-SQUEEZE is that it can be applied to different requirements, such as applications in incomplete situation (without changing data), variable precision model, tolerance model and the combination of them. In a word, it seems that algorithm TWI-SQUEEZE can easily adapt to different models without changing the original algorithm framework. Can all the reduction problems in different rough set models be resolved in one algorithm framework, as this paper has implied? It is still a problem before the other rough set models are studied, such as probabilistic rough set model and fuzzy rough set model, which were introduced by Zhang et al. (2001). The answer is probably no. But the algorithm can do work as long as there is a cover defined on universe. Therefore the author guess a weak conclusion holds: any reduction that can be done traditionally in a rough set model based on equivalence relation or reasonable tolerance relation can also be done by this algorithm framework with some adaptation.

All the concepts in rough set theory not only can be built on the concepts of equivalence relation and equivalence class, but also can be established on the basis of demarcation relation and demarcation information, although "indiscernibility" has been regarded as a building block in rough set theory. And when knowledge reduction is concerned, the combination of both families of concepts can have better achievements than only using one of them. Although I've proved the equivalence of the main ideas of attribute reduction that is based on rough set theory and that underlies algorithm TWI-SQUEEZE (Appendix A), this paper is not a try to substitute rough set theory, a milestone, models of which not only have wide applications based on knowledge reduction, but also have deepened our knowledge of the world. Instead, it offers one possible practical way to realize the ambition of the theory. That is to let the data speak as much as possible for itself, when the data is sufficiently good and representative, without worrying about the scale of data much (Drewry et al., 2002).

There is no doubt that algorithm TWI-SQUEEZE is efficient. However, efficient does not mean effective. Generally speaking, it is hard to achieve ideal efficiency and effectiveness in the same time. As the algorithm is concerned, the problem is that it can only find a suboptimal reduct or approximate reduct efficiently. Although the problem is still remained, I've discussed how to find variant reducts in an efficient way (section 5.3, 7.7). Another shortage should be noted is this algorithm lack a matching efficient enough method that computes some heuristic information (no matter what it is) in the stage of attribute reduction, without violating the ability of finding a good enough suboptimal reduct. It is an open problem.

To overcome noise and overfitting, there are many methods, such as discretization and value induction (section 7.5), approximate reduction (section 6.4, 7.3, 7.8.1), sampling (section 7.6), etc. It deserves to point out again that algorithm TWI-SQUEEZE can achieve good effect only when values have generalized or it has built in data preprocess or it is based on a rightly selected model (such as tolerance model), instead of using it as a panacea.

Algorithm TWI-SQUEEZE also has other shortages. For example, it doesn't build in some mechanism to preclude meaningless attributes (not so called "*irrelevant*" attributes *for the data*, from the viewpoint of classification and description, which is discussed in section 2.4. "Meaningless" is not discussed in the context of the data we are training and testing, but in the context of both seen and *all* unseen data), which will probably influence the result of knowledge reduction that is used for prediction. It is very possible that some attribute happens to partition the examples well, despite being unrelated to the actual target function, as have discussed before. And it is separate-and-conquer in nature, which will limit its application in the situation when partition induced by a set of attributes cannot be fulfilled by linear combination of demarcation effect of individual attributes.

In this paper, knowledge reduction, especially attribute reduction is discussed. Taper-off-Sorting plays an important role in these algorithms. In fact, some techniques used in this paper can also be used in other applications, such as finding large itemsets and association rules. It needs only one scan to find all the large itemsets. It can also be used in unsupervised learning. For example, we can use both the average variety between clusters and inside clusters to find the proper number (sequence) of clusters. And it can also find functional dependency of two attribute subsets very efficiently, and so on.

My future work will continue to study on demarcation information and its applications in data mining and machine learning.

## Appendix A: Soundness of the thoughts

I will prove my thoughts using definitions of demarcation/separation and approximation space models here. The classical approximation space that is based on Pawlak rough set model is a special case of the generalized models. Although there is a restriction in these models, I'll discuss how to loose this restriction for algorithm in paper.

**Definition A1:** $Cov=\{X_1,X_2,\cdots,X_n\}$ is a succinct cover of $U$, which satisfy: $X_i-\bigcup_{j\neq i}X_j\neq\phi$, and $\bigcup_i X_i=U$. From now on, we use "cover" to denote succinct cover. For $\forall x,y\in U$, $(x,y)$ is a ***demarcation*** **with respect to** $X_i$, if $x\in X_i, y\notin X_i$; or $x\notin X_i, y\in X_i$.

**Definition A2:** $Cov=\{X_1,X_2,\cdots,X_n\}$ is a cover of $U$. $x\in X_i, y\in X_j$. $(x,y)$ is called a ***demarcation*** if $x,y\notin X_i\bigcap X_j$. ***Let*** $S(Cov)$ ***be the set of all the demarcations induced by*** $Cov$. It can also be called ***demarcation relation.*** If $Cov$ is induced by attribute set $P$, we use $S(Cov,P)$ to denotes all the demarcations induced by $Cov$.

**Theorem A1:** Let $K=(U,C\bigcup D)$, if $C$ can induce an equivalence relation $IND(C)$, we have:

$\forall x, y \in U, (x, y)$ is a ***demarcation*** induced by $IND(C)$, if $(x, y) \notin IND(C)$. This is a ***narrowed*** definition of demarcation only applicable in classical approximation space based on indiscernibility relation.

**Proof:** $IND(C)$ induce a partition $\{X_1, X_2, \cdots, X_n\}$ of $U$, which satisfy that for $\forall x, y \in X_i$, $(x, y) \in IND(C)$. It is easy to prove that $X_i \cap X_j = \phi$, when $i \neq j$, or else $X_i = X_j$. Hence, $Z_i = X_i$. For $\forall x, y,\ (x, y) \notin IND(C),\ \exists i, j,$ satisfy $x \in X_i, y \in X_j$. If $i = j$, then $(x, y) \in IND(C)$. Therefore, $i \neq j$. And $x \in Z_i, y \in Z_j$. Therefore $(x, y)$ is a ***demarcation*** induced by $IND(C)$. ■

Another simpler but equal definition is that: $\forall x, y \in U$, $(x, y)$ is called a ***demarcation*** induced by $IND(C)$, if $x \neq y$.

**Definition A3:** $Cov = \{X_1, X_2, \cdots, X_n\}$ is a cover of $U$, let $Z_i = \left( X_i - \bigcup_{p \neq i} X_p \right)$, $i \in \{1, 2, \cdots, n\}$, we have: $(x, y)$ is a ***determinate demarcation*** **with respect to** $Cov$ **iff** $\exists i, j \in \{1, 2, \cdots, n\}, i \neq j$ **s.t.** $x \in Z_i, y \in Z_j$. This definition is cover based. ***Let*** $DS(Cov)$ ***be the set of all the determinate demarcations induced by*** $Cov$. If $Cov$ is induced by attribute set $P$, we use $DS(Cov, P)$ to denotes all the determinate demarcations induced by $Cov$.

**Definition A4:** $Cov$ is a cover of $U$. $S(X, Y, Cov)$ is the set of demarcations induced by $X, Y \in Cov$, if $S(X, Y, Cov) = \{(x, y) \mid x \in X - Y \wedge y \in Y - X$ or $y \in X - Y \wedge x \in Y - X\}$.

From the definition, we have:

$S(X, \overline{X}, Cov) = \{(x, y) \in U \times U \mid x \in X \wedge y \in \overline{X}$ or $x \in \overline{X} \wedge y \in X\}$. $S(X, X, Cov) = S_R(X) = \phi$.

**Definition A5(**Skowron&Stepaniuk, 1996**):** $\nu : P(U) \times P(U) \times [0, 0.5) \to [0, 1]$ is a rough inclusion function, which satisfies:

$\nu(X, Y, \beta) = f_\beta(t)$, where $t = Card(X \cap Y) / Card(X)$, for any $\phi \neq X, Y \subseteq U$ and $0 \leq \beta < 0.5$.

Generally, $f_\beta(t) = \begin{cases} 0 & if \quad 0 \leq t \leq \beta \\ (t - \beta)/(1 - 2\beta) & if \quad \beta < t < 1 - \beta \\ 1 & if \quad 1 - \beta \leq t \leq 1 \end{cases}$

***Note that variable precision model can be transformed to classical model as discussed in section 7.3.1. Therefore, from now on, we assume*** $\beta = 0$.

Borrow the thoughts of Skowron and Stepaniuk (1996), we introduce the concept of “structural element” and “structural function”.

**Definition A6 (**Skowron&Stepaniuk, 1996**):** For a given structural function $F : P(U) \to \{0, 1\}$, $\forall X_i \in P(U)$ is called a $F$ -structural element of $U$ if $F(X_i) = 1$.

**Definition A7: An approximation space** of $K = (U, C \cup D)$ is a system

$$R=(U,L,F,S,POS)$$

where $U$ is a non-empty, finite set, $L:U\times U\times \mathrm{P}(C)\to\{0,1\}$ is a map which induces approximation space covers $Cov_C$ and $Cov_D$ of $U$. It must satisfies: $L(x,y,C)=1\Rightarrow L(x,y,A)=1,\forall A\subseteq C$ (*). $F$ is a structural function, which determines approximation space covers with $L$. $S$ is a measure of demarcation information. $POS$ is an approximation operator to be defined, which determines positive region. In fact, the following theorems are just to disclose relation between demarcation information measure and positive region based on given definitions.

**Remark A1: (*) is a restriction for algorithm TWI-SQUEEZE, which means the original algorithm should be amended a bit to fit some cases, such as the example 2 discussed by Skowron and Stepaniuk (1996). I'll discuss this example in details in the paper, and show how to solve this problem if we chose a relative flexible attitude to "demarcation". The tolerance approximation space model proposed by Skowron and Stepaniuk also includes this restriction.**

**1. Tolerance Approximation Space**

**Because classical rough set model, or Pawlak's rough set model, is a special case of tolerance approximation space (Skowron & Stepaniuk, 1996),** we will discuss how algorithm TWI-SQUEEZE can adapt to fit for the requirements of this generalized rough set model.

**Definition A8 (Skowron & Stepaniuk, 1996):** A binary relation $R$ is a tolerance relation if it is reflexive and symmetric.

**Definition A9:** $K=(U,C\bigcup D)$. For $\forall A\subseteq C$, $R_A$ is a tolerance relation induced by $SIM(x,y,A):U\times U\times \mathrm{P}(C)\to\{0,1\}$, if for $\forall(x,y)\in R_A$, we have $SIM(x,y,A)=1$; whereas for $\forall(x,y)\notin R_A$, we have $SIM(x,y,A)=0$. $SIM(x,y,A)$ must satisfies: 1) $SIM(x,x,A)=1$; 2) $SIM(x,y,A)=SIM(y,x,A)$; 3) $\forall A\subseteq B\subseteq C$ s.t. $SIM(x,y,B)=1\Rightarrow SIM(x,y,A)=1$. It means that if $x$ and $y$ are discerned by condition subset $A$, they can also be discerned by the whole condition set $C$.

**Definition A10:** $K=(U,C\bigcup D)$. $R_D$ is a tolerance relation induced by $SIM(x,y,D):U\times U\times D\to\{0,1\}$, if for $\forall(x,y)\in R_D$, we have $SIM(x,y,D)=1$.

Note that maybe $SIM(x,y,A)|_{A\subseteq C}$ is different from $SIM(x,y,D)$.

**Definition A11:** $Cov_A=\{X_1,X_2,\cdots,X_m\}$ is called a tolerance approximation space cover of $U$ with respect to condition attribute set if it is induced by a tolerance relation $R_A$ on $K=(U,C\bigcup D)$, which satisfy:

1. $R_A$ is induced by $A\subseteq C$ with $SIM(x,y,A)$;
2. For $\forall x\in X_i, y\in X_j$, we have $(x,y)\in R_A$ if $i=j$.

3. $x \in X_i, y \in X_j$. If $x, y \notin X_i \cap X_j$, $(x,y) \notin R_A$.

**Definition A12:** $Cov_D = \{Y_1, X_2, \cdots, Y_n\}$ is called a tolerance approximation space cover of $U$ with respect to decision attribute set if it is induced by a tolerance relation $R_D$ on $K=(U, C \cup D)$, which satisfy:

1. $R_D$ is induced by decision attribute set $D$ with $SIM(x,y,D)$;
2. For $\forall x, y \in X_i$, we have $(x,y) \in R_D$.
3. $x \in X_i, y \in X_j$. If $x, y \notin X_i \cap X_j$, $(x,y) \notin R_D$.

**Theorem A2:** $S(Cov_A) = \overline{R_A}, S(Cov_D) = \overline{R_D}$. $Cov_D, Cov_A$ are tolerance approximation space covers.

**Proof:** From Definition A6, condition 3, we know $S(Cov_A) \subseteq \overline{R_A}$.

For $\forall x \in X_i, y \in X_j$, if $(x,y) \in \overline{R_A}$, then $i \neq j$. Therefore, $x, y \notin X_i \cap X_j$, or $x = y$. As a consequent, $(x,y) \in S(Cov_A)$. Hence, $S(Cov_A) = \overline{R_A}$. Similarly, we can prove $S(Cov_D) = \overline{R_D}$. Therefore, $\overline{R_A}$ and $\overline{R_D}$ can also be called demarcation relation or discernibility relation.■

**Definition A13:** $R$ is a tolerance relation in $K=(U, C \cup D)$. For $\forall X \in U$, $x \in X$ is called the ***Core of*** $X$, if $\forall y \in X$ s.t. $(x,y) \in R$.

**Remark A2:** In the tolerance approximation space model proposed by Skowron and Stepaniuk (1996), only the sets that have at least one core can be called structural elements. If $x_i$ is a core of $X_i$, $I(x_i)$ can be used to represent $X_i$.

**Definition A14: For** $\forall A \subseteq C$, $I(x,A) = \{X_i \in Cov_A \mid for \forall y \in U, SIM(x,y,A) = 1 \Rightarrow y \in X_i\}$. Similarly, $I(x,D) = \{Y_i \in Cov_D \mid for \forall y \in U, SIM(x,y,D) = 1 \Rightarrow y \in Y_i\}$.

**Definition A15:** For a given $Cov_A$ $(A \subseteq C)$ and $\forall X \subseteq U$, the tolerance lower approximation of set $X$ is: $tLw(Cov_A, X) = \bigcup \{x \in U \mid \nu(I(x,A), X, \beta) = 1\}$.

**Definition A16:** For a given $Cov_A$ $(A \subseteq C)$ and $\forall X \subseteq U$, the tolerance upper approximation of set $X$ is: $tUp(Cov_A, X) = \bigcup \{x \in U \mid \nu(I(x,A), X, \beta) > 0\}$.

**Definition A17:** In $R=(U,L,F,S,POS)$, the tolerance positive region of $Cov_D$ with respect to $Cov_A$ $(A \subseteq C)$ is:

$$POS(Cov_A, Cov_D) = \bigcup \{x \in U \mid \nu(I(x,A), I(x,D), \beta) = 1\} = \bigcup tLw(Cov_A, I(x,D))|_{x \in U}.$$

**Definition A18:** In $R=(U,L,F,S,POS)$, $L = SIM$, which induces $Cov_C = \{X_1, X_2, \cdots, X_m\}$ And $Cov_D = \{Y_1, Y_2, \cdots, Y_n\}$. For any $x, y \in U$, if $SIM(x,y,A) = 0$ $(A \subseteq C)$ and $SIM(x,y,D) = 0$, $(x,y)$ is called a ***relative tolerance demarcation of*** $R$. **Let** $TS(Cov_D, A)|_{A \subseteq C}$ **be all the relative tolerance demarcations of** $R$ **with respect to** $A$.

**Definition A19:** $R=(U,L,F,S,POS)$ is a tolerance approximation space of $K=(U, C \cup D)$ if

it satisfies:

1. $L=SIM$ is a similarity measure which induces approximation space covers $Cov_C$ and $Cov_D$ of $U$; $SIM$ satisfies definition A15 and A16.

1. $S=TS$;

2. $POS$ satisfies definition A22.

3. $F\left(I(x,B)\right)=1$ for any $x\in U$, and $B\subseteq C$ or $B=D$. Else $F(X)=0$.

**Theorem A3:** $R=(U,L,F,S,POS)$ is a tolerance approximation space of $K=(U,C\bigcup D)$. $POS(Cov_A,Cov_D)\subseteq POS(Cov_C,Cov_D)$, for $\forall A\subset C$.

**Proof:** Since $POS(Cov_A,Cov_D)\subseteq POS(Cov_C,Cov_D)$, for any $x\in POS(Cov_A,Cov_D)-POS(Cov_C,Cov_D)$, $\exists r$ s.t. $r\in I(x,C)-I(x,D)$, but $r\notin I(x,A)$.

Hence, $SIM(x,r,A)=0$, $SIM(x,r,C)=1$. In other words, $L(x,r,A)=0$, but $L(x,r,C)=1$, which is in contradiction with (*).■

**Theorem A4:** $K=(U,C\bigcup D)$, $P=\{c_1,c_2\cdots,c_k\}\subseteq C$. $P^{'}=P-\{c_i\}$. $R=(U,L,F,S,POS)$ is a tolerance approximation space of $K=(U,C\bigcup D)$. We have:

$$POS(Cov_P,Cov_D)=POS(Cov_{P^{'}},Cov_D)\ \text{ if }\ TS(Cov_D,P)=TS(Cov_D,P^{'}).$$

**Proof:** Since $P^{'}\subset P$, from Theorem A2, $POS(Cov_{P^{'}},Cov_D)\subseteq POS(Cov_P,Cov_D)$.

If $POS(Cov_{P^{'}},Cov_D)\neq POS(Cov_P,Cov_D)$, then $POS(Cov_{P^{'}},Cov_D)\subset POS(Cov_P,Cov_D)$.

For any $x\in POS(Cov_P,Cov_D)-POS(Cov_{P^{'}},Cov_D)$, $\exists r$ s.t. $r\in I(x,P^{'})-I(x,D)$, but $r\notin I(x,P)$. Hence, $SIM(x,r,P)=0$ and $SIM(x,r,D)=0$.

Therefore, $(x,r)$ is a relative tolerance demarcation of $R$.

Since $(x,r)\in TS(Cov_D,P)$, but $(x,r)\notin TS(Cov_D,P^{'})$. We have: $TS(Cov_D,P)\neq TS(Cov_D,P^{'})$.

Contradiction. ■

**2. Soundness of Algorithm TWI-SQUEEZE**

**Theorem A5:** Algorithm TWI-SQUEEZE is effective for Tolerance Approximation Space.

**Proof Sketch:**

Because in rough set theory, if $POS(Cov_P,Cov_D)=POS(Cov_{P-\{c_i\}},Cov_D)$, $c_i$ is $D$-dispensable in $P$, else $c_i$ is $D$-indispensable in $P$. Therefore, we can use the concept "demarcations" instead of "positive region" to find the reduct, since theorem A4 is proved. **And as a consequence, for the attribute set $\{C_1,C_2,\cdots C_n\}$, $C_i$ is dispensable to the other attributes and can be removed without violating the consistency of classifications, if and only if the relative demarcations induced by attribute $C_i$ with respect to decision attributes D is the subset of that of the other attributes.**

Similar conclusion can also be found for reduction in a non-decision table. It is because non-decision table can be regarded as a special decision table.

This thought tells us that if we scan condition set $C$ from right to left, we can obtain a subset $s_reduct \subseteq C,$ in which any attribute is indispensable to the attributes in its right side. After the second scan, we obtain another $s_reduct \subseteq C,$ in which any attribute is indispensable to the attributes in its both sides. For attribute $C_i$, cascade task intersection $IST_i$ indicates the demarcations that cannot be offered by other attributes. If $IST_i \neq \phi,$ it means $C_i$ is indispensable to other attributes and therefore should not be removed. Otherwise it is dispensable to others and can be removed. With the check of cascade task intersection, we at last know whether the attribute is indispensable to all the other leftover attributes or not, and gradually get a reduct by removing dispensable attributes. ■

Thus, I've proved the soundness of algorithm by connect "demarcation" to "positive region". We can also prove that **algorithm TWI-SQUEEZE is effective for other rough set models**, such as the rough set model based on Strict Indiscernibility Relation (Slowinski, 1993; Stepaniuk, 1997), in a similar way. In fact, most concepts of classical rough set theory can be represented in a different way. That is not based on equivalence class and indiscernibility, but on the concept of demarcations. For example, the elements in boundary region of a table are those elements that can be demarcated by decision attributes, but cannot be demarcated by condition attributes; the elements in positive region of table are those elements that can be demarcated by both decision attributes and condition attributes, etc.

## Appendix B: How to judge whether there is some objects (record) that cannot be discerned by all the attributes $s_reduct$ except $C_k$?

First of all, we have to answer *how to judge if two segments share some objects (records)?*

This problem is very easy to solve. Assume that these two segment are $seg1$ and $seg2$, and $seg1.addr_first = x1, seg1.addr_last = x2; seg2.addr_first = y1, seg2.addr_last = y2.$ It is easy to prove that if $(x1 - y2)\cdot(x2 - y1) \leq 0$ then $seg1$ and $seg2$ share at least one object. Otherwise they share no objects, and we can further see that if $(x1 - y2) + (x2 - y1) < 0$ then the address of all the objects in $seg1$ is smaller than that of $seg2$.

I call this kind of computation as "conglutinative match", which can determine whether the intersection of two segments is a void set or not.

Now let's answer the question left. If $lT'$ and $lT_{k+1}$ have two or more pairs of D-Regions overlapped respectively, then the overlapped part of these D-Regions cannot be discerned by all the attributes in $s_reduct$ except $C_k$.

We can compare the tasks one by one. Because the segments in any task have an ascendant order, the time complexity of match is only $O(N)$.

Below is the example that has been shown in table 2. After having read the whole algorithm and its example, we can dip into some details now.

If we have:

$lT_6 = \{T_{60}: [1,4]_0, [7,8]_1, [15,18]_2; T_{61}: [5,6]_0, [9,12]_1\}$, and

$lT' = lT_2' = \{T_{20}': [1,2]_0, \{[9,9],[11,12],[14,14]\}_1, [17,18]_2; T_{21}': [3,4,5,6]_0, \{[10,10],[13,13]\}_1, [15,16]_2\}$,

how to compute $lT_2' \overset{+}{\cap} lT_6$ ?

Firstly, we compare $T_{60}$ and $T_{20}'$. Because $[1,4]_0$ contains $[1,2]_0$, $[1,4]_0$ and $[1,2]_0$ are matched. We continue to compare $[7,8]_1$ and $\{[9,9],[11,12],[14,14]\}_1$. Because $[7,8]_1$ is in front of $[9,9]_1$, therefore we continue to compare $[15,18]_2$ and $\{[9,9],[11,12],[14,14]\}_1$. Because 14<15, we continue to compare $[15,18]_2$ and $[17,18]_2$ and they are matched. Therefore, $[1,2]_0$ and $[17,18]_2$ cannot be discerned by $\{C_2, C_5, C_6\}$ except $C_5$.

Afterwards, we compare $T_{60}$ and $T_{21}'$. Then compare $T_{61}$ and $T_{20}'$. And then $T_{61}$ and $T_{21}'$.

At last, we can find that, $lT_2' \overset{+}{\cap} lT_6 = \{[1,2]_0, [17,18]_2; [5,6]_0, [10,10]_1\}$.

Assuming the number of segments in two cascade tasks are *Na* and *Nb* respectively, the time cost in match is $O(Na+Nb)$. Therefore, the time complexity in matching of two cascade task is $O(N)$.

## Appendix C: Compute Heuristic Information (a small example)

Table TC1 (a) is a small decision table $KS = (U, C \cup D)$. $P \leftarrow C$. After removal of repetitive objects, Table TC1 (a) is transformed to Table TC1 (b).

Let $S(C_i, D)$ be the amount of relative demarcation information of $C_i$ relative to $D$. $I(C_i, D)$ be the amount of relative indiscernible information of $C_i$ relative to $D$.

$U/D = \{X_k \mid U = \cup X_k (X_i \cap X_j = \phi, i \neq j)\}$, $S(D) = \sum_{i,j} Card(X_i) \cdot Card(X_j)$. And $S(D) = S(C_i, D) + I(C_i, D)$.

Because $S(D)$ is a constant, either $S(C_i, D)$ or $I(C_i, D)$ can be used as heuristic information.

As for this example, $S(D) = 4 \times 3 = 12$.

Generally speaking, $I(C_3, D)$ should be computed in the way given in section 5 as:

$$I(C_3, D) = 0.5\{((2+1)^2 - (2^2 + 1^2)) + ((2+2)^2 - (2^2 + 2^2))\} = 6;$$

But in this case, the relative indiscernible information will be computed in a simpler way:

| $C_1$ | $C_2$ | $C_3$ | $D$ |
|---|---|---|---|
| 0 | 0 | 0 | 0 |
| 1 | 0 | 0 | 0 |
| 1 | 0 | 1 | 0 |
| 1 | 1 | 1 | 0 |
| 1 | 1 | 1 | 0 |
| 1 | 1 | 1 | 0 |
| 0 | 0 | 0 | 1 |
| 0 | 1 | 1 | 1 |
| 1 | 0 | 1 | 1 |
| 1 | 0 | 1 | 1 |
| 1 | 0 | 1 | 1 |

**Table TC1 (a)**

| $C_1$ | $C_2$ | $C_3$ | $D$ |
|---|---|---|---|
| 0 | 0 | 0 | 0 |
| 1 | 0 | 0 | 0 |
| 1 | 0 | 1 | 0 |
| 1 | 1 | 1 | 0 |
| 0 | 0 | 0 | 1 |
| 0 | 1 | 1 | 1 |
| 1 | 0 | 1 | 1 |

**Table TC1 (b)**

$I(C_3, D) = 2\times 1 + 2\times 2 = 6;$ $S(C_3, D) = S(D) - I(C_3, D) = 6.$

$I(C_2, D) = 3\times 2 + 1\times 1 = 7;$ $S(C_2, D) = S(D) - I(C_2, D) = 5.$

$I(C_1, D) = 1\times 2 + 3\times 1 = 5;$ $S(C_1, D) = S(D) - I(C_1, D) = 7.$

$I(C_1, D) < I(C_3, D) < I(C_2, D).$ Therefore, $C_1$ should be included to $s_reduct$ first.

(*Note that if we compute occurrence frequency of an attribute in discernibility matrix without removal of repetitive objects first,* $C_2$ *should be included to* $s_reduct$ *first.*)

$P \leftarrow \sigma(P)$ is a permutation of $P$. $\sigma = \begin{pmatrix} 1 & 2 & 3 \\ 2 & 3 & 1 \end{pmatrix}.$

Now $T_3$ is formed. $T_3 = \{queue_0 : [1,2]_0, [5,5]_1; queue_1 : [3,4]_0, [6,7]_1\}$. $Q_3 = e(T_D) - e(T_3) = \phi.$

Table TC1 (b) turns to Table TC2 now.

| $C_2$ | $C_3$ | $T_3$ | $D$ |
|---|---|---|---|
| 0 | 0 | x | 0 |
| 0 | 0 | y | 0 |
| 0 | 1 | y | 0 |
| 1 | 1 | y | 0 |
| 0 | 0 | x | 1 |
| 1 | 1 | x | 1 |
| 0 | 1 | y | 1 |

**Table TC2**

In Table TC2, *x* and *y* can be any value only if *x* is not equal to *y*.

$I(lT_3) = 1\times 2 + 3\times 1 = 5.$ Why not count $S(lT_3)$? Because $I(lT_3)$ reflects the amount of

demarcation information needed to be offered by $C_2$ and $C_3$. It acts just like $S(D)$.

$I(C_3, lT_3) = 1\times 1 + 2\times 1 = 3;$ $S(C_3, lT_3) = S(lT_3) - I(C_3, lT_3) = 2.$

$I(C_2, lT_3) = 1\times 1 + 2\times 1 = 3;$ $S(C_2, lT_3) = S(lT_3) - I(C_2, lT_3) = 2.$

$I(C_3, lT_3) = I(C_2, lT_3)$ . Therefore, $C_2$ or $C_3$ should be included to $s_reduct$. Assume $C_3$ is included. $P \leftarrow \sigma(P)$, $\sigma = \begin{pmatrix} 1 & 2 & 3 \\ 1 & 2 & 3 \end{pmatrix}$.

At last, $KS = (U, P \bigcup D)$. **No actual movement of data is needed.**

**Remark:** $C_1$ seems "lost" in Table TC2 because we don't concern it from now on.

## Appendix D: Property of Demarcation Information Measure

**Definitions are described in section 5. To make recursivity property applicable, all the proofs listed below will be built on some distribution.**

**Property 1:** $\forall s \subseteq \mathrm{P}(F), \varphi(x \in s) \geq 0.$

**Proof:** For any set $X$, $Card(X) \geq 0$. Therefore, $\varphi(s) \geq 0$. ■

**Property 2:** Assume $F = (X_1, X_2, \cdots, X_n)$ and $F' = (X_1', X_2', \cdots, X_n')$ are two partitions of universe, which are induced by two attributes $C_1$ and $C_2$ respectively. Let $C_i(X_j)$ be the corresponding attribute value of $X_j$ under $C_i$.

If $S(\{x\}, \overline{X_i}) \bigcap S(\{x\}, \overline{X_j}) = \phi$, then we have:

$$\varphi(\{x \mid x = C_1(X_i) \wedge C_2(X_j)\}) = Card(S(\{x\}, \overline{X_i})) + Card(S(\{x\}, \overline{X_j})).$$

**Proof:** From definition we have it. ■

**Property 3:** Assume $F = (X_1, X_2, \cdots, X_n)$ is a partitions of universe, then, we have:

$$\varphi(x \in X_i \bigcup X_j, X_i \neq X_j) < \varphi(x \in X_i);\ \varphi(x \in X_i \bigcup X_j, X_i \neq X_j) < \varphi(x \in X_j).$$

**Proof:** $\overline{X_i \bigcup X_j} \subset \overline{X_i}$. Therefore, for $\forall y \in \overline{X_i \bigcup X_j}$, it is also true that $y \in \overline{X_i}$. Hence,

$$\varphi(x \in X_i \bigcup X_j, X_i \neq X_j) \leq \varphi(x \in X_i);$$

And since $\overline{X_i \bigcup X_j} \subset \overline{X_i}$, $\exists y \in \overline{X_i} - \overline{X_i \bigcup X_j}$ s.t. $(x, y)$ is counted in $\varphi(x \in X_i)$, but $(x, y)$ isn't counted in $\varphi(x \in X_i \bigcup X_j, X_i \neq X_j)$.

Hence, $\varphi(x \in X_i \bigcup X_j, X_i \neq X_j) < \varphi(x \in X_i);$

Similarly, we can prove $\varphi\left(x \in X_i \bigcup X_j, X_i \neq X_j\right) < \varphi\left(x \in X_j\right)$.

This property can be called monotonic degression with extension of concepts. ■

**Remark D1:** We can use $\varphi\left(x \in X_i\right) - \varphi\left(x \in X_i \bigcup X_j, X_i \neq X_j\right)$ to denote measure of conditional information.

More generally, $s \subseteq e \subseteq U$, $\varphi(x \in s | x \in e) = \varphi(x \in s) - \varphi(x \in e)$ can be regarded as the measure of demarcation information difference between two concepts.

**Property 4:** $\varphi\left(x \in U\right) = 0$.

**Proof:** From definition we have it. ■

**Property 5 (Permutational Symmetry):** $F = \left(X_1, X_2, \cdots, X_n\right)$, $\sigma : \left\{X_1, X_2, \cdots, X_k\right\} \to \left\{Y_1, Y_2, \cdots, Y_n\right\}$ is a permutation where $\sigma(X_i) = Y_i$. Let $F^{'} = \left(Y_1, Y_2, \cdots, Y_n\right)$, we have: $\varpi\left(F\right) = \varpi\left(F^{'}\right)$.

**Proof:** From definition of function $\varpi\left(F\right)$ we have the conclusion. ■

**Property 6:** $\varpi\left(F\right) \geq 0$.

**Proof:** For any set $X$, $Card\left(X\right) \geq 0$. Therefore, $\varpi\left(F\right) \geq 0$. ■

**Property 7:** $F = \left(X_1, \phi, \cdots, \phi\right) = \left(\phi, X_2, \cdots, \phi\right) = \left(\phi, \phi, \cdots, X_n\right) = 0$. $\varpi\left(F\right) = 0$.

**Proof:** From definition of function $\varpi\left(F\right)$ we have the conclusion. ■

***Property 8 (Monotonicity)** $F = \left(X_1, X_2, \cdots, X_n\right)$. If $Card\left(X_1\right) = Card\left(X_2\right) = \cdots = Card\left(X_n\right) = \dfrac{N}{n}$, then $\varpi\left(F\right) = \tau\left(n\right) = \dfrac{n-1}{n} N^2 = \left(1 - \dfrac{1}{n}\right) N^2$. It means in this case $\varpi\left(F\right)$ is a monotonically increasing function of $n$.

**Proof:** From definition of function $\varpi\left(F\right)$ we have the conclusion. ■

**Property 9:** $\varpi\left(F\right)$ reaches its sole maximum when $Card(X_1) = Card(X_2) = \cdots = Card(X_n)$.

**Proof:** It is clear that $\sum_{i<j} (x_i - x_j)^2 \geq 0$.

Hence, $(n-1)\sum_i x_i^2 \geq 2\sum_{i<j} x_i x_j$. Equality holds when $x_1 = x_2 = \cdots = x_n$.

We also have: $N^2 = \left(\sum_i x_i\right)^2 = \sum_i x_i^2 + 2\sum_{i<j} x_i x_j \geq \dfrac{2}{n-1}\sum_{i<j} x_i x_j + 2\sum_{i<j} x_i x_j = \dfrac{2n}{n-1}\sum_{i<j} x_i x_j$.

Equality holds when $x_1 = x_2 = \cdots = x_n$.

Therefore, $\varpi(F) = 2\sum_{i<j} x_i x_j \le \frac{n-1}{n} N^2$, equality holds when $x_1 = x_2 = \cdots = x_n$. ■

**Property 10:** $F = (X_1, X_2, \cdots, X_n)$ and $F' = (X_1', X_2', \cdots, X_n')$ are two partitions of universe.

If $\sum_{i<j}(x_i - x_j)^2 \le \sum_{i<j}(x_i' - x_j')^2$, then $\varpi(F) \ge \varpi(F')$.

**Proof:** $(n-1)\left(\sum_i x_i\right)^2 - \sum_{i<j}(x_i - x_j)^2 = (n-1)N^2 - \sum_{i<j}(x_i - x_j)^2 = 2n\sum_{i<j} x_i x_j$ .

Hence, $\varpi(F) = 2\sum_{i<j} x_i x_j = \frac{(n-1)}{n} N^2 - \frac{1}{n}\sum_{i<j}(x_i - x_j)^2$ . $\varpi(F') = 2\sum_{i<j} x_i' x_j' = \frac{(n-1)}{n} N^2 - \frac{1}{n}\sum_{i<j}(x_i' - x_j')^2$ .

$$\varpi(F) - \varpi(F') = \frac{1}{n}\sum_{i<j}(x_i' - x_j')^2 - \frac{1}{n}\sum_{i<j}(x_i - x_j)^2.$$

Therefore, if $\sum_{i<j}(x_i - x_j)^2 \le \sum_{i<j}(x_i' - x_j')^2$, then $\varpi(F) \ge \varpi(F')$. ■

**Property 11:** $F = (X_1, X_2, \cdots, X_n)$ and $F' = (X_1', X_2', \cdots, X_n')$ are two partitions of universe.

If $\sum_i (x_i - \overline{x})^2 \le \sum_i (x_i' - \overline{x'})^2$, then $\varpi(F) \ge \varpi(F')$. Here, $\overline{x} = \overline{x'} = \frac{1}{n}\sum_i x_i = \frac{N}{n}$ .

**Proof:** $\left(\sum_i x_i\right)^2 - \sum_i (x_i - \overline{x})^2 = N^2 - \sum_i (x_i - \overline{x})^2 = 2\sum_{i<j} x_i x_j + 2\overline{x}\sum_i x_i - n \cdot \overline{x}^2$

$$= 2\sum_{i<j} x_i x_j + \frac{N^2}{n} .$$

Hence, $\varpi(F) = 2\sum_{i<j} x_i x_j = \left(N^2 - \frac{N^2}{n}\right) - \sum_i (x_i - \overline{x})^2$ . $\varpi(F') = 2\sum_{i<j} x_i' x_j' = \left(N^2 - \frac{N^2}{n}\right) - \sum_i (x_i' - \overline{x'})^2$ .

$$\varpi(F) - \varpi(F') = \sum_i (x_i' - \overline{x'})^2 - \sum_i (x_i - \overline{x})^2.$$

Therefore, if $\sum_i (x_i - \overline{x})^2 \le \sum_i (x_i' - \overline{x'})^2$, then $\varpi(F) \ge \varpi(F')$. ■

Although $\varpi$ is defined on discretized universe, we can only define it on continuous universe. Let $\varpi'$ be such a function. Then $\varpi$ can be regarded as the result of periodic sampling of $\varpi'$.

**Property 12:** $C \subset R^n$, which satisfies that for any $a = (x_1, x_2, \cdots, x_n) \in C$, $x_i > 0$ and $\sum_i x_i = N$ .

Then $\varpi^{'}(a)=2\sum_{i<j}x_ix_j$ is a convex down function.

**Proof:** Assume $p,q\in C,\ p=(x_1,x_2,\cdots x_n),q=(y_1,y_2,\cdots y_n).\ 0\le\lambda\le 1.$

Clearly, $\lambda p+(1-\lambda)q\in C.$ Therefore, $C$ is a convex set.

Let $a=(x_1,x_2,\cdots,x_n),b=(x_1^{'},x_2^{'},\cdots,x_n^{'}).$ $a,b\in C$.

$$\left(\sum_i(x_i-x_i^{'})\right)^2=\sum_i(x_i-x_i^{'})^2+2\sum_{i<j}(x_i-x_i^{'})(x_j-x_j^{'})=0.$$

Since $\sum_i(x_i-x_i^{'})^2\ge 0,$ therefore $2\sum_{i<j}(x_i-x_i^{'})(x_j-x_j^{'})\le 0.$

Hence, $2\sum_{i<j}\lambda(1-\lambda)(x_ix_j+x_i^{'}x_j^{'}-x_i^{'}x_j-x_ix_j^{'})\le 0.$

$$2\lambda\sum_{i<j}x_ix_j+2(1-\lambda)\sum_{i<j}x_i^{'}x_j^{'}\le 2\sum_{i<j}(\lambda x_i+(1-\lambda)x_i^{'})(\lambda x_j+(1-\lambda)x_j^{'}).$$

$$\lambda\varpi^{'}(a)+(1-\lambda)\varpi^{'}(b)\le\varpi^{'}(\lambda a+(1-\lambda)b).$$

Therefore, $\varpi^{'}(a)=2\sum_{i<j}x_ix_j$ is a convex down function. ■

***Property 13 (Continuity):**

$a,b\in C\subset R^n.\ a=(x_1,x_2,\cdots,x_k,\cdots,x_n),b=(x_1-\varepsilon_1,x_2-\varepsilon_2,\cdots,x_k+\varepsilon,\cdots,x_n-\varepsilon_n).$

Here: $\varepsilon_j\ge 0\ (j\ne k);\ \sum_{j\ne k}\varepsilon_j=\varepsilon>0;\ \ 0\le x_j-\varepsilon_j\le N\ (j\ne k);\ \ 0\le x_k+\varepsilon\le N.$

We have: $\lim_{\substack{\varepsilon\to 0\\ \varepsilon_j\to 0}}\varpi^{'}(b)=\varpi^{'}(a).$

**Proof:** Because $\varepsilon_j\ge 0\ (j\ne k)$ and $\sum_{j\ne k}\varepsilon_j=\varepsilon>0,$ we have: $\varepsilon\to 0,\ \varepsilon_j\to 0\ (j\ne k).$

It is easy to prove that $(x_i+p)(x_j+q)$ is continuous function, for any $p,q\in R.$

Therefore, $\lim_{\substack{\varepsilon\to 0\\ \varepsilon_j\to 0}}\varpi^{'}(b)=\lim_{\substack{\varepsilon\to 0\\ \varepsilon_j\to 0}}2\left\{\sum_{\substack{i<j\\ i,j\ne k}}(x_i-\varepsilon_i)(x_j-\varepsilon_j)+(x_k+\varepsilon)\sum_{i\ne k}(x_i-\varepsilon_i)\right\}$

$$=2\left\{\sum_{\substack{i<j\\ i,j\ne k}}(x_i)(x_j)+(x_k)\sum_{i\ne k}(x_i)\right\}$$

$$=2\sum_{i<j}x_ix_j=\varpi^{'}(a).$$

■

**Remark:** This property tells us that when distribution has a small change, the result won't have a tremendous change. This implication also holds for $\varpi(F)$.

**Property 14:** $F=(X_1,X_2,\cdots,X_k,\cdots,X_n)$. $F'=(X_1,X_2,\cdots,X_{k-1},X_1',X_2',\cdots,X_t',X_{k+1},\cdots,X_n)$.

$\sum_{i=1}^{n} x_i = \sum_{\substack{i=1\\ i\neq k}}^{n} x_i + \sum_{i=1}^{t} x_i' = N$, when $x_i = Card(X_i)$ and $x_i' = Card(X_i')$.

Let $F_1 = (X_1', X_2', \cdots, X_t')$, we have: $\varpi(F') = \varpi(F) + \varpi(F_1)$.

**Proof:** $x_k = \sum_{i=1}^{t} x_i'$, because $\sum_{i=1}^{n} X_i = \sum_{\substack{i=1\\ i\neq k}}^{n} X_i + \sum_{i=1}^{t} X_i' = N$.

$$\varpi(F') = 2\sum_{\substack{i<j\\ i,j\in[1,n]\\ i,j\neq k}} x_i x_j + 2\sum_{\substack{i\in[1,n]\\ i\neq k\\ j\in[1,t]}} x_i x_j' + 2\sum_{\substack{i<j\\ i,j\in[1,t]}} x_i' x_j'$$

$$= 2\sum_{\substack{i<j\\ i,j\in[1,n]}} x_i x_j + 2\sum_{\substack{i<j\\ i,j\in[1,t]}} x_i' x_j'$$

$$= \varpi(F) + \varpi(F_1).$$

■

***Property 15 (Recursivity):**

$$F_1=(X_1,X_2,\cdots,X_k,\cdots,X_n).\ \ F_2=(Y_1,Y_2,\cdots,Y_k,\cdots,Y_m).\ \ \sum_{i=1}^{n} x_i = \sum_{i=1}^{m} y_i = N.$$

Let $F_1\cap F_2=(X_1\cap Y_1,\cdots,X_1\cap Y_m,X_2\cap Y_1,\cdots,X_2\cap Y_m,\cdots,X_n\cap Y_1,\cdots,X_n\cap Y_m)$. And $F_{2i}=(X_i\cap Y_1,\cdots,X_i\cap Y_m)$

$(i=1,\cdots,n)$. We have: $\varpi(F_1\cap F_2) = \varpi(F_1) + \sum_{i=1}^{n} \varpi(F_{2i})$.

**Proof:** Let $z_{ij} = Card(X_i \cap Y_j)$.

$$\varpi(F_1\cap F_2) = \sum_{\substack{i\in[1,n]\\ j\in[1,m]}} \sum_{\substack{s\in[1,n]\\ t\in[1,m]}} z_{ij} z_{st}$$

$$= \sum_{\substack{i\in[1,n]\\ s\in[1,n]}} \sum_{j\in[1,m]} \sum_{t\in[1,m]} z_{ij} z_{st} + \sum_{i=1}^{n} \sum_{s,t[1,m]} z_{is} z_{it}$$

$$= \varpi(F_1) + \sum_{i=1}^{n} \varpi(F_{2i}).$$

This property is also the corollary of property 14.

■

**Property 16:** $F_1=(X_1,X_2,\cdots,X_k,\cdots,X_n)$. $F_2=(Y_1,Y_2,\cdots,Y_k,\cdots,Y_m)$.

Let $F_1\cap F_2=(X_1\cap Y_1,\cdots,X_1\cap Y_m,X_2\cap Y_1,\cdots,X_2\cap Y_m,\cdots,X_n\cap Y_1,\cdots,X_n\cap Y_m)$. And $F_{2i}=(X_i\cap Y_1,\cdots,X_i\cap Y_m)$ $(i=1,\cdots,n)$. If $X_i\cap Y_1 : X_i\cap Y_2\cdots : X_i\cap Y_m = X_1\cap Y_1 : X_1\cap Y_2\cdots : X_1\cap Y_m$ for any $i\in[1,n]$,

we have: $\varpi(F_1\cap F_2)=\varpi(F_1)+\frac{\varpi(F_2)}{N^2}\sum_{i=1}^{n}x_i^2=\varpi(F_1)+\varpi(F_2)-\frac{1}{N^2}\varpi(F_1)\varpi(F_2)$.

**Proof:** Because $X_i\cap Y_1 : X_i\cap Y_2\cdots : X_i\cap Y_m = X_1\cap Y_1 : X_1\cap Y_2\cdots : X_1\cap Y_m$ for any $i\in[1,n]$,

$\varpi(F_{2i})=\frac{x_i^2}{N^2}\varpi(F_2)$.

Therefore, $\varpi(F_1\cap F_2)=\varpi(F_1)+\sum_{i=1}^{n}\varpi(F_{2i})$

$$=\varpi(F_1)+\sum_{i=1}^{n}\varpi(F_2)\cdot\frac{x_i^2}{N^2}=\varpi(F_1)+\frac{\varpi(F_2)}{N^2}\sum_{i=1}^{n}x_i^2 \cdot$$

Symmetrically, we have: $\varpi(F_1\cap F_2)=\varpi(F_2)+\frac{\varpi(F_1)}{N^2}\sum_{i=1}^{m}y_i^2 \cdot$

Therefore, $\varpi(F_1\cap F_2)=\sum_{i=1}^{n}x_i^2=\varpi(F_1)+\varpi(F_2)-\frac{1}{2N^2}\varpi(F_1)\left[\left(\sum_{i=1}^{n}x_i\right)^2-\sum_{i=1}^{n}x_i^2\right]$

$$-\frac{1}{2N^2}\varpi(F_2)\left[\left(\sum_{i=1}^{m}y_i\right)^2-\sum_{i=1}^{m}y_i^2\right]=\varpi(F_1)+\varpi(F_2)-\frac{1}{N^2}\varpi(F_1)\varpi(F_2).$$

■

**Property 17:** Let $F=(X_1,X_2,\cdots,X_n,Y_1,Y_2,\cdots,Y_m)$ be a partition of universe.

If $F_1=(X_1,X_2,\cdots,X_n)$, $F_2=(Y_1,Y_2,\cdots,Y_m)$ and $F^{'}=(\cup X_i,\cup Y_j)$ then

$\varpi(F)=\varpi(F_1)+\varpi(F_2)+\varpi(F^{'})$.

**Proof:** $\varpi(F)=\sum_{i,j\in[1,n]}x_ix_j+\sum_{i,j\in[1,m]}y_iy_j+\sum_{\substack{i\in[1,n]\\ j\in[1,m]}}x_iy_j$

$$=\sum_{i,j\in[1,n]}x_ix_j+\sum_{i,j\in[1,m]}y_iy_j+\sum_{i\in[1,n]}x_i\sum_{j\in[1,m]}y_j$$

$$=\varpi\left(F_1\right)+\varpi\left(F_2\right)+\varpi\left(F^{'}\right).$$

■

**Property 18:** If $F_1=\left(X_1,X_2,\cdots,X_n\right)$ and $F_2=\left(Y_1,Y_2,\cdots,Y_m\right)$ are two partitions of universe, then $\varpi\left(F_1\cap F_2\right)\le\varpi\left(F_1\right)+\varpi\left(F_2\right)$.

**Proof:** Let $F_{2i}=\left(X_i\cap Y_1,\cdots,X_i\cap Y_m\right)$, $\left(i=1,\cdots,n\right)$. And let $z_{ij}=Card\left(X_i\cap Y_j\right)$.

From property 15, $\varpi\left(F_1\cap F_2\right)=\sum_{\substack{i\in[1,n]\\ j\in[1,m]}}\sum_{\substack{s\in[1,n]\\ t\in[1,m]}}z_{ij}z_{st}=\varpi\left(F_1\right)+\sum_{i=1}^{n}\varpi\left(F_{2i}\right)$.

$$\sum_{i=1}^{n}\varpi\left(F_{2i}\right)=\sum_{i}\sum_{j_1\ne j_2}Z_{ij_1}Z_{ij_2}\le\sum_{i}\sum_{j_1\ne j_2}Z_{ij_1}Z_{ij_2}+\sum_{\substack{i_1\ne i_2\\ j_1\ne j_2}}Z_{i_1j_1}Z_{i_2j_2}=\sum_{j_1\ne j_2}\left(\sum_{i,j_1}Z_{ij_1}\times\sum_{i,j_2}Z_{ij_2}\right)=\varpi\left(F_2\right).$$

Here, $i\in[1,n],\ j_1,j_2\in[1,m]$.

We obtain the equality when $\sum_{\substack{i_1\ne i_2\\ j_1\ne j_2}}Z_{i_1j_1}Z_{i_2j_2}=0$. That is, if $X_{i1}\cap Y_j\ne\phi$ then

$X_{i2}\cap Y_j=\phi\ \left(i_1\ne i_2\right)$. In other words, if $x\in X_{i1}\cap Y_j$ and $x\in X_{i2}\cap Y_j$, then $i1=i2$.

It also means that equality holds only when no demarcation can be induced by both partitions.

■

**Property 19:** Let $F_t=\left(X_1^{\ t},X_2^{\ t},\cdots,X_{nt}^{\ t}\right)\ \left(t\in[1,m]\right)$ is a set of partitions of universe, then $\varpi\left(\cap F_t\right)\le\sum_{t}\varpi\left(F_t\right)$.

**Proof:** It's the corollary of property 18. ■

***Property 20:** Assume the source alphabet is $A=\left\{a_1,a_2,\cdots,a_n\right\}$. Let $OT=\left\{o_1,o_2,\cdots,o_l\right\}\ \left(l\in[0,+\infty)\right)$ be the output letter (object) sequence. Then $\varpi\left(OT\right)\le\varpi\left(A\right)$.

**Proof:** Remove the repetitive objects in $OT$, we obtain $OT^{'}$, which satisfies: $OT^{'}\subseteq A$.

From property 8, we have this property. ■

**Remark:** This property tells us that output variety won't exceed original system variety. As a consequence, any sum of output (demarcation) information is a subset of (demarcation) information contained in the original system.

## Appendix E: Relationship between Uncertainty and Difference

I've proved in this section that difference does can be used as a measure of information. Then what is the relationship between Shannon entropy and (average) demarcation information measure? Or what is the relationship between uncertainty and difference? As we know, Shannon entropy is a measure of uncertainty of a given system. What is uncertainty? A system has some states or objects and we are required to select one and predict what the state or object is. Thus we are faced with uncertainty. To eliminate the uncertainty, some information is needed. Shannon entropy is based on this thought. *The elimination process of uncertainty is a process of selection.* We can choose or select an object rightly only when some information is known. In other words, we are said to be informed only when we can select rightly and definitely. Therefore, information is the thing that makes our selection a *right* one. Then what is the basis of selection and right selection? *The basis of selection is difference and the basis of right selection is awareness of the difference.* In other words, if we know an object is not all the other things in a given system, then we can determine it definitely and select the right label (appellation) for the object. The implication of demarcation information is that we know something because we know it is not other things. The intuition rooted in the classical information theory is that information is the thing that makes someone surprise. Therefore, information measure should be the measure of surprise. Entropy measures are indeed measures for *difference* in distribution, or for equality. But the intuition rooted in demarcation information measure is that information is the thing that tells one object from the others. And hence the measure should be the measure of difference.

In fact, either Shannon entropy or Ashby's variety measure (also is called $H_0$ when it uses logarithm form) can also be regarded as the measure that counts the number of distinct letters in a massage. However, there are two differences between these two measures. One is that, for Ashby, two letters are distinct (or not redundant) if and only if they are different symbol. However, in communication, "*ee*" perhaps has a special meaning and cannot regarded as redundant, e.g. "*bee*" and "*be*". For Shannon, "*ee*" should be counted as two distinct letters. Therefore, Shannon entropy (here means $H_1$) is a measure of the maximum possibly distinct letters in a massage. The other difference is that Shannon entropy also takes the length of codeword [66] into account, which is necessary in communication. The more frequency a letter appears in sequence, the less length it will be coded. Using some *k*-ary unit, this strategy, when it is optimal, will lead to Huffman code, which can make the massage as short as possible[67]. It seems intuitive that the amount of information contained in a message is in direct ratio to the number of distinct letters. Two objects are distinguishable only if their corresponding symbols are different. For communication, this is enough. But in the context of data mining, observer will want to know what an object (which is labeled as "*a*") really is and how can it be made distinct from other objects. He needs more information to know it. As has analyzed, observer can know it only when he is acquainted with the original system, i.e. knowing the demarcation information contained in this system.

Now, we can classify information measures. Shannon entropy and Ashby's variety measure and descriptive complexity can be called linear measure of information. Demarcation information measure can be called quadratic measure, which takes difference into account. Furthermore, we can also define difference of difference, which can be called cubic measure,

[66] It also can be regarded as amount of uncertainty eliminated by the letter.
[67] It can be used as an annotation for why $H_1$ still can not exceed $H_0$ and why Shannon entropy can be generalized to descriptive complexity when output sequence is Bernoulli sequence.

and so on. We can also define different distances, e.g. geodesic distance, for measures.

## Appendix F: Generalized Variety Measure

Demarcation information and its measure is a special case of a more general theory, which I'll only give a concise description here.

Let $A \subseteq U$, $Card(A) = M$, then information function that measure variety (difference) contained in $A$ is $\tau(A) = \sum_{i,j \in [1,M]} w_{ij}$ . $w_{ij}$ is the weighted (semantic) distance of $(x_i, x_j)$, $x_i, x_j \in A$.[68] If $i = j$, $w_{ij} = 0$. When $A = \phi$, let $\tau(A) = 0$. We can use some *k*-ary unit to code the difference, i.e. $\log_k(\tau(A))$.

The difference of $\tau(U)$ from $\varpi(F)$ is that this time we don't discriminate whether $(x_i, x_j)$ is a demarcation or not. In demarcation theory, only demarcations are taken into account to the measure. Therefore, in that theory, not only semantic distance is needed, but a member function is needed. The semantic distances of indiscernible objects are omitted. However, in this more general theory, the semantic distance of indiscernible objects are also taken into account. It is clear that demarcation measure is a special case of this measure. And the key properties of demarcation theory are also true in this theory.

Let $X \subseteq U$, $Card(X) = M$ then $\rho(X) = \frac{\tau(X)}{M^2}$ is the measure of average variety inside $X$. When $X = \phi$, let $\rho(X) = 0$.

Another quantity that can describe variety inside of $X$ is variance of variety: $s = \sqrt{\frac{1}{M^2 - 1} \sum_{i,j \in [1,M]} (w_{ij} - \rho(X))}$ and its relevant statistic quantity: $CV = 100\% \times \frac{s}{\rho(X)}$. Assume $X \neq \phi$.

Let $F = (X_1, X_2, \cdots, X_n)$ is a partition of $U$, $X \subseteq Y \subseteq \mathrm{P}(F)$. $X, Y$ are two concepts. $I(x \in X) = |\overline{X}| \Big/ \left( \frac{1}{|X| \cdot |\overline{X}|} \sum_{\substack{i \in X \\ j \in \overline{X}}} w_{ij} \right) = \frac{|X| \cdot |\overline{X}|^2}{\sum_{i \in X, j \in \overline{X}} w_{ij}}$ is the information contained in "$x \in X$". Here $|X| = Card(X)$. $I(x \in Y) - I(x \in X)$ is the information contained in "$x \in X$" when "$x \in Y$" is already known. Generally, the semantic distance in $X$ is not larger than the semantic distance in $Y$, because when $X \subseteq Y \subseteq \mathrm{P}(F)$, difference in more attributes means larger semantic distance.

---

[68] $w_{ij} = d_{ij} . \delta_{ij}$ . $\delta_{ij}$ can be discrete or continual values, e.g. {0, 1} or [0, 1]. In fact, we can still introduce a more general form. That is, to use $f(\{x_i\}, \{x_j\})$ ($\{x_i\}, \{x_j\} \subset \mathrm{P}(U)$) instead of ordinary distance measure of objects. And $\tau(A) = Card(\{f(\{x_i\}, \{x_j\})\})$. Thus we can talk about some special difference, such as "$\leq$" or even vectors of Haar wavelet coefficients. In my view, difference is not necessary a quantity defined on $R^+$. However, this generalization need more careful and strenuous study and won't be discussed here. Readers are suggested to find appropriate forms of difference to their own advantages.

From the view of this theory, similarity and difference are two sides of one coin, and we can therefore measure "difference" and "similarity" of two sets or concepts in the same way.

(Jain, Murty, Flynn, 1999) has introduced several (dis)similarity measures because of their importance in most clustering procedures. Various distances are discussed when talking about dissimilarity measure, since what is concerned is dissimilarity between two patterns or objects. However, when we compare two sets, we must compare every pairs of objects, which are in these two sets respectively. Therefore, we have a definition from the nature of things:

Let $X,Y \subseteq U$, $X,Y \neq \phi$. $\varphi(X,Y)=\sum_{i\in X, j\in Y} w_{ij}$. $\bar{\varphi}(X,Y)=\frac{d(X,Y)}{|X|\cdot|Y|}$. If $X=\phi$ or $Y=\phi$, $\varphi(X,Y)=0$.

Clearly, $\varphi(X,X)=\tau(X)$. $\bar{\varphi}(X,X)=\rho(X)$. When $X \cap Y=\phi$, $\bar{\varphi}(X,Y)=\frac{1}{|X|\cdot|Y|}\sum_{i\in X, j\in Y} w_{ij}$.

(*)Indeed, this is just the average distance of two classes in clustering and statistic literature (Fan, Mei, 2002). The difference between $X$ and $Y$ can be: $d(X,Y)=\varphi(X,Y)-\varphi(X\cap Y, X\cap Y)$.

The average difference between $X$ and $Y$ can be: $\bar{d}(X,Y)=\frac{\varphi(X,Y)-\varphi(X\cap Y, X\cap Y)}{|X|\cdot|Y-X|+|Y|\cdot|X-Y|+|X\cap Y|^2}$.

When $|X|=|Y|=1$, $\bar{d}(X,Y)$ is just the distance of two objects. It deserves to explain the soundness of $|X\cap Y|^2$ in denominator. One is that when $X=Y$, it is necessary for the soundness of definition. Another more natural reason is that intersection of $X$ and $Y$ is an important element that helps count so called difference and similarity. A small example will cast some light on it.

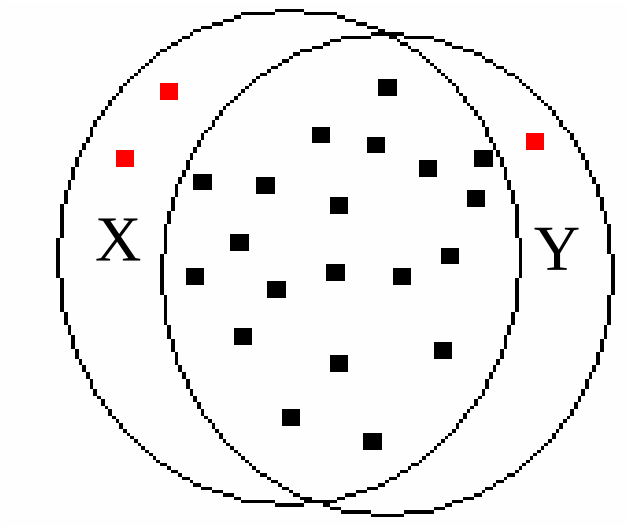

Figure F (a)

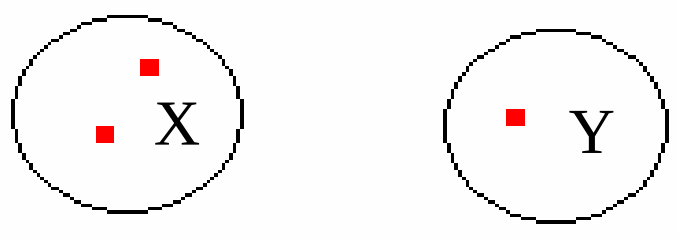

Figure F (b)

In Figure F (a), almost all the elements of set $X$ and $Y$ are in $X \cap Y$. In Figure F (b), the elements in $X-Y$ and $Y-X$ are remained, which are colored with red. It is intuitive that the similarity of $X$ and $Y$ in Figure F (a) are larger than that in Figure F (b). It implies that we cannot ignore the influence of intersection of two sets.

Having the definition of difference between two sets, we can rewrite $I(x \in X)$ as:

$$I(x \in X) = \frac{|\overline{X}|}{\overline{d}(X, \overline{X})}.$$

The measure of similarity is similar. We can use $\frac{1}{\overline{\overline{d}}(X,Y)}$ or $-\overline{d}(X,Y)$ or $\frac{-\overline{d}(X,Y)}{\rho(X \cup Y)}$ to measure similarity. Note that here we is not talking about *degree* of similarity, which is usually regarded as a normalized quantity. In my view, it is not always possible and natural to talk about the degree *exactly*.

We've talked about distance without specifying it. Here the "distance" is not limited to a space distance, but also include semantic distance. Semantemes and meanings come from background knowledge and concepts. In fact, when we discuss similarity by distance here, we first refer to semantic distance. If two objects belong to one concept according to background knowledge, they have zero semantic distance and therefore are extremely similar. We use numerical taxonomy only when there is no semantic distance can be used. The above definition is mainly a kind of "independent" proximity measure, which doesn't take previous knowledge, concept into account. We can also use a complex proximity measure which highlight semantic ingredient.

## Appendix G: Measure of Demarcation Knowledge

It seems we are still debating the criterion of genuine knowledge. It also seems that there are various aspects need to be considered. Truth is surely one aspect of knowledge, but how can we measure truth? Can we say one proposition is "truer" than the other? I doubt about it. Usefulness is another important aspect of knowledge. Can we measure usefulness? I think it is difficult, because it is to do with different people and different interests.

One simple way is to estimate the quantity of positive evidences that support a proposition. But how can we compare two propositions which can apply to infinite situations?

Therefore, a general measure of knowledge is very difficult, if not impossible. What I will discuss here is limited to a given finite system, especially a table, and in a given sense. In (Pawlak, 1991), Pawlak has argued that knowledge can be regarded as abilities to classify objects[69]. I will narrow my discussion in this sense. If there is such a measure, it will benefit some fields, especially classification and pattern recognition.

Generally, a proposition that is composed of higher level concepts contains more knowledge than a proposition that is composed of lower level concepts. We know higher level concepts contain less information, and it is interesting that combination of them will express more knowledge. For example, there are two propositions, "bird can fly" and "bird with yellow speckle wings can fly". These propositions can be transformed to a more general form "A is/belong to B" (in fact, most knowledge can be transformed to this form). For instance, "bird can fly" is identical to "bird is *a kind of creature that can fly*". "Bird with yellow speckle wings" has more information than "bird" in the view of Shannon entropy. But it is clear that "birds can fly" contains more knowledge than "birds with yellow speckle wings can fly". We can explain that by sum up all the uncertainty eliminated by the knowledge. In this view, Shannon entropy can do the work.

[69] "Our claim is that knowledge is deep-seated in the classificatory abilities of human beings and other species". Z. Pawlak (Pawlak, 1991).

Some people will also argue that "bird can fly" is trivial knowledge which offers little information. Yes, "bird can fly" is trivial knowledge, but only for us, who hold it as common sense, not for machines, which do not have such "trivial" knowledge. We should be cautious that "trivial" is relative.

In my view, demarcation information, which tells us an object is not B but A, can also be used to measure information. It reflects a different aspect of information from Shannon entropy. Shannon entropy tells us information is the thing that eliminates uncertainty. And demarcation tells us information is the thing telling one object from another objects. And it can also be used to measure demarcation knowledge naturally. Here, knowledge can be partly false. For example, "bird can fly" is true in most case, but ostrich cannot fly. Therefore, we should not only concern about right demarcations but also false demarcations. For instance, in a table, there is an object which implies "owl can fly", and there is another object which implies "ostrich cannot fly". But we have claimed that ostrich can fly because it is also a kind of bird. Therefore, some false demarcations occur. For example, the object which implies "mouse cannot fly" now is demarcated form the object which implies "ostrich can fly". This demarcation is a false one. Hence, we use a demarcation information measure triplet to measure knowledge: one reflects actual demarcations implied in one proposition; the second reflects mendacious demarcations contained in that proposition; the third reflects the actual demarcation information that is lost in the proposition.

Assume there is a proposition "A is B". A' $\subseteq$ A, which makes "A' is B" a truth. Then the amount of demarcation knowledge contained in this proposition can be measured as:

$$\left(Card\left(A^{'}\right)\times Card\left(\overline{B}\right), Card\left(A-A^{'}\right)\times Card\left(\overline{B}\right), Card\left(A^{'}\right)\times Card\left(A-A^{'}\right)\right)^{70}.$$

Note that, this is not a universal measure of knowledge. The author believes there exist a set of measures which can disclose different aspects of knowledge.

[70] We can use the definition A4 in Appendix A to rewrite it, in the form of demarcation information.